\documentclass[10pt,twocolumn,letterpaper]{article}

\usepackage[final,algorithms]{wacv}      

\usepackage{graphicx}
\usepackage{amsmath}
\usepackage{amssymb}
\usepackage{booktabs}

\usepackage{cuted} 

\usepackage{graphicx}
\usepackage{multirow}
\usepackage{tabularx}
\usepackage[dvipsnames]{xcolor}
\usepackage{booktabs}
\usepackage{colortbl}
\usepackage{amsfonts}
\usepackage{amssymb}
\usepackage{bbding}
\usepackage{xcolor}
\usepackage{adjustbox}
\usepackage{makecell}
\usepackage[accsupp]{axessibility} 

\usepackage[pagebackref,breaklinks,colorlinks]{hyperref}

\usepackage[capitalize]{cleveref}
\crefname{section}{Sec.}{Secs.}
\Crefname{section}{Section}{Sections}
\Crefname{table}{Table}{Tables}
\crefname{table}{Tab.}{Tabs.}

\def\wacvPaperID{251} 
\def\confName{WACV}
\def\confYear{2025}

\begin{document}

\title{Sigma\raisebox{-0.19em}{\includegraphics[height=1.1em]{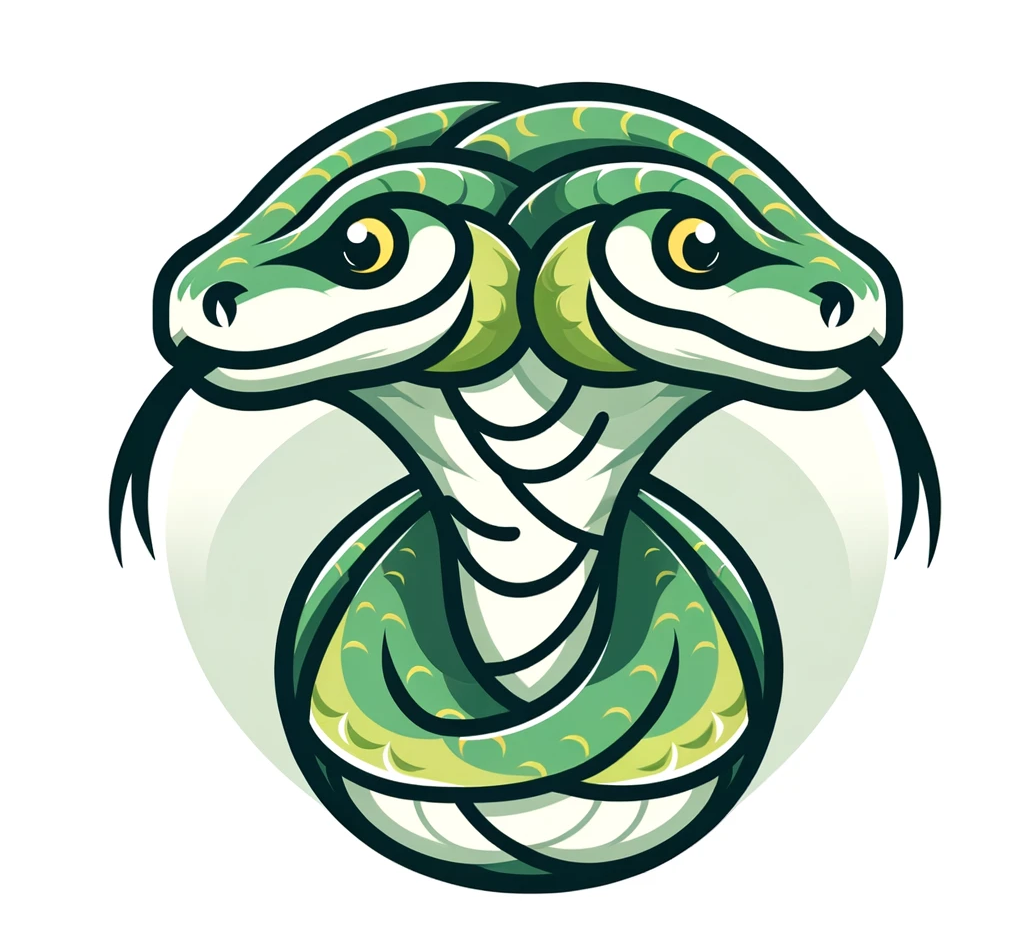}}: Siamese Mamba Network for Multi-Modal Semantic Segmentation} 

\author{Zifu Wan \quad
Pingping Zhang \quad
Yuhao Wang \quad
Silong Yong  \\ \vspace{1mm}
Simon Stepputtis \quad
Katia Sycara \quad
Yaqi Xie \\ 
Robotics Institute, Carnegie Mellon University 
\\ 
}
\maketitle

\begin{abstract}
\looseness = -1
Multi-modal semantic segmentation significantly enhances AI agents' perception and scene understanding, especially under adverse conditions like low-light or overexposed environments. Leveraging additional modalities (\textit{X-modality}) like thermal and depth alongside traditional RGB provides complementary information, enabling more robust and reliable prediction. 
In this work, we introduce Sigma, a Siamese Mamba network for multi-modal semantic segmentation utilizing the advanced Mamba. Unlike conventional methods that rely on CNNs, with their limited local receptive fields, or Vision Transformers (ViTs), which offer global receptive fields at the cost of quadratic complexity, our model achieves global receptive fields with linear complexity. By employing a Siamese encoder and innovating a Mamba-based fusion mechanism, we effectively select essential information from different modalities. A decoder is then developed to enhance the channel-wise modeling ability of the model. Our proposed method is rigorously evaluated on both RGB-Thermal and RGB-Depth semantic segmentation tasks, demonstrating its superiority and marking the first successful application of State Space Models (SSMs) in multi-modal perception tasks.
Code is available at \url{https://github.com/zifuwan/Sigma}.
\end{abstract}

\section{Introduction}
\label{sec:intro}

\looseness = -1
Semantic segmentation, aiming to assign a semantic label for each pixel within an image, has been increasingly significant for AI agents to accurately perceive their environment~\cite{guo2018review, chen20242023}.
However, current vision models still struggle in challenging conditions like low light or obstructive elements such as sun glare and fire~\cite{zhang2023robust,zhang2024hikersgg, chen20242023}. 
With the goal of enhancing segmentation under such challenging conditions, additional modalities like thermal and depth are utilized to enhance vision system robustness. 
With the supplementary information, the robustness and capabilities of vision pipelines can be improved~\cite{zhang2023delivering, liu2022cmx, lv2024context}.
However, utilizing multiple modalities introduces additional challenges, namely the alignment and fusion of the information provided through these additional channels~\cite{feng2020deep}.


Previous approaches in multi-modal semantic segmentation rely on Convolutional Neural Networks (CNN) or Vision Transformers (ViT). 
While CNN-based approaches~\cite{feng2023cekd, liang2023explicit,zhang2025spectral} are known for their scalability and linear complexity, they suffer from the small receptive field limited by the kernel size, leading to the local reductive bias. 
Besides, CNNs utilize a weight-sharing mechanism across different parts of the input, limiting its flexibility in adapting to unseen or low-quality images. 
In contrast, ViT-based methods~\cite{vaswani2017attention, liu2022cmx, bachmann2022multimae} offer enhanced visual modeling by leveraging global receptive fields and dynamic weights. 
However, their self-attention mechanism leads to quadratic complexity in terms of input sizes~\cite{gu2023mamba}, raising efficiency concerns. 
Attempts to improve this efficiency by reducing the dimensions or strides of the processing windows compromise the extent of the receptive fields~\cite{yang2023swin3d}.


\begin{figure}[t]
    \centering
        \centering
        \includegraphics[width=0.95\linewidth]{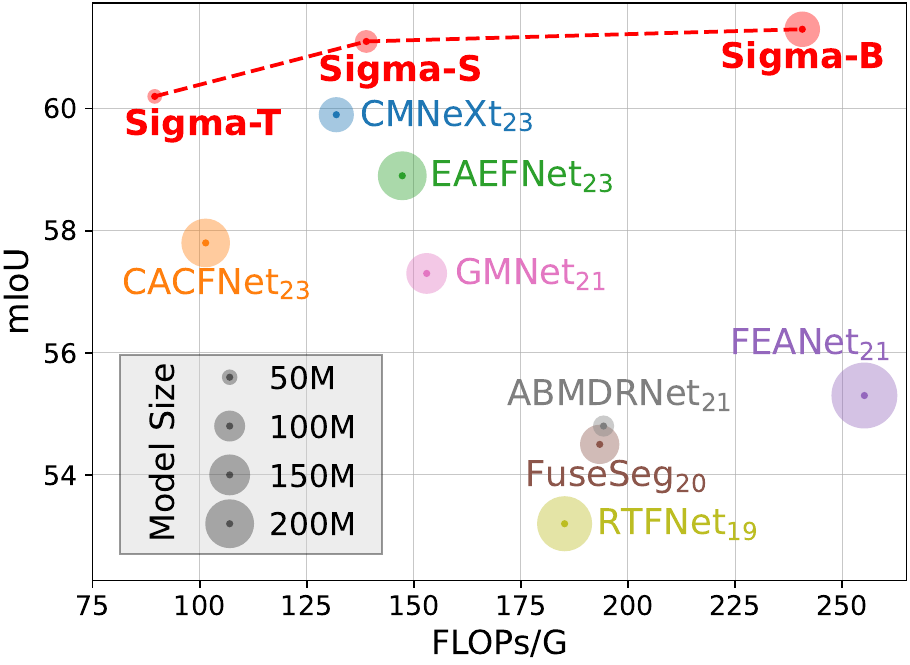}
        \label{fig:overall_flops}
    \vspace{-6pt}
    \caption{Computation and model size comparison of Sigma and other methods on MFNet dataset~\cite{ha2017mfnet}. The size of each circle indicates the model size (parameters).
    }
    \vspace{-12pt}
    \label{fig:teaser}
\end{figure}

To address the aforementioned limitations, Selective Structured State Space Models (S6), or Mamba~\cite{gu2023mamba} have gained more popularity due to their global receptive field coverage and dynamic weights with linear complexity.
Mamba has shown outstanding effectiveness in tasks involving long sequence modeling, notably in natural language processing~\cite{gu2023mamba}. Furthermore, more studies have explored its potential for vision-related applications, such as image classification~\cite{liu2024vmamba}, medical image segmentation~\cite{ma2024u,ruan2024vm,xing2024segmamba,wang2024weakmambaunet,yang2024vivim}, and 3D scene comprehension~\cite{liang2024pointmamba}. 
Inspired by these benefits, we introduce \textbf{Sigma}\raisebox{-0.19em}{\includegraphics[height=1.2em]{figs/snake.png}}, a Siamese Mamba network for multi-modal information fusion and apply it to the challenging domain of semantic segmentation.

Our Sigma integrates a Siamese encoder~\cite{yan2022fully, yan2023transy} for modality-specific feature extraction, fusion modules for aggregating information from different modalities, and a decoder for enhancing both spatial and channel-specific information.
The encoder backbone utilizes cascaded \textit{Visual State Space Blocks (VSSB)} to extract multi-scale information from various modalities.
Subsequently, the extracted features are inputted to a fusion module at each level, where multi-modal features initially interact through a \textit{Cross Mamba Block (CroMB)} to augment cross-modal information.
Then, the enhanced features are processed by a \textit{Concat Mamba Block (ConMB)}, which employs an attention mechanism to select pertinent information from each modality.
Finally, the fused features are sent to multi-level \textit{Channel-Aware Visual State Space Blocks (CAVSSB)} to effectively capture multi-scale long-range information. 

\looseness=-1
Comprehensive experiments on both RGB-Thermal~\cite{ha2017mfnet, shivakumar2020pst900} and RGB-Depth datasets~\cite{silberman2012indoor, song2015sun} demonstrate that Sigma outperforms state-of-the-art models in both accuracy and efficiency, as depicted in Fig.~\ref{fig:teaser}. 

\bigbreak
Our contributions can be summarized as follows:
\begin{itemize}
\item To our best knowledge, our work marks the first successful application of state space models, specifically Mamba, in multi-modal semantic segmentation.
\item We introduce a Mamba-based fusion mechanism alongside a channel-aware decoder, to efficiently extract information across different modalities and integrate them seamlessly.
\looseness=-1
\item Comprehensive evaluations in RGB-Thermal and RGB-Depth domains showcase our method's superior accuracy and efficiency, setting a new benchmark for future investigations into Mamba's potential in multi-modal learning.
\end{itemize}

\section{Related Work}
\label{sec: related work}
\begin{figure*}[t]
\vspace{-10pt}
\begin{center}
    \includegraphics[width=1\linewidth]{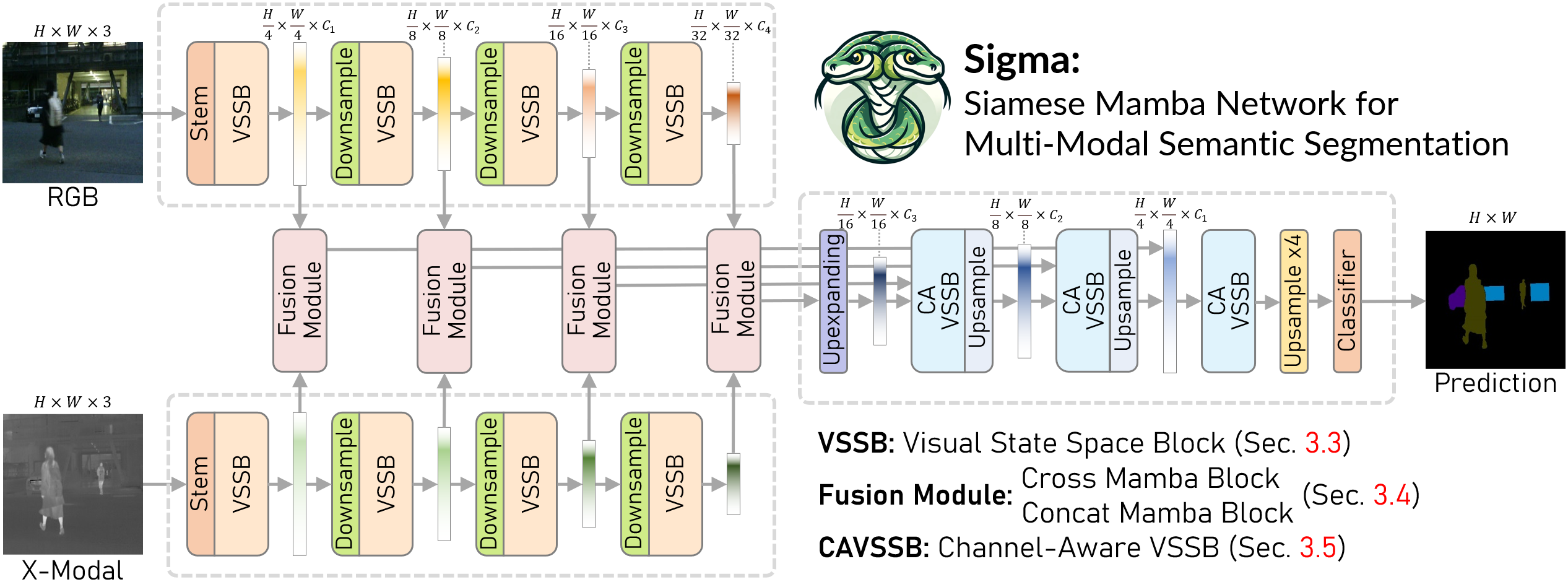}
\end{center}
\vspace{-6pt}
   \caption{Overall architecture of the proposed Sigma. }
\label{fig:overall}
\vspace{-6pt}
\end{figure*}
\textbf{Multi-Modal Semantic Segmentation.}
Multi-modal semantic segmentation typically integrates the RGB modality with other complementary modalities such as thermal, depth, and LiDAR~\cite{feng2020deep, zhang2021deep}. These supplementary modalities provide crucial information to the vision system across various scenarios. 
To optimize the utilization of these additional modalities, the development of effective feature extractors and fusion mechanisms has been of paramount importance.

In RGB-Thermal semantic segmentation, early approaches commonly employ encoder-decoder architectures with shortcut connections~\cite{ha2017mfnet, shivakumar2020pst900, sun2020fuseseg}, dense connections~\cite{sun2020fuseseg, xu2020u2fusion, zhou2021gmnet}, and dilated convolutions~\cite{xu2021attention, zhou2023cacfnet}. To address the limited context understanding of CNNs, many methods apply attention mechanisms during feature fusion~\cite{liang2023explicit, sun2020fuseseg, zhou2021gmnet}. With the rise of Transformers, methods like CMX~\cite{liu2022cmx} utilize SegFormer~\cite{xie2021segformer} for feature extraction and incorporate cross-attention modules for feature fusion. Building on CMX, CMNeXt~\cite{zhang2023delivering} introduces a self-query hub to select informative features from various modalities. SegMiF~\cite{liu2023multi} employs a cascading structure with a hierarchical interactive attention mechanism to ensure precise mapping across modalities.

In RGB-Depth semantic segmentation, approaches effective in RGB-Thermal semantic segmentation, such as CMX~\cite{liu2022cmx} and CMNeXt~\cite{zhang2023delivering}, also show impressive performance. Additionally, self-supervised pre-training is being explored in RGB-Depth perception. MultiMAE~\cite{bachmann2022multimae} uses a masked autoencoder with pseudo-labeling, while DFormer~\cite{yindformer} integrates RGB and Depth modalities within the pre-training architecture to learn transferable multi-modal representations.
Although the aforementioned Transformer-based methods have shown promising results in RGB-X semantic segmentation, the quadratic scaling nature of the self-attention mechanism limits the length of input sequences. 
%
In contrast, our proposed Sigma processes very long sequences, preserving all valuable information while requiring significantly less computation.

\textbf{State Space Models.}
State Space Models (SSM)~\cite{gu2021efficiently,smith2022simplified}, inspired by linear time-invariant systems, are considered as efficient sequence-to-sequence models.
Recently, the Structured State-Space Sequence model (S4)~\cite{gu2021efficiently} has emerged as a pioneering work in capturing long-range dependencies.
Furthermore, with the selective mechanism introduced into S4, Mamba~\cite{gu2023mamba} surpasses Transformers and other advanced architectures.
%
%
Due to the remarkable performance, researchers have extended it to the field of computer vision.
%
%
For example, Vision Mamba~\cite{zhu2024vision} integrates SSM with bidirectional scanning, making each patch related to another.
VMamba~\cite{liu2024vmamba} extends scanning in four directions to fully capture interrelations among image patches.
%
%
Besides, state space models have been extended to medical image segmentation~\cite{ma2024u,ruan2024vm,xing2024segmamba,wang2024weakmambaunet}, image restoration~\cite{guo2024mambair} and point cloud analysis~\cite{liang2024pointmamba}, all showing competitive results with lower complexity.
However, these recent works directly employ SSM as a plug-and-play module, without the in-depth design tailored to specific tasks.
Additionally, there is a lack of exploration of SSM in multi-modal tasks.
%
Thus, we propose a Mamba-based fusion mechanism for cross-modality exchange and concatenated feature enhancement, along with a channel-aware decoder for channel information selection.
%
By leveraging the specialized design of SSM for multi-modal tasks, our approach attains enhanced accuracy while maintaining low complexity.

\section{Sigma: Siamese Mamba Network}
\label{sec:Method}

In this section, we give a detailed illustration of our proposed \textbf{Siamese Mamba Network (Sigma)} for multi-modal semantic segmentation.
Firstly, we provide basic information of State Space Models.
Subsequently, we present an overview of our Sigma architecture, followed by in-depth discussions of the encoder, fusion module, and decoder.

\subsection{Preliminaries}
\textbf{State Space Models.}
\label{sec:preliminaries}
State Space Models (SSM)~\cite{gu2021efficiently,gu2021combining,smith2022simplified} represent a class of sequence-to-sequence modeling systems characterized by constant dynamics over time, a property also known as linear time-invariant.
With linear complexity, SSM can effectively capture the inherent dynamics through an implicit mapping to latent states, which can be defined as:
\begin{equation}
\small
    y(t) = Ch(t) + Dx(t), 
    \dot{h}(t) = Ah(t) + Bx(t).
\end{equation}
Here, $x(t) \in \mathbb{R},h(t) \in \mathbb{R}^{N}$, \text{and} $ y(t) \in \mathbb{R}$ denotes the input, hidden state, and the output, respectively.
$N$ is the state size and $\dot{h}(t)$ refers to the time derivative of $h(t)$.
Additionally, ${A} \in \mathbb{R}^{N \times N}$, ${B} \in \mathbb{R}^{N \times 1}$, ${C} \in \mathbb{R}^{1 \times N}$, and ${D} \in \mathbb{R}$ are the system matrices.
To process discrete sequences like image and text, SSMs adopt Zero-Order Hold (ZOH) discretization~\cite{gu2021efficiently} to map the input sequence $\{x_1, x_2, ..., x_K\}$ to the output sequence $\{y_1, y_2, ..., y_K\}$.
Specifically, suppose $\mathrm{\Delta} \in \mathbb{R}^{D}$ is the pre-defined timescale parameter to map continuous parameters $A$, $B$ into a discrete space, the discretization process can be formulated as:
\begingroup
\begin{gather}
\overline{A}=\exp (\mathrm{\Delta} A),\; \overline{B} = (\mathrm{\Delta} A)^{-1}(\exp (A) - I)\mathrm{\Delta} B, \;\overline{C} = C, \\
  y_k = \overline{C}{h_k} + \overline{D}{x_k},
  h_{k} = \overline{A}{h_{k-1}} +\overline{B}{x_k}.
  \label{eq:ss}
\end{gather}
\endgroup
Here, all the matrices keep the same dimension as the operation iterates.
Notably, $\overline{D}$, serving as a residual connection, is often discarded in the equation:
\begingroup
\begin{equation}
    y_k =  \overline{C}{h_k}.
    \label{eq:y=ch}
\end{equation}
\endgroup
Besides, following Mamba~\cite{gu2023mamba}, the matrix $\overline{B}$ can be approximated by the first-order Taylor series:
\begingroup
\begin{equation}
    \overline{B} =(\exp (A) - I) A^{-1} B \approx (\mathrm{\Delta} A)(\mathrm{\Delta} A)^{-1}\mathrm{\Delta}B = \mathrm{\Delta}B.
\end{equation}
\endgroup

\textbf{Selective Scan Mechanism.}
While SSMs are effective for modeling discrete sequences, they encounter invariant parameters regardless of differences in the input.
To address this limitation, the Selective State Space Model (S6, \textit{a.k.a} Mamba)~\cite{gu2023mamba} is introduced, ensuring the input-dependency.
In Mamba, the matrices $B \in \mathbb{R}^{L \times N}$, $C \in \mathbb{R}^{L \times N}$, and $\mathrm{\Delta} \in \mathbb{R}^{L \times D}$ are derived from the input $x \in \mathbb{R}^{L \times D}$, enabling the model to be contextually aware of the input.
With this selection mechanism, Mamba is capable of effectively modeling the complex interactions present in long sequences.

\subsection{Overall Architecture}
\looseness = -1
As illustrated in Fig.~\ref{fig:overall}, our proposed method comprises a siamese mamba encoder (Sec.~\ref{sec:method encoder}), a fusion module (Sec.~\ref{sec:method fusion}), and a channel-aware mamba decoder (Sec.~\ref{sec:method decoder}).
During the encoding phase, four Visual State Space Blocks (VSSB) with downsampling operations are sequentially cascaded to extract multi-level image features. 
The two encoder branches share weights to reduce computational complexity.
Subsequently, features from each level, derived from the two branches, are processed through a fusion module. 
%
%
In the decoding phase, the fused features at each level are further enhanced by a Channel-Aware Visual State Space Block (CAVSSB) with an upsampling operation. 
Ultimately, the final feature is forwarded to a classifier to generate the prediction.

\subsection{Siamese Mamba Encoder}
\label{sec:method encoder}
\begin{figure}[h]
\begin{center}
    \includegraphics[width=0.8\linewidth]{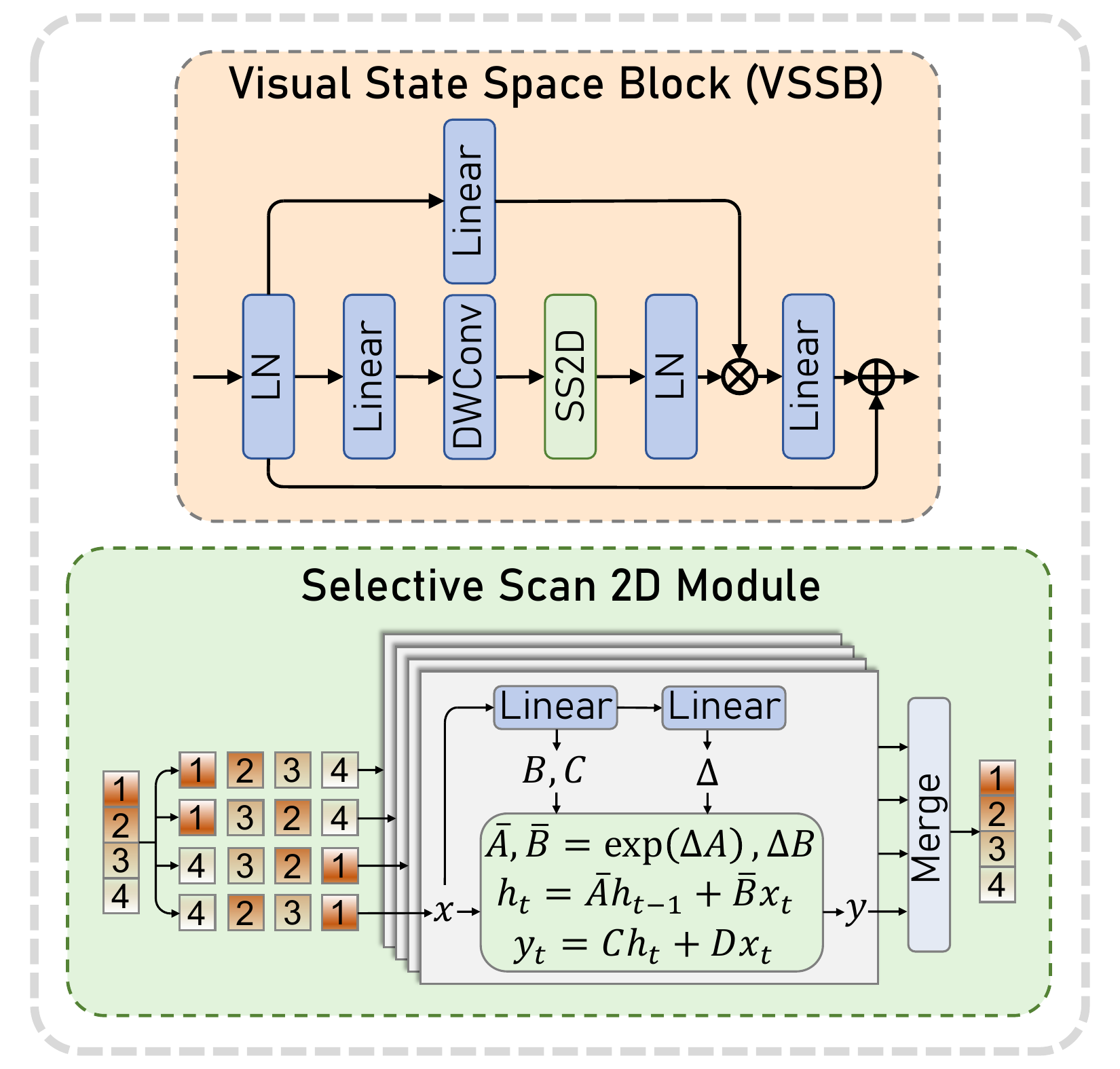}
\end{center}
\vspace{-9pt}
   \caption{\looseness=-1 The Visual State Space Block (VSSB) used in the encoder.}
\label{fig:encoder structure}
\vspace{-1pt}
\end{figure}

\looseness = -1
The encoder has two weight-sharing branches which take an RGB image and an X-modality image as input, denoted as $I_{\mathsf{rgb}}, I_{\mathsf{x}} \in \mathbb{R}^{H \times W \times 3}$. Here, $H$ and $W$ represent the height and width of the input modalities.
The encoder starts with a stem module similar to ViT~\cite{dosovitskiy2020image} which partitions the input into patches, generating feature maps $\mathcal{F}_i^1 \in \mathbb{R}^{\frac{H}{4}\times \frac{W}{4}\times C_1} $, where $i \in \{\mathsf{rgb}, \mathsf{x}\}$ refers to the RGB or X modality.
%
%
The features are continually processed by three sets of downsampling and VSSB, yielding multi-scale features $\{\mathcal{F}_i^2 \in \mathbb{R}^{\frac{H}{8}\times \frac{W}{8}\times C_2}, \mathcal{F}_i^3 \in \mathbb{R}^{\frac{H}{16}\times \frac{W}{16}\times C_3}, \mathcal{F}_i^4 \in \mathbb{R}^{\frac{H}{32}\times \frac{W}{32}\times C_4}\}$.
The details of VSSB are introduced as follows.

\textbf{Visual State Space Block (VSSB).}
Following VMamba~\cite{liu2024vmamba}, we implement the VSSB with Selective Scan 2D (SS2D) modules.
As shown in Fig.~\ref{fig:encoder structure}, the input feature is processed by a sequence of linear projection ($\texttt{Linear}$), Depth-wise Convolution ($\texttt{DWConv}$). Then, an SS2D module with a residual connection is used to model long-range spatial information.

\textbf{SS2D Module.}
Within the SS2D module, the input feature is first flattened to four sequences from four directions (top-left to bottom-right, bottom-right to top-left, top-right to bottom-left, and bottom-left to top-right).
Then four distinctive selective scan modules~\cite{gu2023mamba} are used to extract multi-direction information, where each of them captures the long-range dependencies of the sequence with the operation in Eq.~\ref{eq:ss}.
Lastly, the four sequences are reversed to the same direction and summed.

\subsection{Fusion Module}
\label{sec:method fusion}
\begin{figure}[t]
\vspace{-7pt}
\begin{center}
    \includegraphics[width=\linewidth]{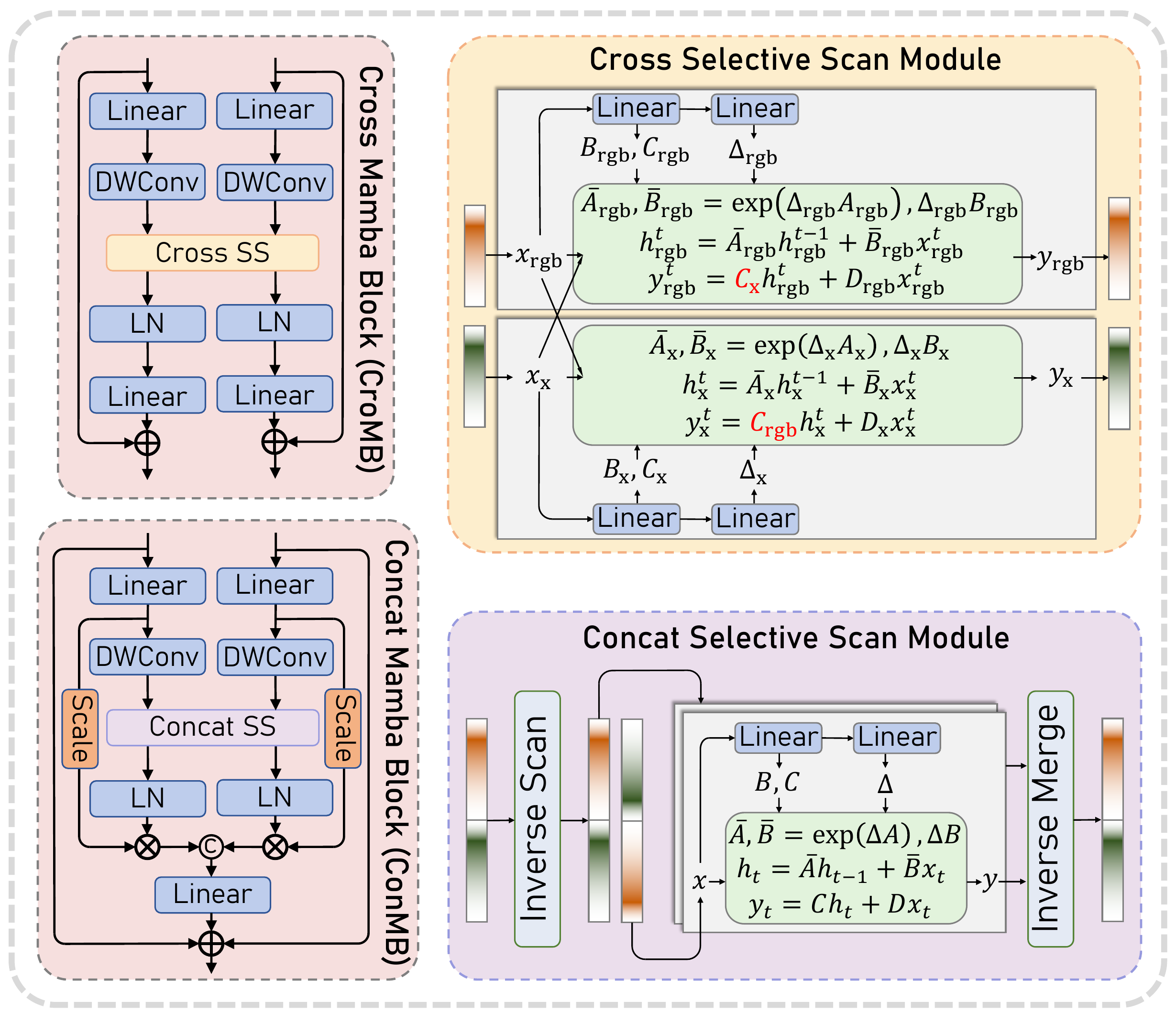}
\end{center}
\vspace{-10pt}
   \caption{The left part shows the Cross Mamba Block (CroMB) and Concat Mamba Block (ConMB). The right part shows the Cross Selective Scan (Cross SS) module and Concat Selective Scan (Concat SS) module.}
\label{fig:fusion}
\vspace{-2pt}
\end{figure}

The detailed architecture of the feature fusion module is illustrated in Fig.~\ref{fig:fusion}. It consists of a Cross Mamba Block (CroMB) and a Concat Mamba Block (ConMB). 
Specifically, CroMB employs a cross-multiplication mechanism to enhance the features with one another, and ConMB applies the selective scan mechanism to concatenated features to obtain the fusion result.
Suppose the features from the $k$-th Siamese encoder block are represented as $\mathcal{F}_{\mathsf{rgb/x}}^k \in \mathbb{R}^{H_k\times W_k\times C_k}$, then the entire fusion process can be represented as:
\begingroup
\begin{gather}
    \hat{\mathcal{F}}_{\mathsf{rgb}}^k,\; \hat{\mathcal{F}}_{\mathsf{x}}^k = \texttt{CroMB}(\mathcal{F}_{\mathsf{rgb}}^k, \; \mathcal{F}_{\mathsf{x}}^k), \\
    \mathcal{F}_{\mathsf{fuse}}^k = \texttt{ConMB}(\hat{\mathcal{F}}_{\mathsf{rgb}}^k,\; \hat{\mathcal{F}}_{\mathsf{x}}^k).
\end{gather}
\endgroup
Here, $\hat{\mathcal{F}}_{\mathsf{rgb}}^k, \hat{\mathcal{F}}_{\mathsf{x}}^k, \text{and }\mathcal{F}_{\mathsf{fuse}}^k$ remains the original dimension as $\mathbb{R}^{H_k\times W_k\times C_k}$.
Details of CroMB and ConMB are as follows.

\label{sec:method fusion}

\textbf{CroMB.}
As demonstrated in the upper part of Fig.~\ref{fig:fusion}, two input features are first processed by linear layers and depth-wise convolutions, then sent to the Cross Selective Scan (Cross SS) module.
According to the selection mechanism of Mamba mentioned in Sec.~\ref{sec:preliminaries}, the system matrices $B,\; C$ and $\mathrm{\Delta}$ are generated to enable the context-aware ability of the model.
Here, linear projection layers are utilized to generate the matrices.
According to Eq.~\ref{eq:y=ch}, matrix $C$ is used to decode the information from the hidden state $h_k$ to obtain the output $y_k$. 
Inspired by the cross-attention mechanism~\cite{chen2021crossvit}, we facilitate information exchange among multiple selective scan modules.
%
%
In particular, the process can be represented as:
\begingroup
\begin{gather}
\overline{A}_{\mathsf{rgb}} = \exp (\mathrm{\Delta_{\mathsf{rgb}}} A_{\mathsf{rgb}}), \ \overline{A}_{\mathsf{x}} = \exp (\mathrm{\Delta_{\mathsf{x}}} A_{\mathsf{x}}), \\
\overline{B}_{\mathsf{rgb}} = \mathrm{\Delta_{rgb}} B_{\mathsf{rgb}},\ \overline{B}_{\mathsf{x}} = \mathrm{\Delta_{x}} B_{\mathsf{x}}, \\
  h_{\mathsf{rgb}}^{t} = \overline{A}_{\mathsf{rgb}}{h_{\mathsf{rgb}}^{t-1}} +\overline{B}_{\mathsf{rgb}}{x_{\mathsf{rgb}}^t},\ h_{\mathsf{x}}^{t} = \overline{A}_{\mathsf{x}}{h_{\mathsf{x}}^{t-1}} +\overline{B}_{\mathsf{x}}{x_{\mathsf{x}}^t}, \label{Eq:hidden state}\\
  y_{\mathsf{rgb}}^t = \textcolor{red}{{C}_{\mathsf {x}}}{h_{\mathsf{rgb}}^t} + {D}_{\mathsf{rgb}}{x_{\mathsf{rgb}}^t},\ y_{\mathsf{x}}^t = \textcolor{red}{{C}_{\mathsf {rgb}}}{h_{\mathsf{x}}^t} + {D}_{\mathsf{x}}{x_{\mathsf{x}}^t}, \label{Eq:cross attention}\\
  y_{\mathsf{rgb}} = [y_{\mathsf{rgb}}^1, y_{\mathsf{rgb}}^2, \ldots, y_{\mathsf{rgb}}^l], \ y_{\mathsf{x}} = [y_{\mathsf{x}}^1, y_{\mathsf{x}}^2, \ldots, y_{\mathsf{x}}^l].
\end{gather}
\endgroup
\looseness = -1
Here, $x_{\mathsf{rgb/x}}^t$ represents the input at time step $t$, and $y_{\mathsf{rgb/x}}$ denotes the selective scan output.
\(\textcolor{red}{{C}_{\mathsf {x}}}\) and \(\textcolor{red}{{C}_{\mathsf {rgb}}}\) are the cross-modal matrices used for recovering the outputs at each time step from the hidden states.

\textbf{ConMB.}
In CroMB, the features from two modalities interact with each other and obtain the cross-modal-enhanced features.
To further obtain a fused feature containing vital information from both modalities, we propose ConMB to integrate the outputs from CroMB.
%
%

\looseness = -1
Specifically, the outputs $\hat{\mathcal{F}}_{\mathsf{rgb}}^k, \hat{\mathcal{F}}_{\mathsf{x}}^k \in \mathbb{R}^{H_k\times W_k\times C_k}$ from CroMB are first processed by linear and depth-wise convolution layers, then sent to the Concat Selective Scan (Concat SS) module.
Within the Concat SS module, the two features are first flattened and then concatenated in the sequence length dimension.
This provides a sequence $\mathcal{S}_{\mathsf{Concat}}^k$ of shape $\mathbb{R}^{(2\times H_k\times W_k)\times C_k}$.
Besides, to comprehensively capture information from two modalities, we inversely scan the concatenated sequence $\mathcal{S}_{\mathsf{Concat}}^k$ to obtain an additional sequence $\mathcal{S}_{\mathsf{Inverse}}^k \in \mathbb{R}^{(2\times H_k\times W_k)\times C_k}$.
Afterwards, each sequence is processed to capture long-range dependencies, obtaining $\hat{\mathcal{S}}_{\mathsf{Concat}}^k$ and $\hat{\mathcal{S}}_{\mathsf{Inverse}}^k$.
Then the inversed sequence output is flipped back and added with the processed sequence.
The summed sequence is separated to recover two outputs.
This process can be represented as:
\begingroup
\begin{gather}
    \widetilde{\mathcal{F}}_{\mathsf{rgb}}^k = \texttt{DWConv}(\texttt{Linear}(\hat{\mathcal{F}}_{\mathsf{rgb}}^k)),\\ \widetilde{\mathcal{F}}_{\mathsf{x}}^k = \texttt{DWConv}(\texttt{Linear}(\hat{\mathcal{F}}_{\mathsf{x}}^k)),
    \\
    \mathcal{S}_{\mathsf{Concat}}^k = \texttt{Concat}(\widetilde{\mathcal{F}}_{\mathsf{rgb}}^k, \widetilde{\mathcal{F}}_{\mathsf{x}}^k),\\
    \mathcal{S}_{\mathsf{Inverse}}^k = \texttt{Inverse}(\mathcal{S}_{\mathsf{Concat}}^k),\\
    \hat{\mathcal{S}}_{\mathsf{Concat}}^k, \hat{\mathcal{S}}_{\mathsf{Inverse}}^k = \texttt{SSM}(\mathcal{S}_{\mathsf{Concat}}^k), \texttt{SSM}(\mathcal{S}_{\mathsf{Inverse}}^k), \\
    \overline{\mathcal{F}}_{\mathsf{RGB}}^k, \overline{\mathcal{F}}_{\mathsf{X}}^k = \texttt{Seperate}(\hat{\mathcal{S}}_{\mathsf{Concat}}^k + \texttt{Inverse}(\hat{\mathcal{S}}_{\mathsf{Inverse}}^k)).
\end{gather}
\endgroup
After obtaining the scanned features $\overline{\mathcal{F}}_{\mathsf{rgb}}^k$ and $\overline{\mathcal{F}}_{\mathsf{x}}^k$, they are multiplied with two scaling parameters and concatenated in the channel dimension.
Finally, a linear projection layer is used to reduce the feature shape to $\mathbb{R}^{H_k\times W_k\times C_k}$.

\subsection{Channel-Aware Mamba Decoder}
\label{sec:method decoder}
\begin{figure}[h]
\begin{center}
    \includegraphics[width=0.7\linewidth]{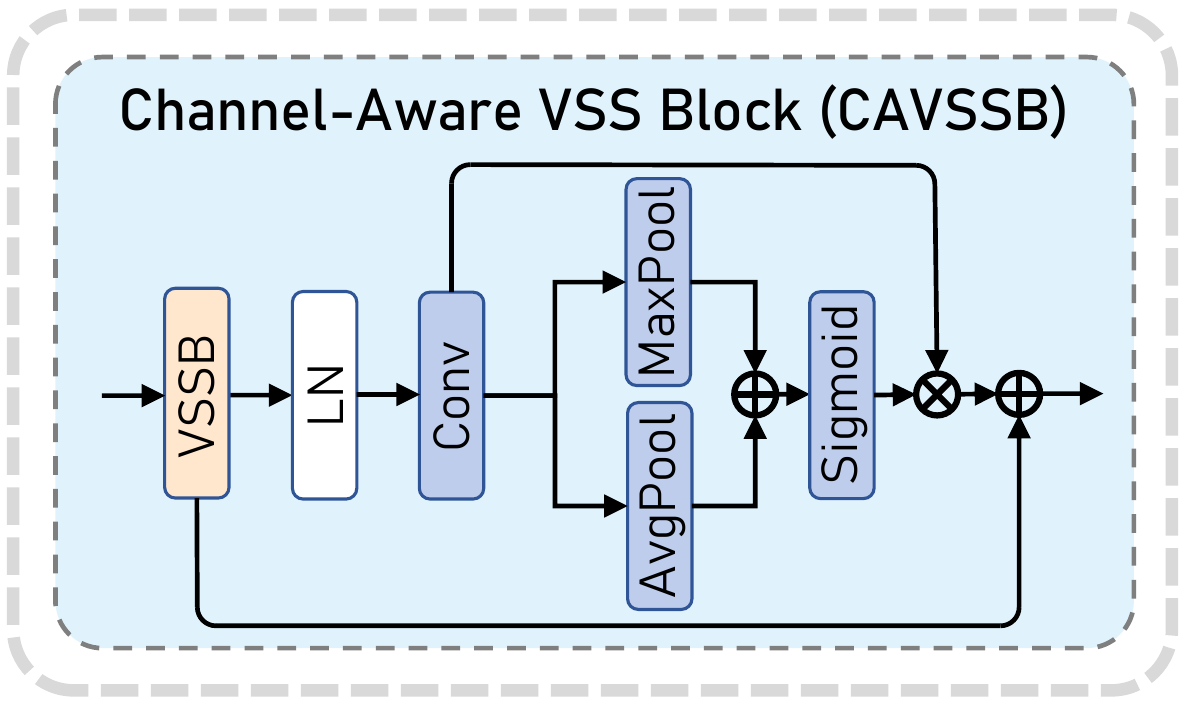}
\end{center}
\vspace{-15pt}
   \caption{The Channel-Aware Visual State Space Block (CAVSSB) used in the Sigma decoder.}
\label{fig:decoder structure}
\vspace{-6pt}
\end{figure}
To obtain the final prediction, we introduce a channel-aware Mamba decoder, as shown in Figure~\ref{fig:overall}. The decoder consists of three groups of Channel-Aware Visual State Space Blocks (CAVSSB) and upsampling operations. Finally, the decoded features are classified by MLP layers for semantic segmentation prediction. Details of the CAVSSB are shown in Figure~\ref{fig:decoder structure}.

The CAVSSB incorporates a VSSB to process the fused features. While VSSBs are adept at extracting global spatial context, they fall short in learning inter-channel information. To address this issue, we apply channel attentions to select vital channel-wise representations. Specifically, our channel-attention operation consists of the average pooling and max pooling. This straightforward implementation offers effective channel selection while maintaining efficiency. Additionally, two residual connections are used to acquire the final output. In this manner, we form a spatial-channel-aware scheme for robust decoding.
\section{Experiments}

\begin{table*}[h]
\begin{center}

\subfloat[Quantitative comparisons on MFNet day-night evaluation set~\cite{ha2017mfnet} (9 classes)] 
{
    \resizebox{\linewidth}{!}
    {
        \footnotesize
        \begin{tabular}{c|c|cc|p{0.8cm}<{\centering}p{0.8cm}<{\centering}p{0.8cm}<{\centering}p{0.8cm}<{\centering}p{0.8cm}<{\centering}p{0.8cm}<{\centering}p{0.8cm}<{\centering}p{0.8cm}<{\centering}p{0.8cm}<{\centering}|p{0.7cm}<{\centering}}
        \toprule
        Method & Backbone & Params (M) & FLOPs (G) & Unlabeled & Car & Person & Bike & Curve & Stop & Guardrail & Cone & Bump & mIoU\\ 
        \midrule \midrule
        MFNet$_{\mathsf{17}}$~\cite{ha2017mfnet} & -- & -- & -- & 96.9 & 65.9 & 58.9 & 42.9 & 29.9 & 9.9 & 0.0 & 25.2 & 27.7 & 39.7 \\
        RTFNet$_{\mathsf{19}}$~\cite{sun2019rtfnet} & ResNet-152 & 245.7 & 185.2 & \underline{98.5} & 87.4 & 70.3 & 62.7 & 45.3 & 29.8 & 0.0 & 29.1 & 55.7 & 53.2 \\
        PSTNet$_{\mathsf{20}}$~\cite{shivakumar2020pst900} & ResNet-18 & 105.8 & 123.4 & 97.0 & 76.8 & 52.6 & 55.3 & 29.6 & 25.1 & \textbf{15.1} & 39.4 & 45.0 & 48.4 \\
        FuseSeg$_{\mathsf{20}}$~\cite{sun2020fuseseg} & DenseNet-161 & 141.5 & 193.4  & 97.6 & 87.9 & 71.7 & 64.6 & 44.8 & 22.7 & 6.4 & 46.9 & 47.9 & 54.5 \\
        U2Fusion$_{\mathsf{20}}$~\cite{xu2020u2fusion} & VGG-16 & -- & --  & 97.7 & 82.8 & 64.8 & 61.0 & 32.3 & 20.9 & -- & 45.2 & 50.2 & 50.8 \\
        AFNet$_{\mathsf{21}}$~\cite{xu2021attention} & ResNet-50 & -- & --  & 98.0 & 86.0 & 67.4 & 62.0 & 43.0 & 28.9 & 4.6 & 44.9 & 56.6 & 54.6 \\
        ABMDRNet$_{\mathsf{21}}$~\cite{zhang2021abmdrnet} & ResNet-50 & 64.6 & 194.3  & \textbf{98.6} & 84.8 & 69.6 & 60.3 & 45.1 & 33.1 & 5.1 & 47.4 & 50.0 & 54.8 \\
        FEANet$_{\mathsf{21}}$~\cite{deng2021feanet} & ResNet-152 & 337.1 & 255.2  & 98.3 & 87.8 & 71.1 & 61.1 & 46.5 & 22.1 & 6.6 & {55.3} & 48.9 & 55.3 \\
        GMNet$_{\mathsf{21}}$~\cite{zhou2021gmnet} & ResNet-50 & 149.8 & 153.0  & 97.5 & 86.5 & 73.1 & 61.7 & 44.0 & \underline{42.3} & 14.5 & 48.7 & 47.4 & 57.3 \\
        TarDAL$_{\mathsf{22}}$~\cite{liu2022target} & -- & 297 & --  & 97.6 & 80.7 & 67.1 & 60.1 & 34.9 & 10.5 & -- & 38.7 & 45.5 & 48.6 \\
        CMX$_{\mathsf{22}}$~\cite{liu2022cmx} & MiT-B4 & 139.9 & 134.3  & 98.3 & {90.1} & {75.2} & 64.5 & {50.2} & 35.3 & 8.5 & 54.2 & \underline{60.6} & 59.7 \\
        EAEFNet$_{\mathsf{23}}$~\cite{liang2023explicit} & ResNet-152 & 200.4 & 147.3  & -- & 87.6 & 72.6 & 63.8 & 48.6 & 35.0 & \underline{14.2} & 52.4 & 58.3 & 58.9 \\
        CACFNet$_{\mathsf{23}}$~\cite{zhou2023cacfnet} & ConvNeXt-B & 198.6 & 101.4  & -- & 89.2 & 69.5 & 63.3 & 46.6 & 32.4 & 7.9 & 54.9 & 58.3 & 57.8 \\
        PAIF$_{\mathsf{23}}$~\cite{liu2023paif} & -- & 260 & --  & -- & 88.1 & 72.4 & 60.8 & -- & -- & -- & \underline{56.0} & 57.2 & 56.5 \\
        CENet$_{\mathsf{23}}$~\cite{feng2023cekd} & ResNet-50 & -- & --  & -- & 85.8 & 70.0 & 61.4 & 46.8 & 29.3 & 8.7 & 47.8 & 56.9 & 56.1 \\
        SegMiF$_{\mathsf{23}}$~\cite{liu2023multi} & MiT-B3 & -- & --  & 98.1 & 87.8 & 71.4 & 63.2 & 47.5 & 31.1 & -- & 48.9 & 50.3 & 56.1 \\
        CMNeXt$_{\mathsf{23}}$~\cite{zhang2023delivering} & MiT-B4 & 119.6 & 131.9  & 98.4 & \textbf{91.5} & \underline{75.3} & 67.6 & 50.5 & 40.1 & 9.3 & 53.4 & 52.8 & 59.9 \\
        \rowcolor{gray!20}
        Sigma (Ours) & VMamba-T & 48.3 & 89.5  & 98.4 & 90.8 & 75.2 & 66.6 & 48.2 & 38.0 & 8.7 & 55.9 & 60.4 & 60.2 \\
        \rowcolor{gray!20}
        Sigma (Ours) & VMamba-S & 69.8 & 138.9  & \underline{98.5} & \textbf{91.5} & \textbf{75.8} & 67.8 & 49.6 & 41.8  & 9.6 & 54.8 & 60.4 & \underline{61.1} \\
        \rowcolor{gray!20}
        Sigma (Ours) & VMamba-B & 121.4 & 240.7  & \underline{98.5} & \underline{91.1} & 75.2 & \underline{68.0} & \underline{50.8} & \textbf{43.0} & 9.7 & \textbf{57.6} & 57.9 & \textbf{61.3} \\
        \bottomrule
        \end{tabular}
        \label{tab:MF_benchmark}
    }
} 
\vspace{1pt}
\subfloat[Quantitative comparisons on PST900 evaluation set~\cite{shivakumar2020pst900} (5 classes)] 
{
    \resizebox{0.72\linewidth}{!}
    {
        \footnotesize
        \begin{tabular}{c|c|p{1.2cm}<{\centering}p{1.2cm}<{\centering}p{1.2cm}<{\centering}p{1.3cm}<{\centering}p{1.2cm}<{\centering}|p{1.2cm}<{\centering}}
        \toprule
        Method & Backbone & Background & Extinguisher & Backpack & Hand-Drill & Survivor &  mIoU \\ 
        \midrule \midrule
        MFNet$_{\mathsf{17}}$~\cite{ha2017mfnet} & -- & 98.6 & 60.4 & 64.3 & 41.1 & 20.7 & 57.0 \\ 
        RTFNet$_{\mathsf{19}}$~\cite{sun2019rtfnet} & ResNet-152 & 98.9 & 52.0 & 75.3 & 25.4 & 36.4 & 57.6 \\ 
        PSTNet$_{\mathsf{20}}$~\cite{shivakumar2020pst900} & ResNet-18 & 98.9 & 70.1 & 69.2 & 53.6 & 50.0 & 68.4 \\ 
        ABMDRNet$_{\mathsf{21}}$~\cite{zhang2021abmdrnet} & ResNet-50 & 99.0 & 66.2 & 67.9 & 61.5 & 62.0 & 71.3 \\
        GMNet$_{\mathsf{21}}$~\cite{zhou2021gmnet} & ResNet-50 & 99.4 & 73.8 & 83.8 & 85.2 & 78.4 & 84.1 \\ 
        CCFFNet$_{\mathsf{22}}$~\cite{wu2022complementarity} & ResNet-101 & 99.4 & \textbf{82.8} & 75.8 & 79.9 & 72.7 & 82.1 \\ 
        EGFNet$_{\mathsf{23}}$~\cite{dong2023egfnet} & ResNet-101 & \textbf{99.6} & 80.0 & \textbf{90.6} & 76.1 & 80.9 & 85.4 \\ 
        CACFNet$_{\mathsf{22}}$~\cite{zhou2023cacfnet} & ConvNeXt-B & \textbf{99.6} & \underline{82.1} & 89.5 & 80.9 & 80.8 & 86.6 \\ 
        
        CAINet$_{\mathsf{24}}$~\cite{lv2024context} & MobileNet-V2 & \underline{99.5} & 80.3 & 88.0 & 77.2 & 78.7 & 84.7 \\  
        \rowcolor{gray!20}
        Sigma (Ours) & VMamba-T & \textbf{99.6} & 81.9 & \underline{89.8} & \underline{88.7} & \textbf{82.7} & \textbf{88.6} \\ 
        \rowcolor{gray!20}
        Sigma (Ours) & VMamba-S & \textbf{99.6} & 79.4 & 88.7 & \textbf{90.2} & \underline{81.2} & \underline{87.8} \\ 
        \bottomrule
        \end{tabular}
        \label{tab:PST_benchmark}
    }
} 
\vspace{-0.1in}
\caption{Quantitative comparisons for semantic segmentation of RGB-T images on MFNet~\cite{ha2017mfnet} and PST900~\cite{shivakumar2020pst900} datasets. 
The best and second best results in each block are highlighted in \textbf{bold} and \underline{underline}, respectively.}
\label{tab:rgbt}
\vspace{-0.15in}
\end{center}
\end{table*}

\subsection{Experimental Settings}
\textbf{Datasets.}
\looseness=-1
To verify the effectiveness of Sigma, we conduct extensive experiments on two publicly available RGB-Thermal (RGB-T) semantic segmentation datasets, namely MFNet~\cite{ha2017mfnet} and PST900~\cite{shivakumar2020pst900}.
In addition, to better understand the generalizability of Sigma, we conduct experiments on two RGB-Depth (RGB-D) datasets, including NYU Depth V2~\cite{silberman2012indoor} and SUN RGB-D~\cite{song2015sun}.
The details of these datasets are as follows.
\begin{itemize}
    \item \textbf{MFNet dataset} contains 820 daytime and 749 nighttime RGB-T images with a resolution of $640\times480$. The dataset includes eight common classes of objects in driving scenarios. We follow the training/testing split of~\cite{liu2022cmx}.
    \item \textbf{PST900 dataset}
    provides 597 and 288 calibrated RGB-T images with a resolution of $1280\times720$ for training and validation. This dataset is collected from DARPA Subterranean Challenge and annotated with four classes.
    \item \textbf{NYU Depth V2 dataset} contains 1449 RGB-D images annotated with 40 semantic classes with a resolution of $640\times480$. We divide them into 795/654 for training/testing following previous work~\cite{hu2019acnet, girdhar2022omnivore}.
    \item \textbf{SUN RGB-D dataset} incorporates 10335 RGB-D images with 37 classes. We follow the common setting~\cite{hu2019acnet} to split 5285/5050 for training/testing, and reshape the images to $640\times480$.
\end{itemize}

\textbf{Evaluation Metrics.}
Following previous work~\cite{zhang2023delivering, hu2019acnet}, we report the Intersection over Union (mIoU) averaged across the semantic classes for evaluation. 

\textbf{Training Settings.}
We follow \cite{liu2022cmx} to use the AdamW optimizer~\cite{loshchilov2017decoupled} with an initial learning rate $6e^{-5}$ and weight decay $0.01$. 
The model is trained with a batch size of 8 for 500 epochs.
We utilize the ImageNet-1K~\cite{russakovsky2015imagenet} pre-trained model provided by VMamba~\cite{liu2024vmamba} for the Siamese image encoder, leading to three different sizes of models (Sigma-Tiny, Sigma-Small, and Sigma-Base).
%
%
%
More details of the experimental settings are described in the supplementary material.

\subsection{Quantitative and Qualitative Results}
\begin{figure*}[t]
	\centering
	\vspace{-2mm}
	\resizebox{1\textwidth}{!}
	{
		\begin{tabular}{@{}c@{}c@{}c@{}c@{}c@{}c@{}c@{}c@{}c@{}c@{}c}
			\vspace{-0.5mm}
			\includegraphics[width=1.85cm,height=1.85cm]{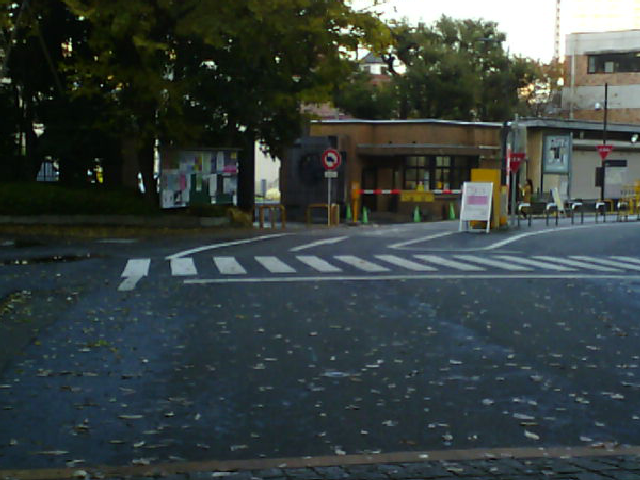}\hspace{0.5mm} &
			\includegraphics[width=1.85cm,height=1.85cm]{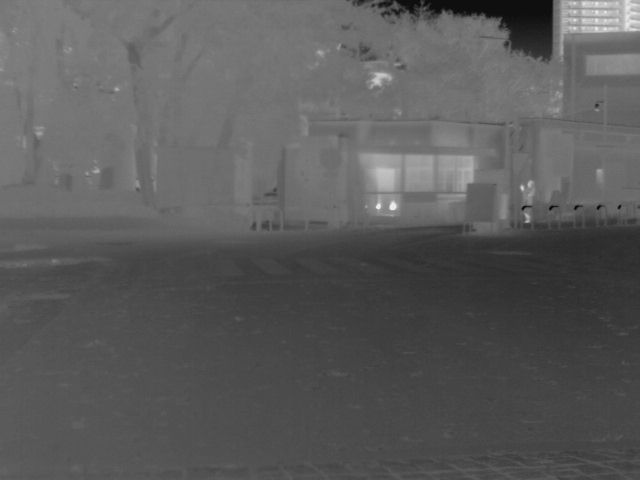}\hspace{0.5mm} &
			\includegraphics[width=1.85cm,height=1.85cm]{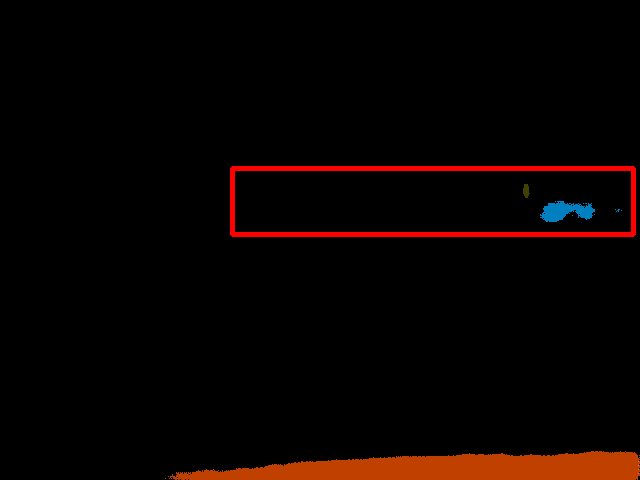}\hspace{0.5mm} &
			\includegraphics[width=1.85cm,height=1.85cm]{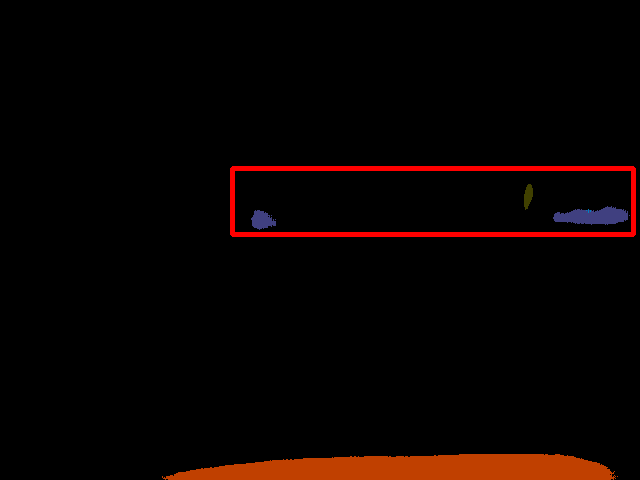}\hspace{0.5mm} &
			\includegraphics[width=1.85cm,height=1.85cm]{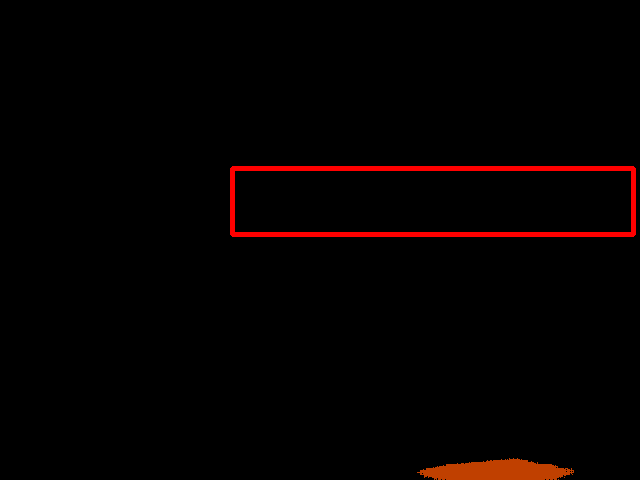}\hspace{0.5mm} &
                \includegraphics[width=1.85cm,height=1.85cm]{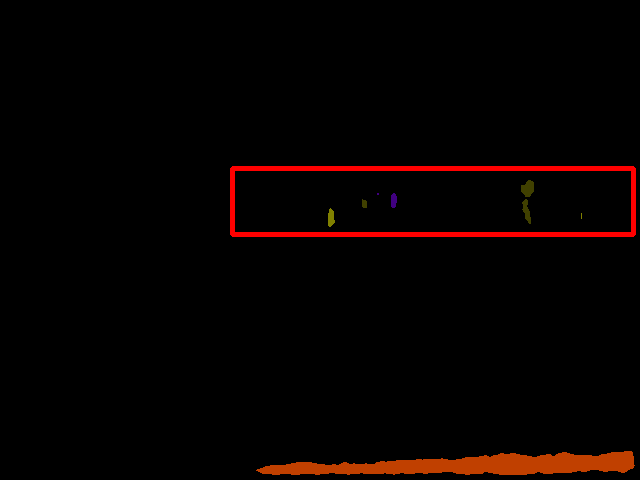}\hspace{0.5mm} &
                \includegraphics[width=1.85cm,height=1.85cm]{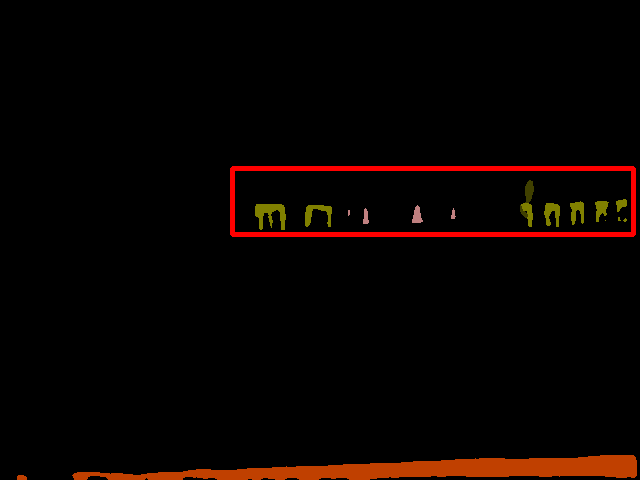}\hspace{0.5mm} &
                \includegraphics[width=1.85cm,height=1.85cm]{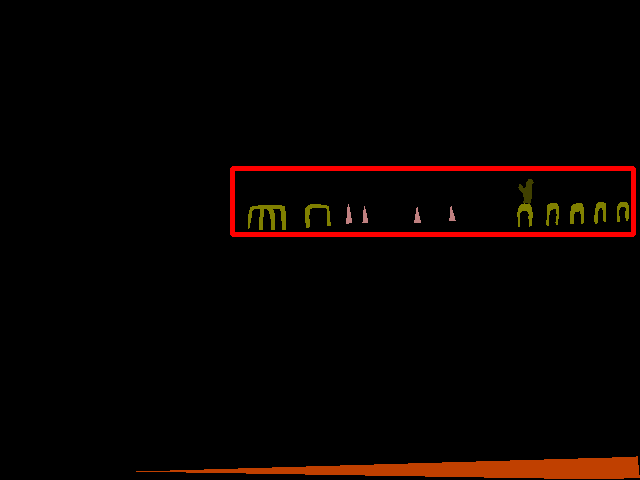}\hspace{0.5mm} \\
			\vspace{-0.5mm}
   \includegraphics[width=1.85cm,height=1.85cm]{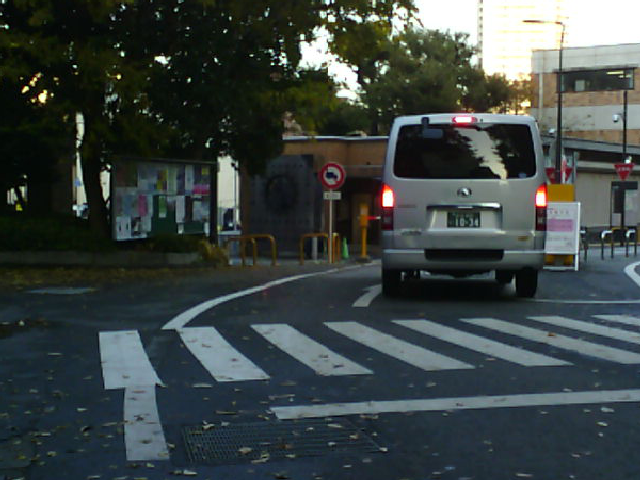}\hspace{0.5mm} &
			\includegraphics[width=1.85cm,height=1.85cm]{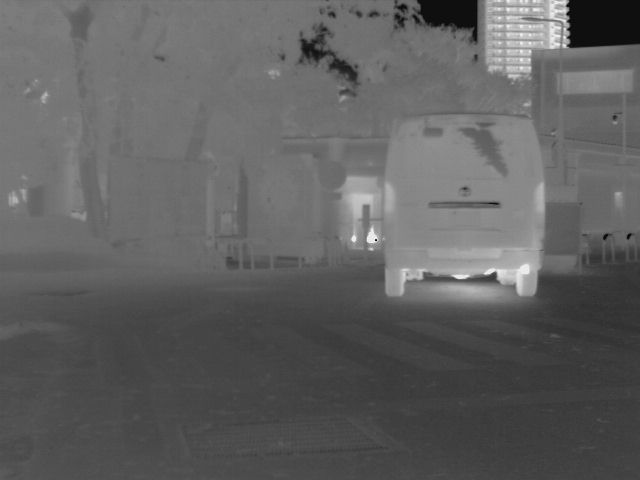}\hspace{0.5mm} &
			\includegraphics[width=1.85cm,height=1.85cm]{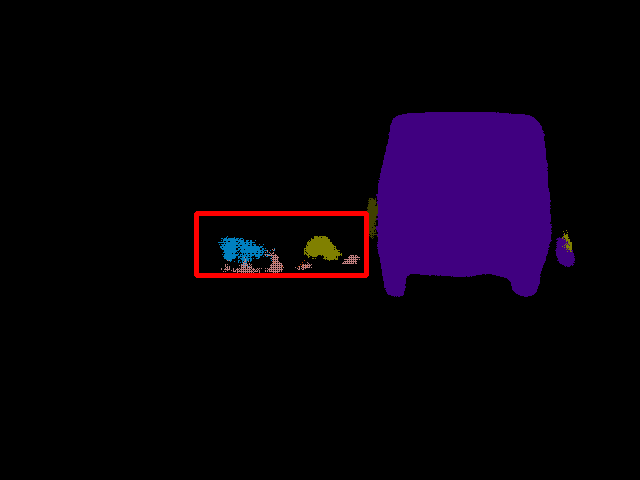}\hspace{0.5mm} &
			\includegraphics[width=1.85cm,height=1.85cm]{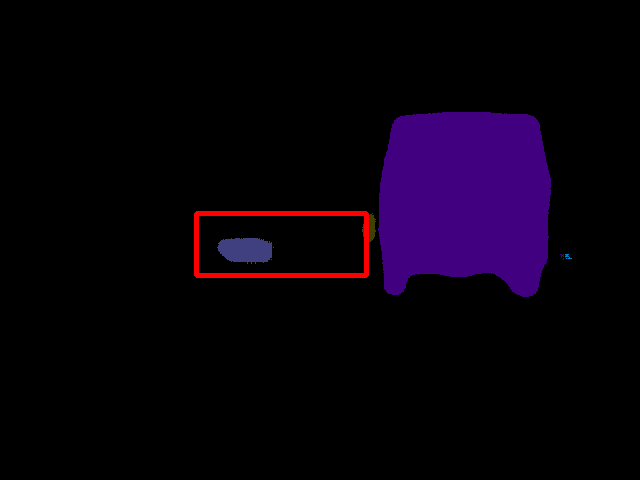}\hspace{0.5mm} &
			\includegraphics[width=1.85cm,height=1.85cm]{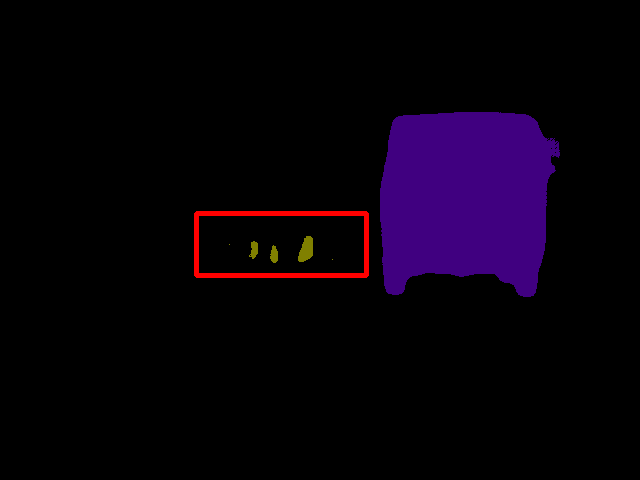}\hspace{0.5mm} &
                \includegraphics[width=1.85cm,height=1.85cm]{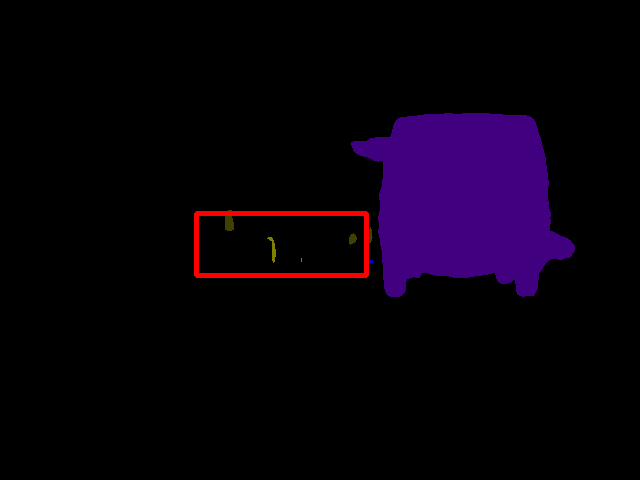}\hspace{0.5mm} &
                \includegraphics[width=1.85cm,height=1.85cm]{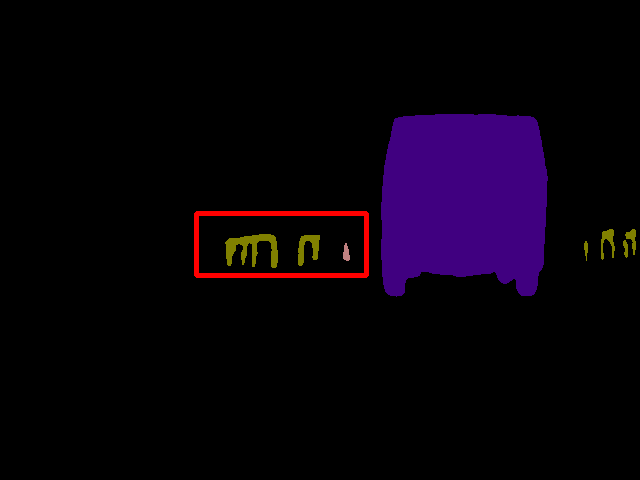}\hspace{0.5mm} &
                \includegraphics[width=1.85cm,height=1.85cm]{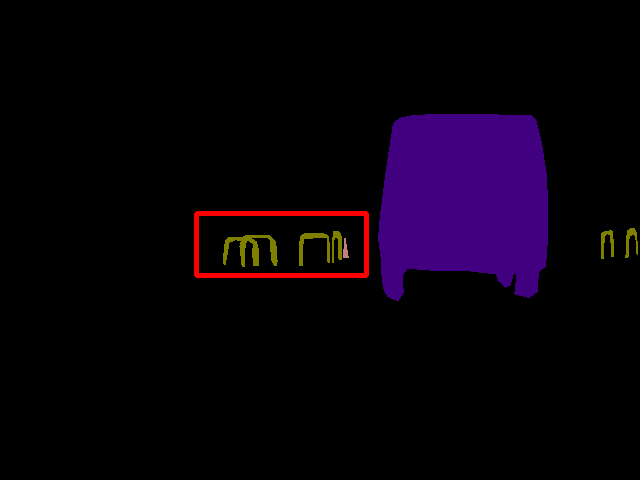}\hspace{0.5mm} \\
			\vspace{-0.5mm}
			\includegraphics[width=1.85cm,height=1.85cm]{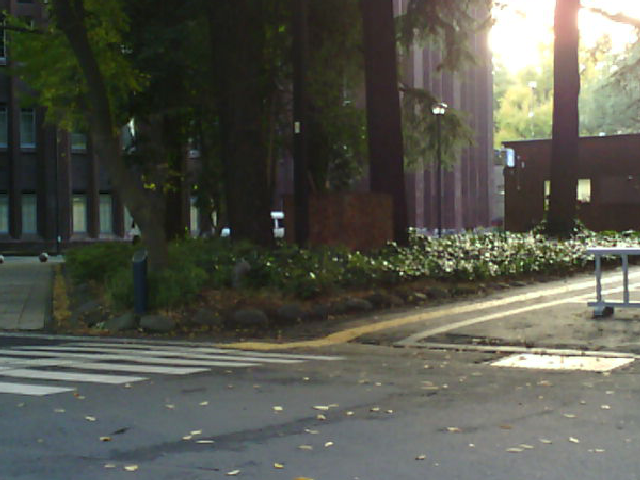}\hspace{0.5mm} &
			\includegraphics[width=1.85cm,height=1.85cm]{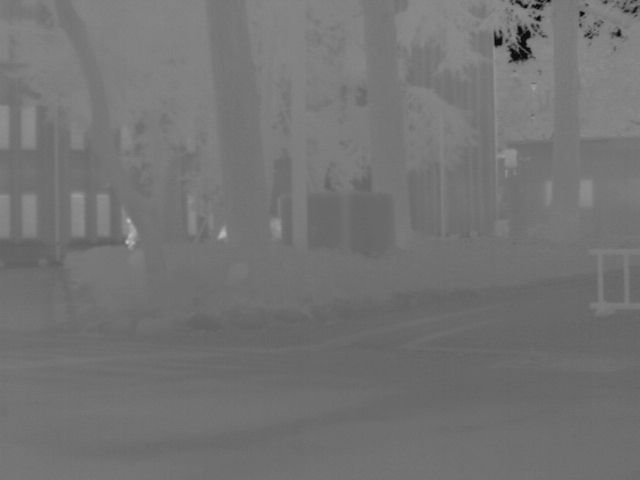}\hspace{0.5mm} &
			\includegraphics[width=1.85cm,height=1.85cm]{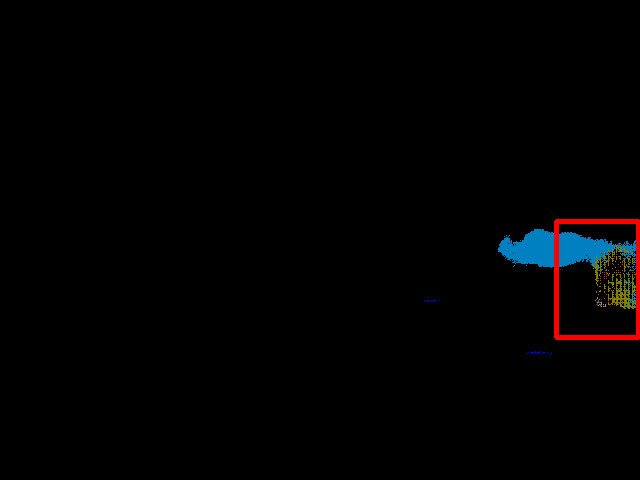}\hspace{0.5mm} &
			\includegraphics[width=1.85cm,height=1.85cm]{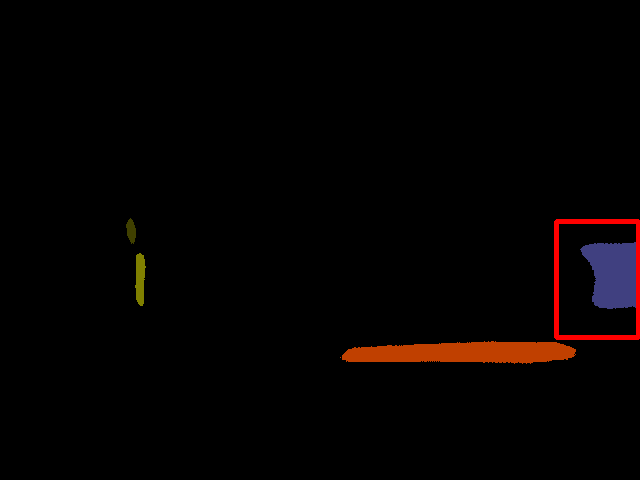}\hspace{0.5mm} &
			\includegraphics[width=1.85cm,height=1.85cm]{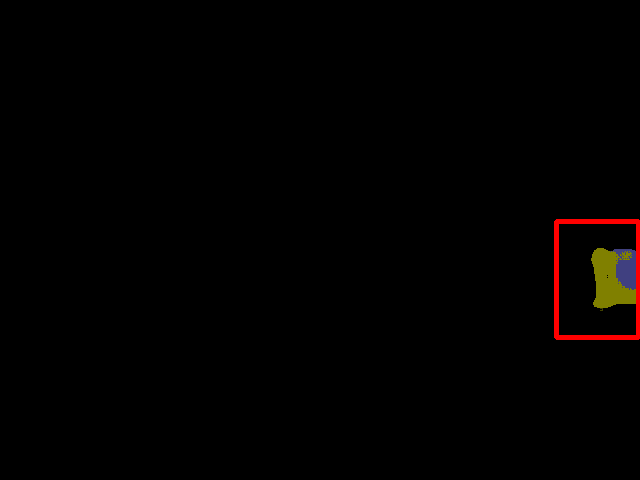}\hspace{0.5mm} &
                \includegraphics[width=1.85cm,height=1.85cm]{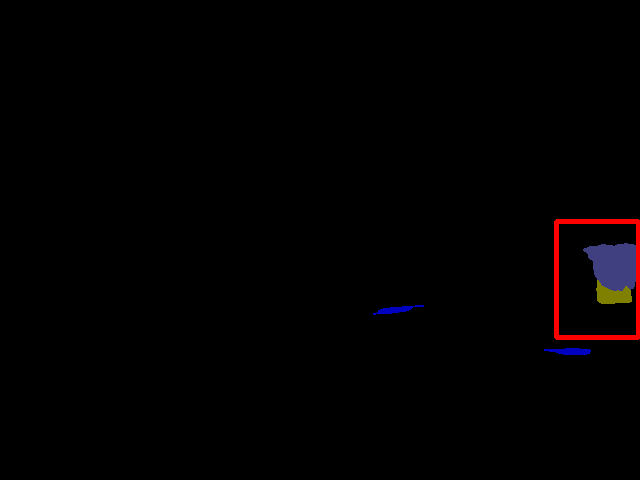}\hspace{0.5mm} &
                \includegraphics[width=1.85cm,height=1.85cm]{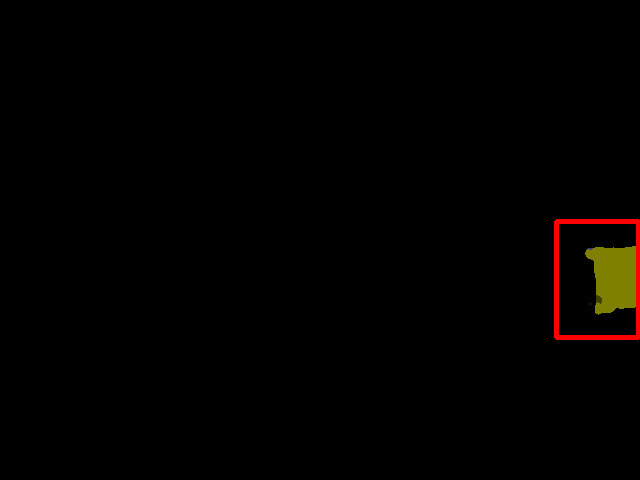}\hspace{0.5mm} &
                \includegraphics[width=1.85cm,height=1.85cm]{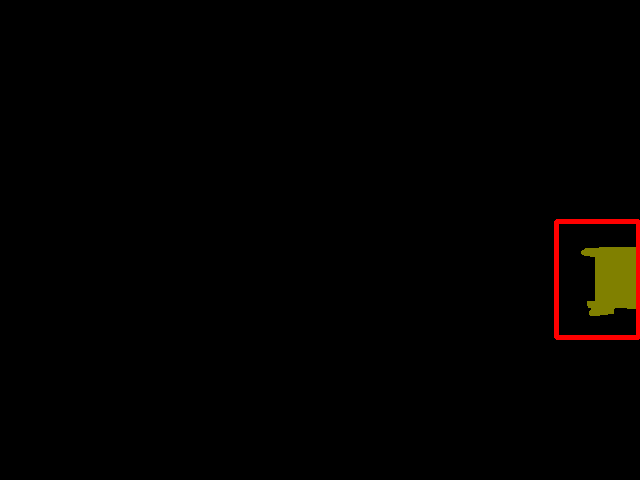}\hspace{0.5mm} & \vspace{-1.2mm} \\ [1pt]
			{\scriptsize RGB} & {\scriptsize Thermal} & {\scriptsize RTFNet~\cite{sun2019rtfnet}} & {\scriptsize FEANet~\cite{deng2021feanet}} & {\scriptsize EAEFNet~\cite{liang2023explicit}} & {\scriptsize CMX~\cite{liu2022cmx}} & {\scriptsize Sigma (Ours)} & {\scriptsize Ground Truth}\ \\
		\end{tabular}
	}
	\vspace{-3.4mm}
	\caption{Qualitative comparison on MFNet~\cite{ha2017mfnet} dataset. More results can be found in the supplementary material.}
	\label{fig:qualitative mfnet}

\end{figure*}
\begin{figure*}[!]
	\centering
 \vspace{-6pt}
	\resizebox{1\textwidth}{!}
	{
		\begin{tabular}{@{}c@{}c@{}c@{}c@{}c@{}c@{}c@{}c@{}c@{}c@{}c}
			\vspace{-0.5mm}
   \includegraphics[width=1.85cm,height=1.85cm]{figs/NYU_Results/RGB/1298.jpg}\hspace{0.5mm} &
			\includegraphics[width=1.85cm,height=1.85cm]{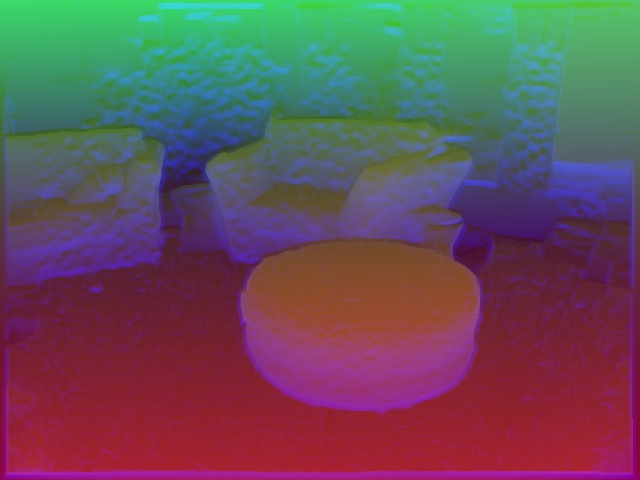}\hspace{0.5mm} &
			\includegraphics[width=1.85cm,height=1.85cm]{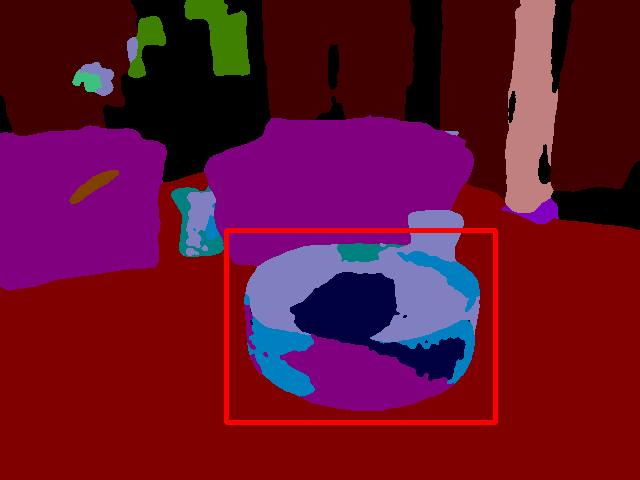}\hspace{0.5mm} &
			\includegraphics[width=1.85cm,height=1.85cm]{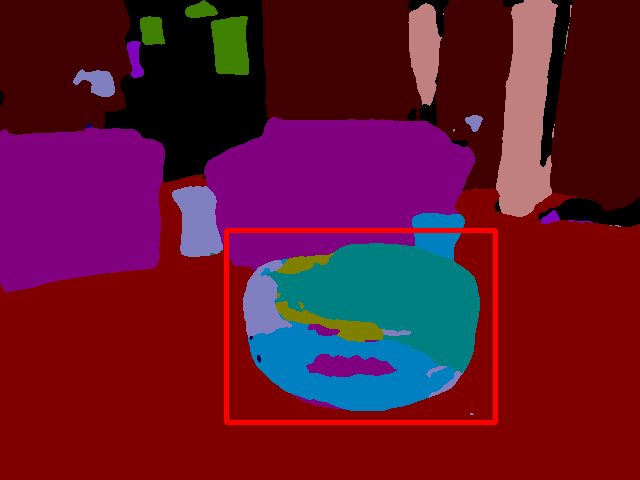}\hspace{0.5mm} &
			\includegraphics[width=1.85cm,height=1.85cm]{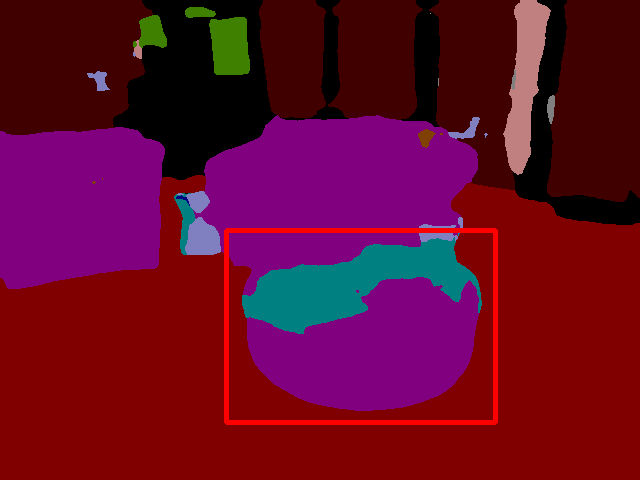}\hspace{0.5mm} &
                \includegraphics[width=1.85cm,height=1.85cm]{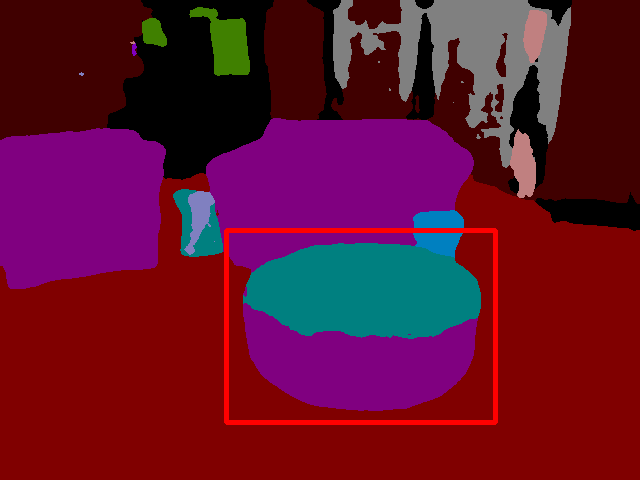}\hspace{0.5mm} &
                \includegraphics[width=1.85cm,height=1.85cm]{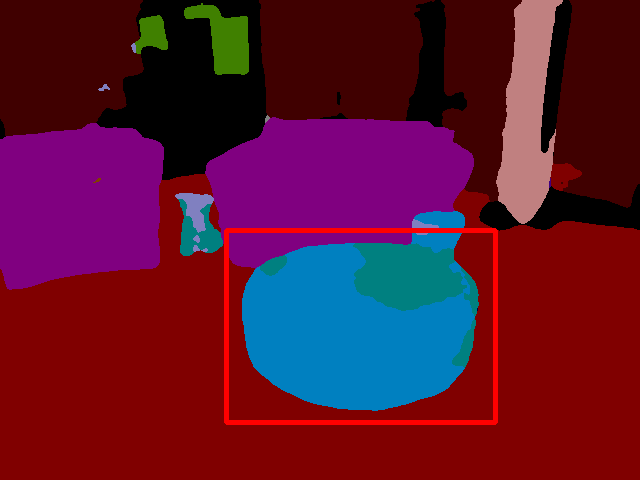}\hspace{0.5mm} &
                \includegraphics[width=1.85cm,height=1.85cm]{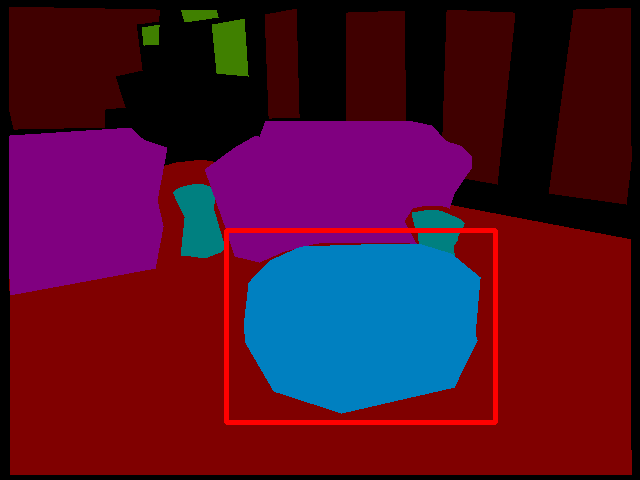}\hspace{0.5mm} 
			 \\
			\vspace{-0.5mm}
   \includegraphics[width=1.85cm,height=1.85cm]{figs/NYU_Results/RGB/870.jpg}\hspace{0.5mm} &
			\includegraphics[width=1.85cm,height=1.85cm]{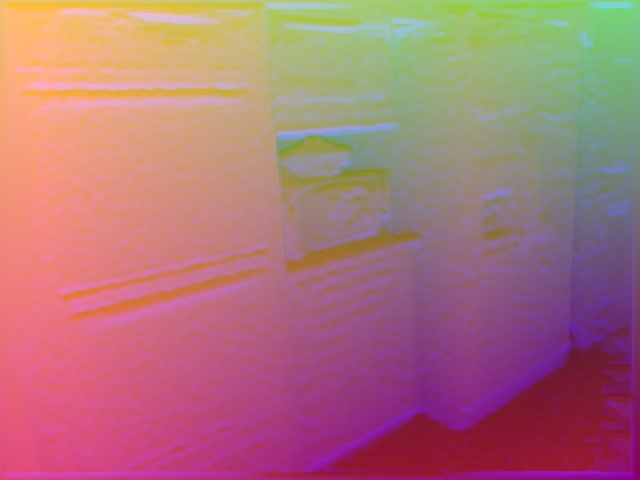}\hspace{0.5mm} &
			\includegraphics[width=1.85cm,height=1.85cm]{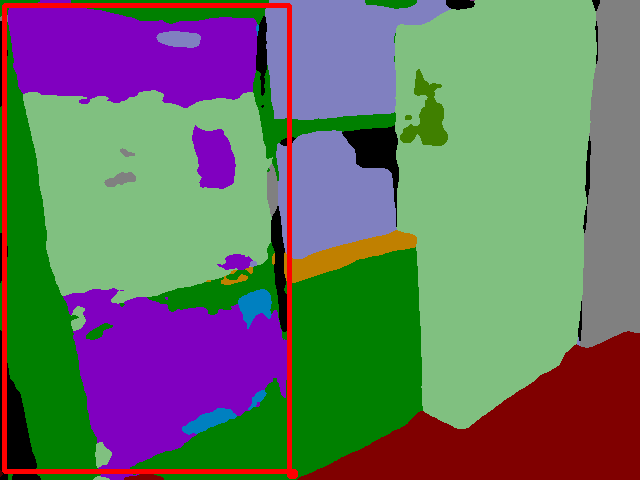}\hspace{0.5mm} &
			\includegraphics[width=1.85cm,height=1.85cm]{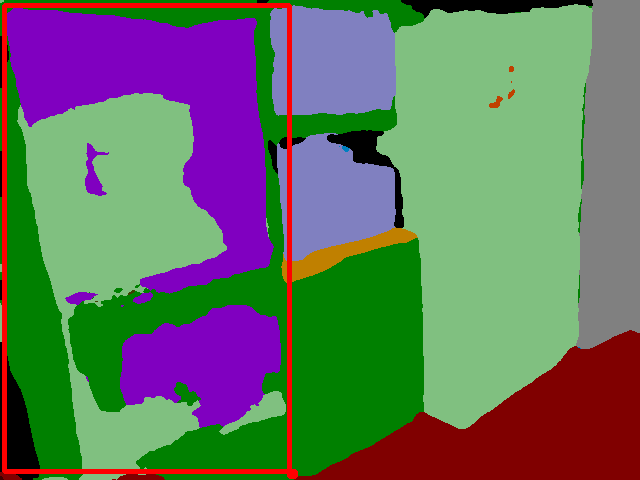}\hspace{0.5mm} &
			\includegraphics[width=1.85cm,height=1.85cm]{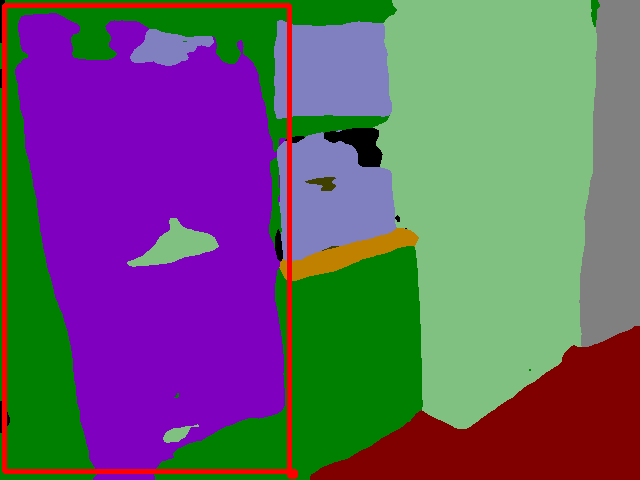}\hspace{0.5mm} &
                \includegraphics[width=1.85cm,height=1.85cm]{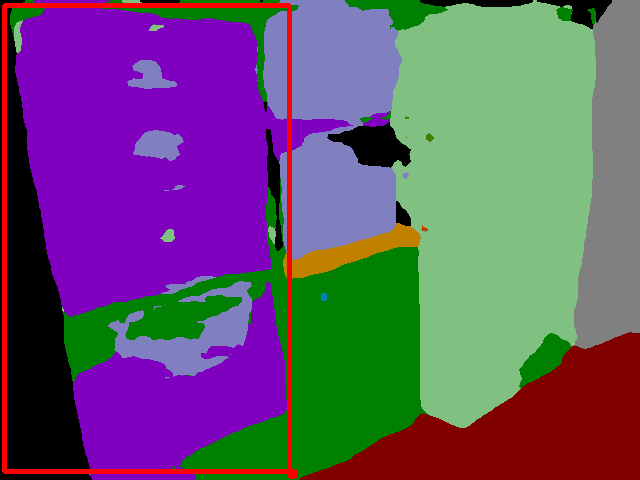}\hspace{0.5mm} &
                \includegraphics[width=1.85cm,height=1.85cm]{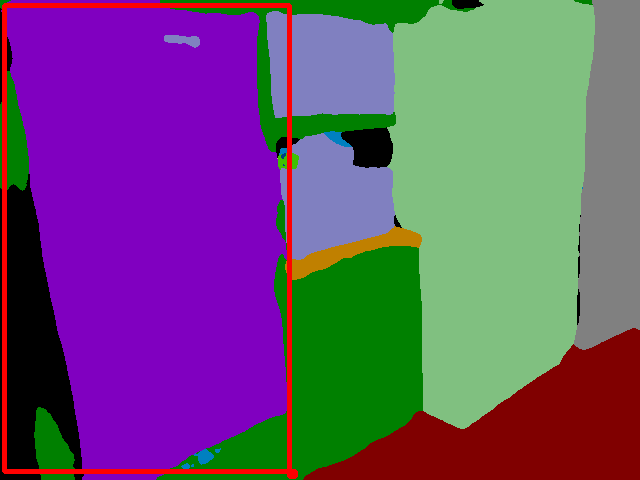}\hspace{0.5mm} &
                \includegraphics[width=1.85cm,height=1.85cm]{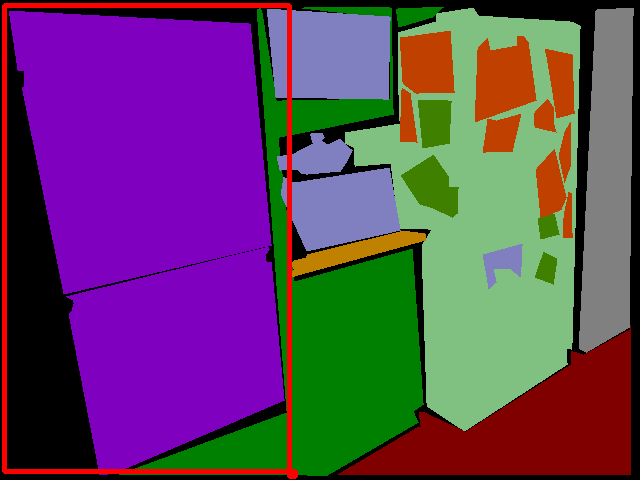}\hspace{0.5mm} \\
			\vspace{-0.5mm}
        \includegraphics[width=1.85cm,height=1.85cm]{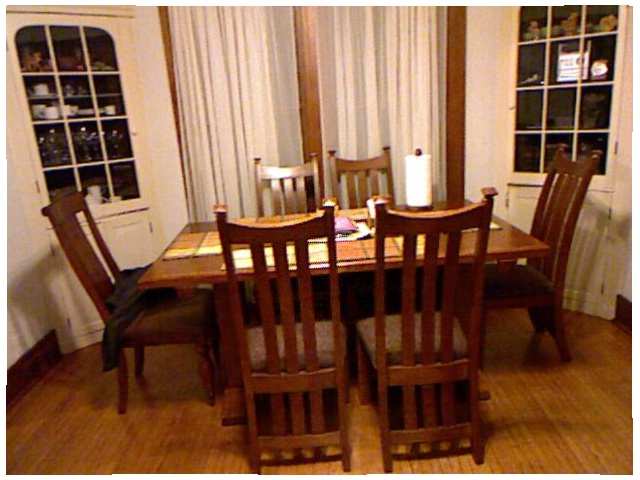}\hspace{0.5mm} &
\includegraphics[width=1.85cm,height=1.85cm]{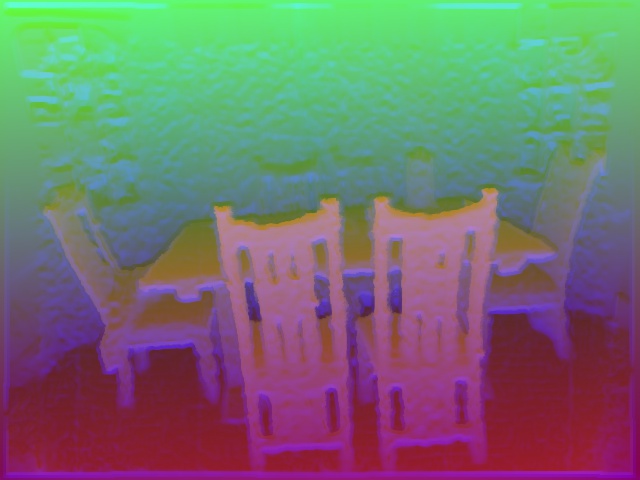}\hspace{0.5mm} &
\includegraphics[width=1.85cm,height=1.85cm]{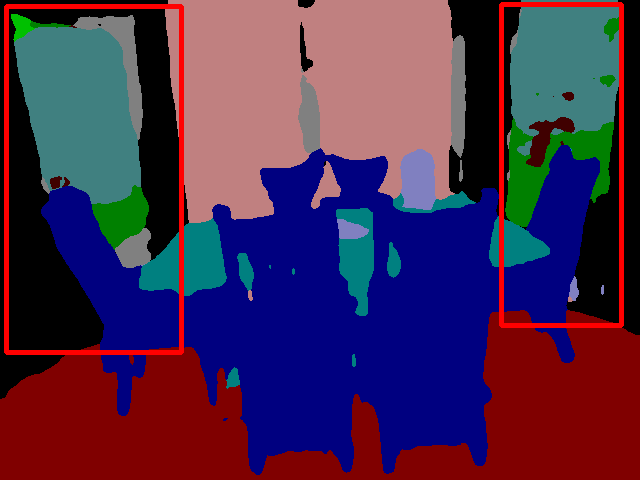}\hspace{0.5mm} &
\includegraphics[width=1.85cm,height=1.85cm]{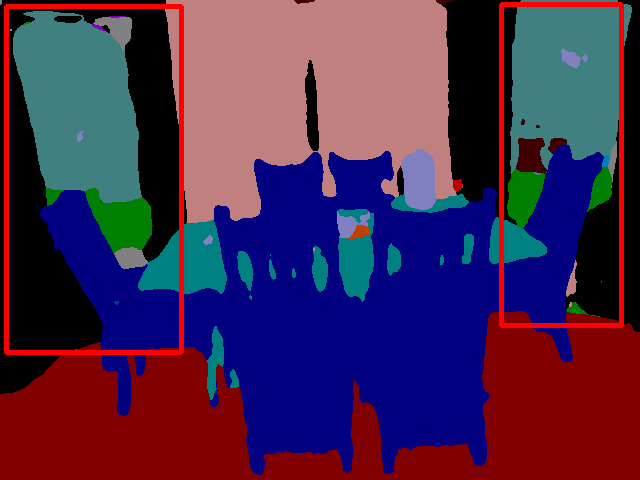}\hspace{0.5mm} &
\includegraphics[width=1.85cm,height=1.85cm]{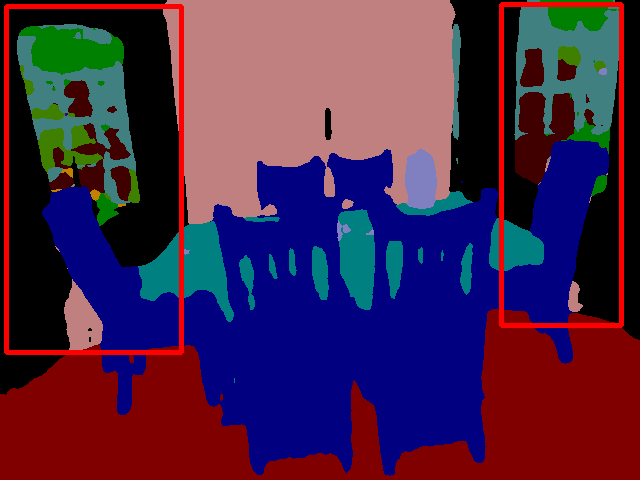}\hspace{0.5mm} &
\includegraphics[width=1.85cm,height=1.85cm]{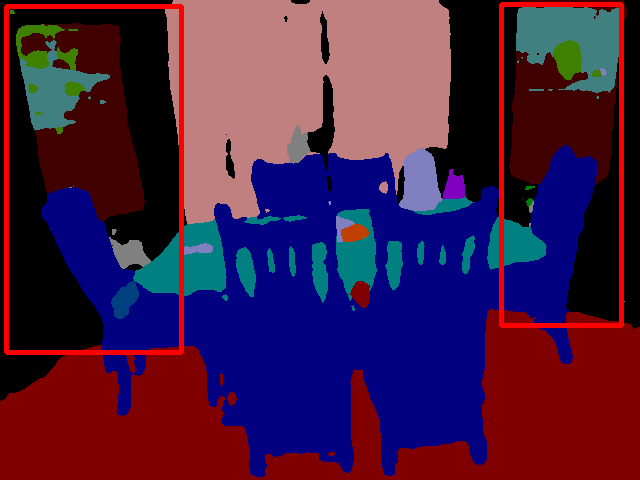}\hspace{0.5mm} &
\includegraphics[width=1.85cm,height=1.85cm]{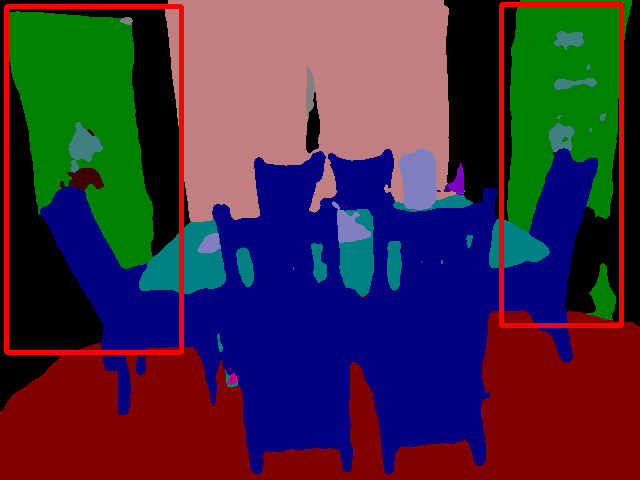}\hspace{0.5mm} &	
\includegraphics[width=1.85cm,height=1.85cm]{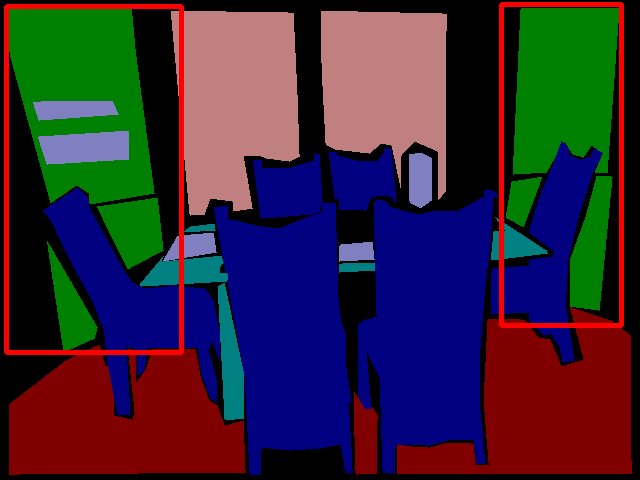}\hspace{0.5mm}&
			\vspace{-1.2mm} \\ [1pt]
			{\scriptsize RGB} & {\scriptsize HHA} & {\scriptsize SA-Gate~\cite{chen2020bi}} & {\scriptsize MultiMAE~\cite{bachmann2022multimae}} & {\scriptsize CMX~\cite{liu2022cmx}} & {\scriptsize CMNeXt~\cite{zhang2023delivering}} & {\scriptsize Sigma (Ours)} & {\scriptsize Ground Truth}\ \\
		\end{tabular}
	}
	\vspace{-2.8mm}
	\caption{Qualitative comparison on NYU Depth V2~\cite{silberman2012indoor} dataset. We use HHA images for better visualization of depth modality. More results can be found in the supplementary material.}
	\label{fig:qualitative nyudepth}
\vspace{-15pt}
\end{figure*}
\begin{figure*}[h]
\begin{center}
    \includegraphics[width=\linewidth]{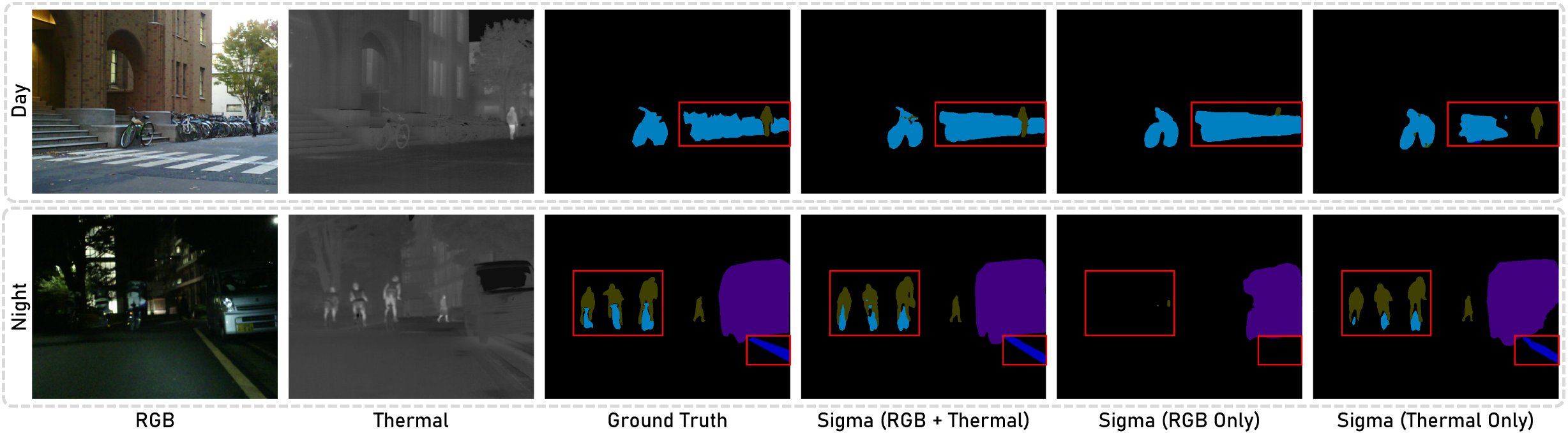}
\end{center}
\vspace{-19pt}
   \caption{Comparative analysis of semantic segmentation results: single-modal vs. multi-modal approach.}
\label{fig:motivation}
\vspace{-8pt}
\end{figure*}

\textbf{RGB-T Semantic Segmentation.}
Table~\ref{tab:rgbt}(a) shows the per-class semantic segmentation results, alongside comparisons of the model size and computational complexity, on the MFNet dataset.
It is observed that our tiny model surpasses other compared methods with fewer model parameters and FLOPs, and our base model achieves a $1.1\%$ performance improvement compared to the tiny variant.
In addition, as shown in Table~\ref{tab:rgbt}(b), our method outperforms other methods by more than $2\%$ on the PST900 dataset,
demonstrating the superiority of our proposed method.

The qualitative analysis in Fig.~\ref{fig:qualitative mfnet} reveals that our model outperforms the baseline models by generating more comprehensive segmentations and accurate classifications, notably in identifying intricate features like tactile paving and bollards. This indicates that our method can extract vital features and detailed information to improve the prediction.

\textbf{RGB-D Semantic Segmentation.}
In Table~\ref{table:rgbd results}, we compare Sigma against various RGB-D methods to validate its generalization capability.
Remarkably, our small model surpasses the performance of CMNeXt~\cite{zhang2023delivering} while employing only $69.8$M parameters, which is $49.8$M less than CMNeXt. This demonstrates our method's superior balance between accuracy and efficiency.

Fig.~\ref{fig:qualitative nyudepth} shows the capability to generate more coherent segmentation by utilizing depth information effectively. For instance, in the case of the round chair adjacent to the sofa, shadows cause baseline models to fragment the chair into multiple segments. Our method successfully recognizes it as a singular entity, highlighting its proficiency in leveraging depth data for segmentation.

\textbf{Qualitative Analysis of Dual-Modality Learning.}
In Figure~\ref{fig:motivation}, we examine the contributions of both RGB and thermal modalities to the final prediction. By leveraging information from both modalities, our model achieves more comprehensive segmentation and more precise boundary delineation.
The enhanced outcomes stem from fusing valuable information in both RGB and thermal modalities. Specifically, RGB enhances color distinction, whereas thermal imaging excels in texture differentiation. Integrating these modalities results in improved segmentation accuracy.

\begin{table}[h]
\centering
\resizebox{\linewidth}{!}{%
\begin{tabular}{c|c|c|cc|cc}
\toprule
\multirow{2}{*}[-0.5ex]{Method} & \multirow{2}{*}[-0.5ex]{Backbone} & \multirow{2}{*}[-0.5ex]{Params} & \multicolumn{2}{c|}{NYU Depth V2} & \multicolumn{2}{c}{SUN RGB-D} \\
\cmidrule(lr){4-5} \cmidrule(lr){6-7}& & & Input Size  & mIoU & Input Size  & mIoU \\
\midrule \midrule
        ACNet$_{\mathsf 19}$~\cite{hu2019acnet}&ResNet-50&116.6&$480\times 640$  & 48.3 &$530\times 730$& 48.1 \\
        SA-Gate$_{\mathsf 20}$~\cite{chen2020bi}&ResNet-101&110.9&$480\times 640$  & 52.4 &$530\times 730$& 49.4\\
        CEN$_{\mathsf 20}$~\cite{wang2020deep}&ResNet-101&118.2&$480\times 640$ &51.7&$530\times 730$ &50.2\\
        CEN$_{\mathsf 20}$~\cite{wang2020deep}&ResNet-152&133.9&$480\times 640$& 52.5&$530\times 730$ &51.1\\
        SGNet$_{\mathsf 21}$~\cite{chen2021spatial}&ResNet-101&64.7&$480\times 640$  & 51.1 &$530\times 730$& 
 48.6\\
        ShapeConv$_{\mathsf 21}$~\cite{cao2021shapeconv}&ResNext-101&86.8&$480\times 640$  & 51.3 &$530\times 730$ &48.6 \\ 
        ESANet$_{\mathsf 21}$~\cite{seichter2021efficient}&ResNet-34&31.2&$480\times 640$& 50.3&$480\times 640$&48.2\\
        FRNet$_{\mathsf 22}$~\cite{zhou2022frnet}&ResNet-34&85.5&$480\times 640$ &53.6&$530\times 730$& 51.8\\
        PGDENet$_{\mathsf 22}$~\cite{zhou2022pgdenet}&ResNet-34&100.7&$480\times 640$ 
 &53.7&$530\times730$  &51.0\\
        EMSANet$_{\mathsf 22}$~\cite{seichter2022efficient}&ResNet-34&46.9&$480\times 640$&51.0&$530\times 730$ &48.4\\ 
        TokenFusion$_{\mathsf 22}$~\cite{wang2022multimodal}& MiT-B2&26.0&$480\times640$  &53.3&$530\times 730$ 
 &50.3\\
        TokenFusion$_{\mathsf 22}$~\cite{wang2022multimodal} &MiT-B3&45.9&$480\times640$  &54.2&$530\times 730$ 
 &51.0\\
        MultiMAE$_{\mathsf 22}$~\cite{bachmann2022multimae}&ViT-B&95.2&$640\times640$ &56.0&$640\times 640$ 
 &51.1\\
        Omnivore$_{\mathsf 22}$~\cite{girdhar2022omnivore}&Swin-B&95.7&$480\times 640$  &54.0&--&--\\
        PDCNet$_{\mathsf 23}$~\cite{10185116}&ResNet-101&--&$480\times 480$&53.5&$480\times480$&49.6\\
        InvPT$_{\mathsf 23}$~\cite{invpt2022}&ViT-L&423$^{\dag}$&$480\times 640$&53.6&--&--\\
        T-Prompter$_{\mathsf 23}$~\cite{taskprompter2023}&T-Prompter-L&401$^{\dag}$&$480\times 640$&55.3&--&--\\
        SMMCL$_{\mathsf 24}$~\cite{dong2024understanding}&SegNeXt-B&--&$480\times 640$&55.8&--&--\\
        CMNeXt$_{\mathsf 23}$~\cite{zhang2023delivering} &MiT-B4&119.6&$480\times640$  &\underline{56.9}&$530\times 730$ 
 &\underline{51.9}\\
        \midrule
        \rowcolor{gray!20}
        Sigma (Ours)&VMamba-T&48.3&$480\times 640$&53.9&$480\times640$&50.0\\
        \rowcolor{gray!20}
        Sigma (Ours)&VMamba-S&69.8&$480\times 640$&\textbf{57.0}&$480\times640$&\textbf{52.4}\\
\bottomrule
\end{tabular}%
}
\vspace{-6pt}
\caption{Comparison of RGB-D semantic segmentation on NYU Depth V2~\cite{silberman2012indoor} and SUN RGB-D~\cite{song2015sun}. $^{\dag}$ indicates the parameters for multi-task learning reported from \cite{taskprompter2023}.}
\label{table:rgbd results}
\end{table}

\subsection{Ablation Studies}

\begin{table}[h]
\setlength\tabcolsep{9pt}
\centering
\resizebox{0.98\linewidth}{!}{
\begin{tabular}{c|c|cc|c|cc}
\toprule   
\#&Encoder&CroMB & ConMB & Decoder & mIoU ($\nabla$)\\ \midrule
\rowcolor{gray!20}
1&VMamba-T&\Checkmark & \Checkmark &  CMD &  60.5 (\textcolor{MidnightBlue}{0.0}) 
  \\
2&VMamba-T&\XSolidBrush & \Checkmark& CMD & 59.9 (\textcolor{MidnightBlue}{-0.6})
   \\
3&VMamba-T&\Checkmark& \XSolidBrush& CMD &  59.7 (\textcolor{MidnightBlue}{-0.8})
\\
4&VMamba-T&\XSolidBrush& \XSolidBrush& CMD &  58.4 (\textcolor{MidnightBlue}{-2.1})
\\
5&VMamba-T&\Checkmark & \Checkmark &  MLP &  59.4 (\textcolor{MidnightBlue}{-1.1})
 \\
6&VMamba-T&\XSolidBrush & \XSolidBrush& MLP & 57.5 (\textcolor{MidnightBlue}{-3.0})
\\ 
7&VMamba-T&\Checkmark& \Checkmark& Swin & 59.6 (\textcolor{MidnightBlue}{-0.9})
\\ 
8&VMamba-T&\Checkmark& \Checkmark& UperNet & 59.0 (\textcolor{MidnightBlue}{-1.5})
\\ 
9&Swin-T&\XSolidBrush& \XSolidBrush& Swin & 56.3 (\textcolor{MidnightBlue}{-4.2})
\\ 
\bottomrule
\end{tabular}
}
\vspace{-4pt}
\caption{Effectiveness of different components. We report the mIoU metric and the relative decrease ($\nabla$) on the MFNet~\cite{ha2017mfnet} dataset. When both CroMB and ConMB show \scalebox{0.7}{\raisebox{-0.19em}\XSolidBrush} marks, we use a simple feature summation operation to obtain the fused feature.}
\label{tab:ablation components}
\vspace{-12pt}
\end{table}

\looseness = -1
\textbf{Effectiveness of Key Components.}
As detailed in Table~\ref{tab:ablation components}, we carried out ablation studies with Sigma-Tiny on the MFNet~\cite{ha2017mfnet} dataset. Compared to the complete model, eliminating CroMB and ConMB individually results in decreases of $0.6\%$ and $0.8\%$ in performance, respectively. The removal of both blocks leads to a performance decline of $2.1\%$, highlighting the effectiveness of our proposed fusion module.
Additionally, we evaluate our proposed decoder against a simple Multi-Layer Perceptron (MLP) decoder, a Swin-Transformer decoder~\cite{liu2021swin}, and an UperNet decoder~\cite{xiao2018unified}, discovering that our decoder surpasses them by $1.1\%$, $0.9\%$, and $1.5\%$, respectively.
To further assess the effectiveness, we substitute our encoder with a pre-trained Swin Transformer~\cite{liu2021swin} Tiny backbone and apply Swin Transformer blocks in the decoder.
The results suggest that our design with SSM can be more effective than a simple integration of Transformers.

\begin{table}[h]
\setlength\tabcolsep{9pt}
\centering
\resizebox{0.89\linewidth}{!}{
\begin{tabular}{c|c|c|c}
\toprule   
\#&Encoder& Cross Attention Matrix & mIoU ($\nabla$)\\ \midrule
\rowcolor{gray!20}
1&VMamba-T& C &  60.5 (\textcolor{MidnightBlue}{0.0}) 
  \\
2&VMamba-T& B & 59.3 (\textcolor{MidnightBlue}{-1.2})
   \\
3&VMamba-T& D & 59.6 (\textcolor{MidnightBlue}{-0.9})
   \\
4&VMamba-T& B \& C &  59.7 (\textcolor{MidnightBlue}{-0.8})
\\ 
5&VMamba-T& C \& D &  59.8 (\textcolor{MidnightBlue}{-0.7})
\\ 
\bottomrule
\end{tabular}
}
\vspace{-4pt}
\caption{Comparison of attention matrix used in CroMB. 
}
\label{tab:ablation cromb}
\vspace{-4pt}
\end{table}
\textbf{Comparison of Attention Matrix in CroMB.}
In CroMB, we exchange matrix \(C\) of two modalities for cross-modal information interaction. For comparison, we conduct experiments to use matrix \(C\), \(D\), \(B \& C\), and \(C \& D\) in CroMB. As shown in Table~\ref{tab:ablation cromb}, using the matrix \(C\) achieves the optimal performance. This supports our claim that matrix \(C\) is used to decode the information from the hidden state and can guide the reconstruction for complementary modality.
\section{Conclusion}
\looseness = -1
In this work, we propose Sigma, a novel Siamese Mamba Network for multi-modal semantic segmentation, which explores the application of state space models in multi-modal scene understanding for the first time.
A Siamese backbone is first applied to extract robust global long-range dependencies with linear complexity. 
Then we introduce a fusion module that incorporates a cross-selective scan and a concat-selective scan.
Finally, we design a channel-aware Mamba decoder to extract essential information from the fused features for predictions.
Extensive experiments on RGB-T and RGB-D semantic segmentation benchmarks show the superiority of our method in both accuracy and efficiency.

\section*{Acknowledgement}
\looseness=-1
This work has been funded in part by the Army Research Laboratory (ARL) award W911NF-23-2-0007, DARPA award FA8750-23-2-1015, and ONR award N00014-23-1-2840.
We also appreciate the constructive suggestions provided by Ce Zhang.


{\small
\bibliographystyle{ieee_fullname}
\bibliography{egbib}
}

\clearpage
\appendix
\setcounter{page}{1}
\maketitlesupplementary

\renewcommand{\thesection}{\Alph{section}}
\renewcommand\thefigure{\Alph{section}\arabic{figure}} 
\renewcommand\thetable{\Alph{section}\arabic{table}}  
\setcounter{section}{0}
\setcounter{figure}{0} 
\setcounter{table}{0} 


\section{Experimental Details}
During training, we perform data augmentation, including random flipping and scaling with random scales $[0.5,1.75]$, to all datasets.
We adopt VMamba~\cite{liu2024vmamba} pre-trained on ImageNet~\cite{russakovsky2015imagenet} as the backbone, which includes three versions, namely VMamba-Tiny, VMamba-Small, and VMamba-Base. The detailed settings of the three models are listed in Table~\ref{tab:vmamba size}.  We select AdamW optimizer~\cite{kingma2014adam} with weight decay $0.01$.
The original learning rate is set to $6e^{-5}$ and we employ a poly learning rate schedule with 10 warm-up epochs.
We use cross-entropy as the loss function.
When reporting testing results on NYU Depth V2~\cite{silberman2012indoor} and SUN RGB-D~\cite{song2015sun} datasets, we use multiple scales $\{0.75,1,1.25\}$ according to most previous RGB-Depth semantic segmentation methods~\cite{liu2022cmx, zhang2023delivering}.
We use mean Intersection over Union (mIoU) averaged across semantic classes as the evaluation metric to measure the segmentation performance.
For each of the datasets, more implementation details are described as follows.

\noindent\textbf{MFNet dataset.} The tiny and small backbones are trained on four 3090Ti GPUs and the base backbone is trained on four A6000 GPUs. We use the original image size of $640\times480$ for training and inference. The batch size is set to $8$ for training. A single 3090Ti GPU is used for inferencing all the models.

\noindent\textbf{PST900 dataset.} The tiny and small backbones are trained on two A6000 GPUs. We use the original image size of $1280\times720$ for training and inference. The batch size is set to $4$ for training. A single A6000 GPU is used for inferencing all the models.

\noindent\textbf{NYU Depth V2 dataset.} Unlike other methods~\cite{liu2022cmx, chen2020bi} to use HHA format of depth images for training, we directly use raw depth images and we found no apparent performance difference between the formats. We take the whole image with the size $640{\times}480$ for training and inference. 4 3090Ti GPUs are used to train the tiny and small backbones with batch size 8, and 4 A6000 GPUs are used to train the base model.

\noindent\textbf{SUN-RGBD dataset.}
Unlike previous methods which use larger resolution input ($730\times530$~\cite{zhang2023delivering, wang2022multimodal} or $640\times640$~\cite{bachmann2022multimae}), we adopt the input resolution of $640{\times}480$ and keep the same training settings as NYU Depth V2 dataset. We also use raw depth images instead of HHA format for training.

\begin{table}[h]
\setlength\tabcolsep{9pt}
\centering
\resizebox{\linewidth}{!}{
\begin{tabular}{c|cccc|c}
\toprule   
\multirow{2}{*}[-0.7ex]{Backbone} & \multicolumn{4}{c|}{VSS Block Number} & \multirow{2}{*}[-0.7ex]{\makecell{Embedded \\ Dimension}} \\
\cmidrule{2-5}
 & Stage 1 & Stage 2 & Stage 3 & Stage 4 & \\
\midrule \midrule
VMamba-Tiny & 2 & 2 & 9 & 2 &  96 
  \\
VMamba-Small & 2 & 2 & 27 & 2 &  96 
  \\
VMamba-Base & 2 & 2 & 27 & 2 &  128 
  \\
\bottomrule
\end{tabular}
}
\vspace{-5pt}
\caption{Details about three versions of backbone.}
\label{tab:vmamba size}
\vspace{-10pt}
\end{table}

\section{Daytime and Nighttime
Performance}
To explore the effectiveness of our method on daytime and nighttime RGB-T images, we use the MFNet~\cite{ha2017mfnet} dataset and follow CMX~\cite{liu2022cmx} to use 205 daytime images and 188 nighttime images in the test set for evaluation. As shown in Table~\ref{tab:day_night}, our method delivers better results on both daytime and nighttime results, demonstrating the effectiveness of our proposed method.

\begin{table}[h]
\setlength\tabcolsep{9pt}
\centering
\resizebox{\linewidth}{!}{
\begin{tabular}{c | c | c | c}
    \toprule
    \textbf{Method} & \textbf{Modal} & \textbf{Daytime} & \textbf{Nighttime}\\
    \midrule\midrule
    FRRN~\cite{pohlen2017full} & RGB & 40.0 & 37.3 \\
    DFN~\cite{yu2018dfn} & RGB & 38.0 & 42.3\\
    BiSeNet~\cite{yu2018bisenet} & RGB & 44.8 & 47.7 \\
    SegFormer-B2~\cite{xie2021segformer} & RGB & 48.6 & 49.2\\
    SegFormer-B4~\cite{xie2021segformer} & RGB & 49.4 & 52.4 \\
    \midrule
    MFNet~\cite{ha2017mfnet} & RGB-T & 36.1 & 36.8\\
    FuseNet~\cite{hazirbas2016fusenet} & RGB-T & 41.0 & 43.9\\
    RTFNet~\cite{sun2019rtfnet} & RGB-T & 45.8 & 54.8\\
    FuseSeg~\cite{sun2020fuseseg} & RGB-T & 47.8 & 54.6\\
    GMNet~\cite{zhou2021gmnet} & RGB-T & 49.0 & 57.7\\
    CMX (MiT-B2)~\cite{liu2022cmx} & RGB-T & 51.3 & 57.8\\
    CMX (MiT-B4)~\cite{liu2022cmx} & RGB-T & {52.5} & {59.4}\\
    \midrule
    \rowcolor{gray!15} Sigma (VMamba-T) & RGB-T & \underline{54.1} & 59.0\\
    \rowcolor{gray!15} Sigma (VMamba-S) & RGB-T & \textbf{55.0} & \underline{60.0}\\
    \rowcolor{gray!15} Sigma (VMamba-B) & RGB-T & \underline{54.1} & \textbf{60.9}\\
    \bottomrule
\end{tabular}
}
\vspace{-5pt}
\caption{Performance comparison on daytime and nighttime MFNet~\cite{ha2017mfnet} dataset. We use mIoU ($\%$) for evaluation.}
\label{tab:day_night}
\vspace{-10pt}
\end{table}

\section{Ablation Studies}
Apart from the ablation studies on the effect of each of our components, we further conduct experiments on the detailed design of the State Space Models. 
In Table~\ref{tab:ablation_supp}, we compare the effect of the state size in State Space Models and the number of CVSS blocks in our Mamba decoder.
From the table, we can find that setting the state size to 4 and the decoder layers to [4,4,4] leads to the optimal result.
\begin{table}[h]
\setlength\tabcolsep{9pt}
\vspace{-7pt}
\centering
\resizebox{\linewidth}{!}{
\begin{tabular}{c|c|cc|c}
\toprule   
\#&Encoder&State Size & Decoder Layers & mIoU ($\nabla$)\\ \midrule
\rowcolor{gray!20}
1&VMamba-T& 4 &  [4,\;4,\;4] &  60.5 (\textcolor{MidnightBlue}{0.0}) 
  \\
2&VMamba-T& 4 &  [3,\;3,\;3] &  60.2 (\textcolor{MidnightBlue}{0.3}) 
  \\
3&VMamba-T& 4 &  [2,\;2,\;2] &  59.4 (\textcolor{MidnightBlue}{1.1}) 
    \\
4&VMamba-T& 8 &  [4,\;4,\;4] &  60.3 (\textcolor{MidnightBlue}{0.2}) 
  \\
5&VMamba-T& 16 &  [4,\;4,\;4] &  59.7 (\textcolor{MidnightBlue}{0.8}) 
  \\
\bottomrule
\end{tabular}
}
\vspace{-5pt}
\caption{Ablation studies of decoder layers and the space size of the state space models on the MFNet~\cite{ha2017mfnet} dataset.}
\vspace{-5pt}
\label{tab:ablation_supp}
\end{table}



\section{Complexity Comparison of CroMB and Self-Attention}
\begin{figure}[h]
\begin{center}
    \includegraphics[width=0.8\linewidth]{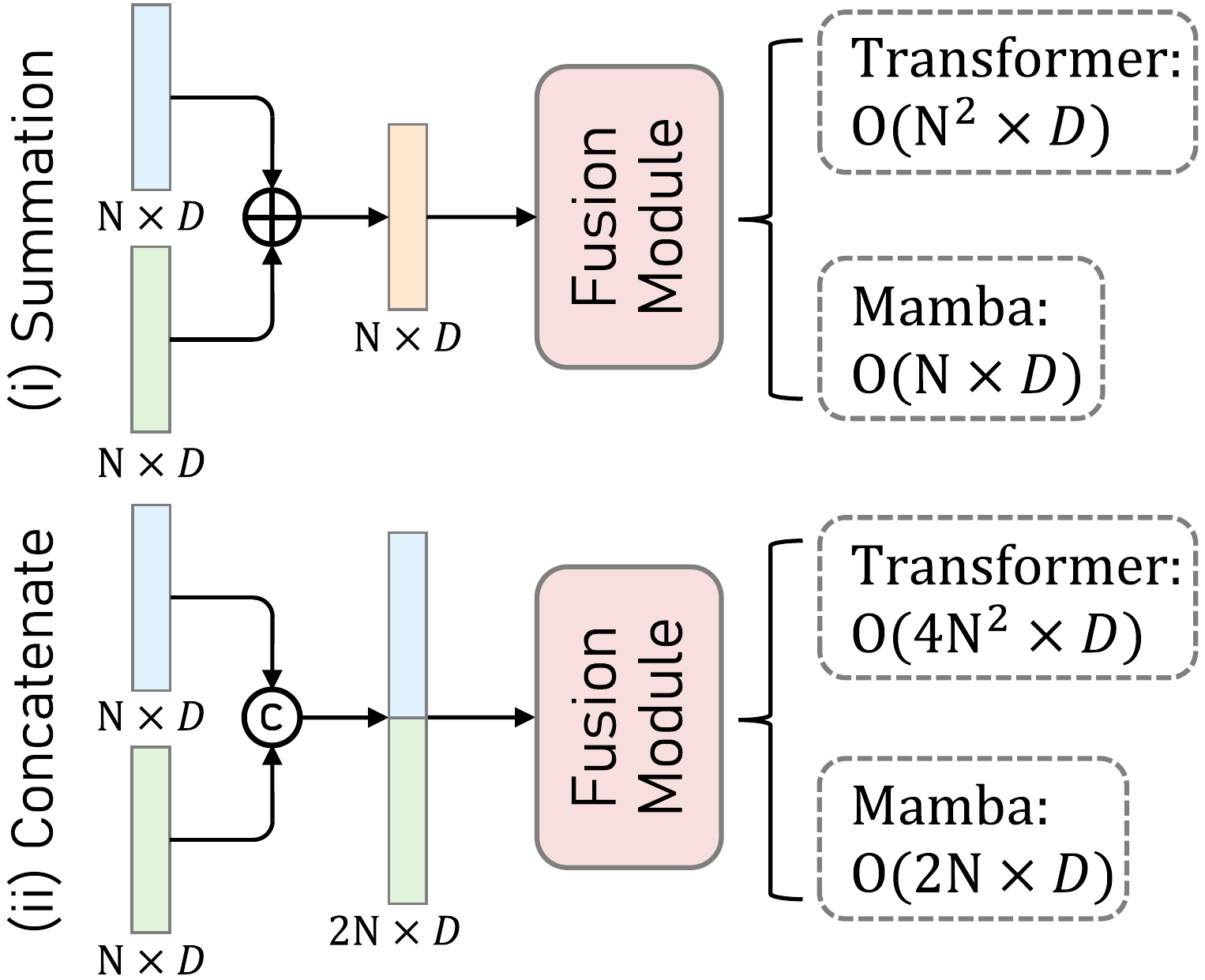}
\end{center}
\vspace{-12pt}
   \caption{Comparative analysis of complexity across different fusion methods utilizing Transformer and Mamba: Mamba-based fusion approaches significantly reduce complexity by an order of magnitude compared to their Transformer-based counterparts.}
\label{fig:theoretical comparison}
\vspace{-8pt}
\end{figure}

\begin{figure}[h]
\begin{center}
    \includegraphics[width=0.99\linewidth]{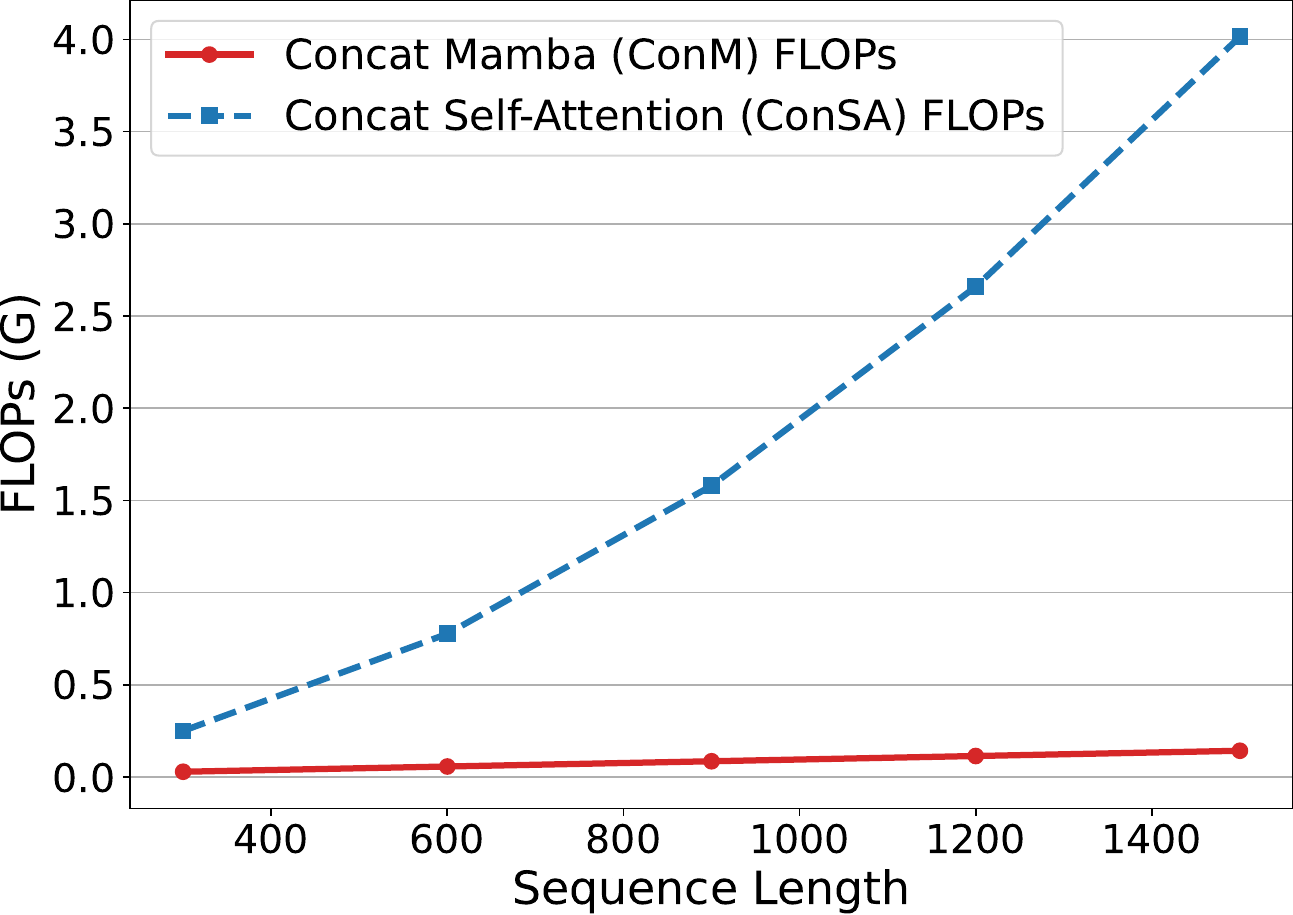}
\end{center}
\vspace{-12pt}
   \caption{Qualitative computation comparison of \textcolor[RGB]{31,119,180}{Concat Self-Attention (ConSA)} and our \textcolor[RGB]{214,39,40}{Concat Mamba (ConM)} mechanism.}
\label{fig:cromb_flops}
\vspace{-5pt}
\end{figure}

As shown in Fig.~\ref{fig:theoretical comparison}, we compare the theoretical complexity of fusion methods using Mamba and Transformer. This shows the linear scalability advantage of Mamba over quadratic Transformer-based methods. In Fig.~\ref{fig:cromb_flops}, we illustrate the qualitative growth in FLOPs as the input sequence length increases. It is evident that our ConM mechanism has much less computation consumption than constituting the State Space Model with Self-Attention. This underscores the exceptional efficiency of our proposed ConM in integrating multi-modal features.
\begin{table}[h]
\setlength\tabcolsep{9pt}
\centering
\resizebox{\linewidth}{!}{
\begin{tabular}{c|ccc|cc}
\toprule   
\multirow{2}{*}[-0.7ex]{Stage} & \multicolumn{3}{c|}{Feature Size} & \multicolumn{2}{c}{FLOPs (G)} \\
\cmidrule{2-6}
 & Height & Weight & Channel 
 & ConM & ConSA \\
\midrule \midrule
1 & 120 & 160 & 96 & 1.82 &  --
  \\
2 & 60 & 80 & 192 & 1.71 &  77.89 
  \\
3 & 30 & 40 & 384 & 1.65 &  15.94 
  \\
4 & 15 & 20 & 768 & 1.62 &  8.19
  \\
\bottomrule
\end{tabular}
}
\vspace{-5pt}
\caption{Quantitative comparison of computation complexity between Concat Self-Attention (ConSA) and our proposed Concat Mamba (ConM) mechanism.}
\label{tab:flops quantitative}
\vspace{-4pt}
\end{table}

In Table~\ref{tab:flops quantitative}, we compare the floating-point operations per second (FLOPS) of our proposed ConMB and Concat Mamba (ConSA), which employs self-attention instead of SSM. The ``Stage'' column indicates the four encoding stages, with the input feature size for each fusion block also provided. The findings reveal that ConMB maintains low FLOPs across all stages, whereas the FLOPs for the self-attention mechanism escalate significantly with increases in height and width.

\section{Limitations and Future Work}
\label{sec:limitations}
While Sigma has achieved outstanding results in various RGB-X semantic segmentation tasks, two main limitations remain.
1) \textit{Underutilization of Mamba for Longer Sequences:} Mamba's capability to handle extremely long sequences is a significant advantage, particularly beneficial in fusion tasks involving more than two modalities. However, our current exploration primarily focuses on the application of Mamba for two modalities, potentially not fully leveraging its capacity for modeling longer sequences. Future work will aim to investigate Mamba's performance on datasets with a greater variety of modalities, such as the DELIVER benchmark. This exploration is pivotal for advancing research on enabling autonomous agents to navigate environments using multiple sensors, including RGB, depth, thermal, and LiDAR.
2) \textit{Memory Consumption in the Mamba Encoder:} The Mamba encoder scans image features from four directions, allowing each pixel to assimilate information from its surrounding pixels. This approach, however, quadruples memory usage, posing a challenge for deployment on lightweight edge devices. Future endeavors will seek to incorporate positional information through alternative methods, such as positional encoders, and employ a 1D SSM to diminish computational and memory demands.

\section{More Qualitative Results}
\begin{figure*}[h]
	\centering
	\vspace{-2mm}
	\resizebox{1\textwidth}{!}
	{
		\begin{tabular}{@{}c@{}c@{}c@{}c@{}c@{}c@{}c@{}c@{}c@{}c@{}c}
			\vspace{-0.5mm}
			\includegraphics[width=1.85cm,height=1.85cm]{figs/MFNet_Results/RGB/01402D.png}\hspace{0.5mm} &
			\includegraphics[width=1.85cm,height=1.85cm]{figs/MFNet_Results/Modal/01402D.png}\hspace{0.5mm} &
			\includegraphics[width=1.85cm,height=1.85cm]{figs/MFNet_Results/Pred_RTFNet/01402D.png}\hspace{0.5mm} &
			\includegraphics[width=1.85cm,height=1.85cm]{figs/MFNet_Results/Pred_FEANet/01402D.png}\hspace{0.5mm} &
			\includegraphics[width=1.85cm,height=1.85cm]{figs/MFNet_Results/Pred_EAEFNet/01402D.png}\hspace{0.5mm} &
                \includegraphics[width=1.85cm,height=1.85cm]{figs/MFNet_Results/Pred_CMX/01402D.png}\hspace{0.5mm} &
                \includegraphics[width=1.85cm,height=1.85cm]{figs/MFNet_Results/Pred_Sigma/01402D.png}\hspace{0.5mm} &
                \includegraphics[width=1.85cm,height=1.85cm]{figs/MFNet_Results/gt/01402D.png}\hspace{0.5mm} \\
			\vspace{-0.5mm}
			\includegraphics[width=1.85cm,height=1.85cm]{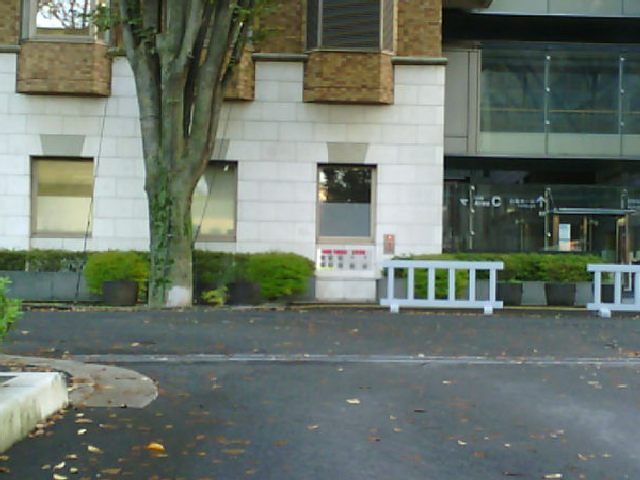}\hspace{0.5mm} &
			\includegraphics[width=1.85cm,height=1.85cm]{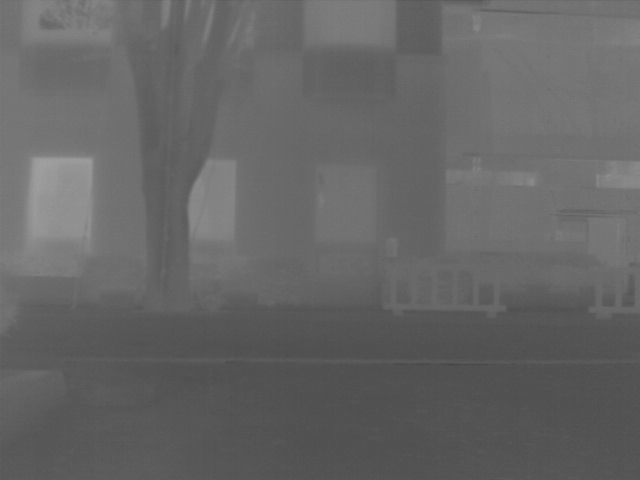}\hspace{0.5mm} &
			\includegraphics[width=1.85cm,height=1.85cm]{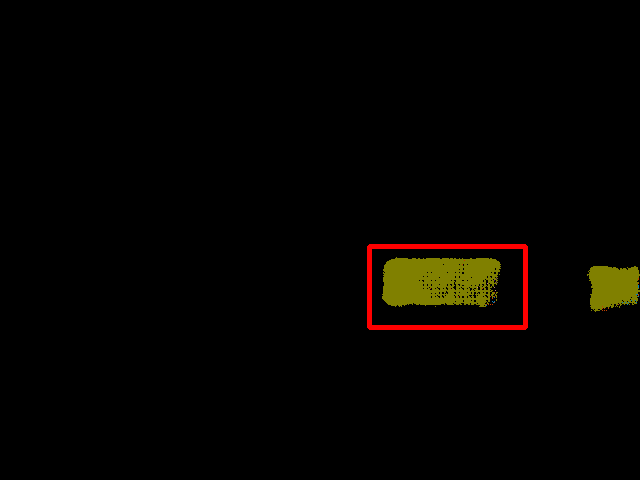}\hspace{0.5mm} &
			\includegraphics[width=1.85cm,height=1.85cm]{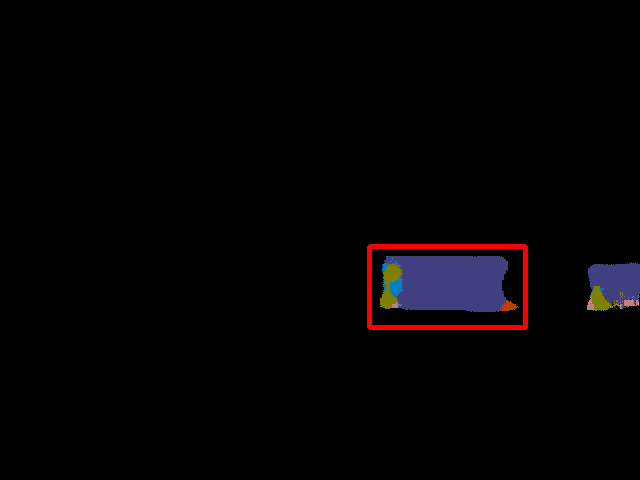}\hspace{0.5mm} &
			\includegraphics[width=1.85cm,height=1.85cm]{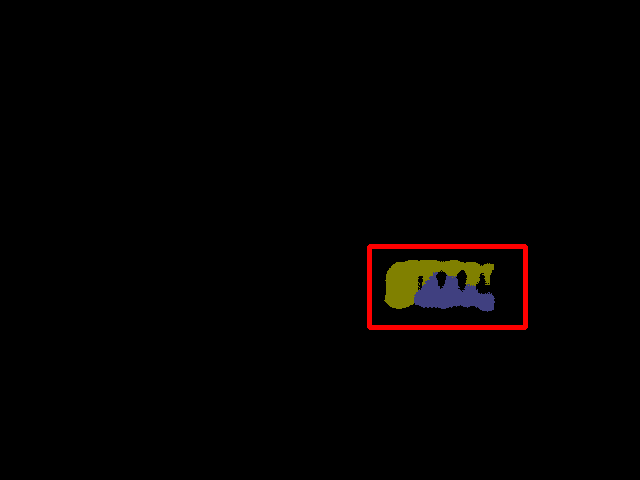}\hspace{0.5mm} &
                \includegraphics[width=1.85cm,height=1.85cm]{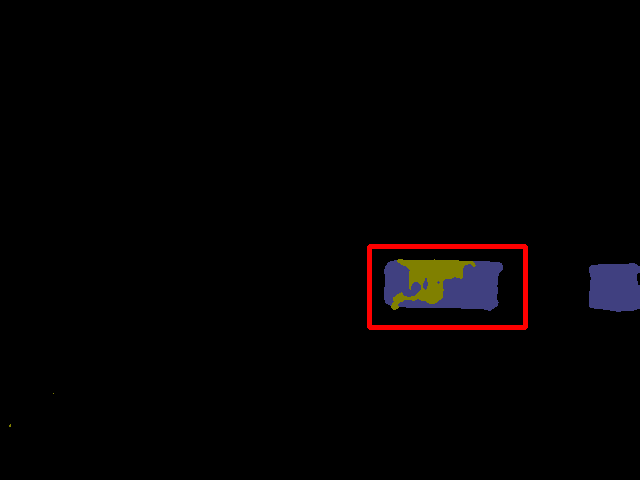}\hspace{0.5mm} &
                \includegraphics[width=1.85cm,height=1.85cm]{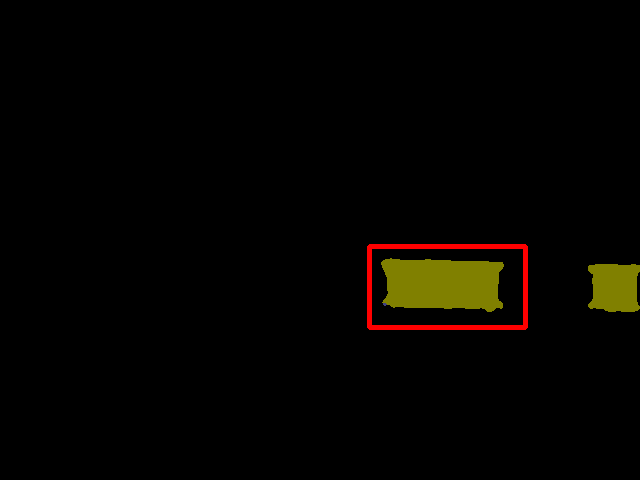}\hspace{0.5mm} &
                \includegraphics[width=1.85cm,height=1.85cm]{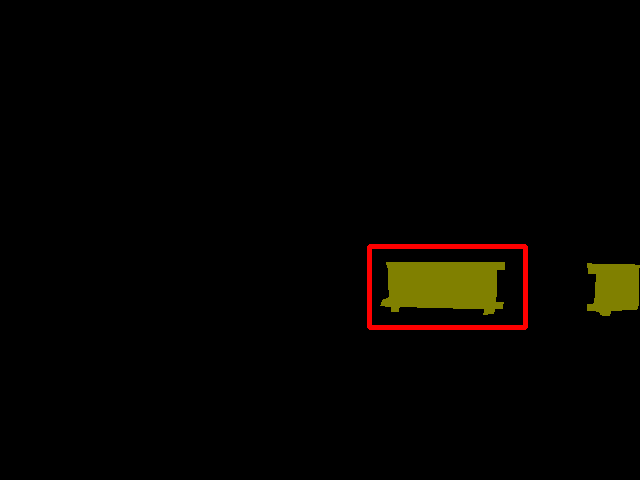}\hspace{0.5mm} \\
			\vspace{-0.5mm}
   \includegraphics[width=1.85cm,height=1.85cm]{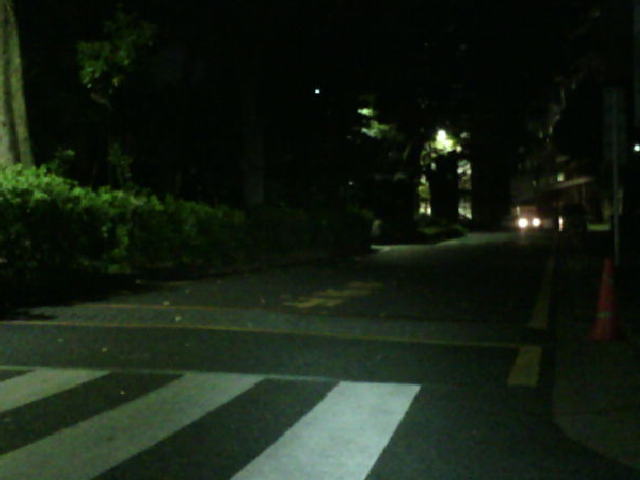}\hspace{0.5mm} &
			\includegraphics[width=1.85cm,height=1.85cm]{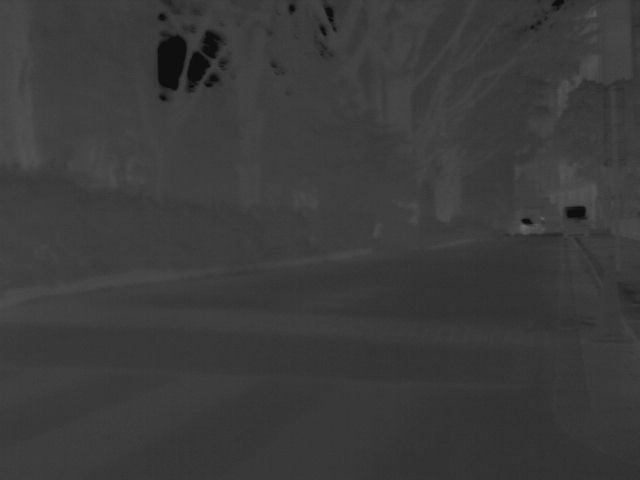}\hspace{0.5mm} &
			\includegraphics[width=1.85cm,height=1.85cm]{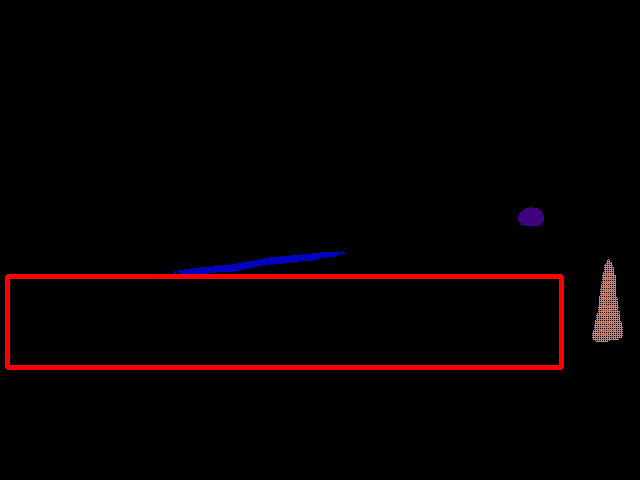}\hspace{0.5mm} &
			\includegraphics[width=1.85cm,height=1.85cm]{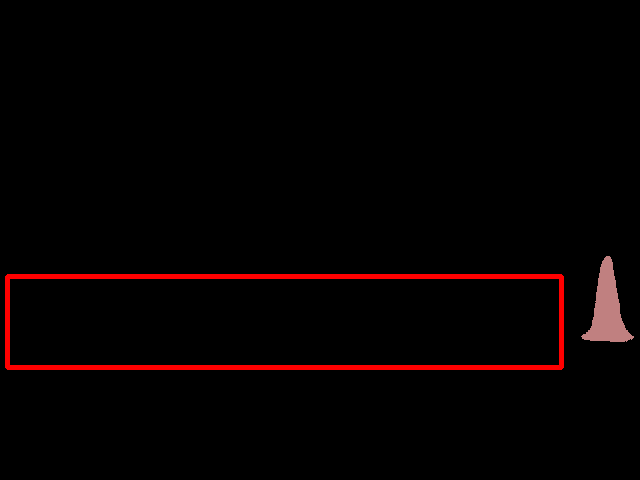}\hspace{0.5mm} &
			\includegraphics[width=1.85cm,height=1.85cm]{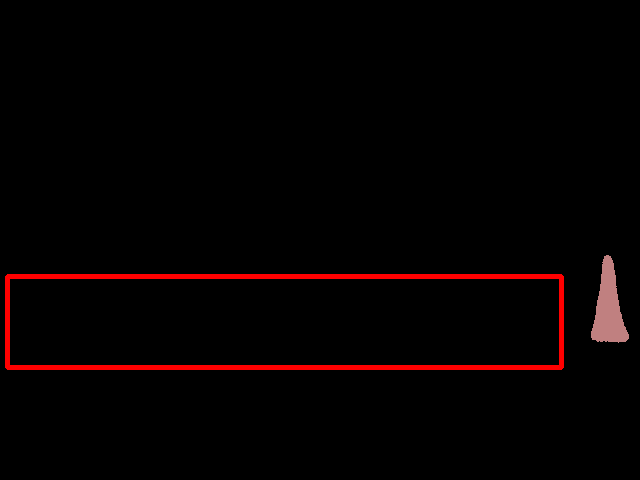}\hspace{0.5mm} &
                \includegraphics[width=1.85cm,height=1.85cm]{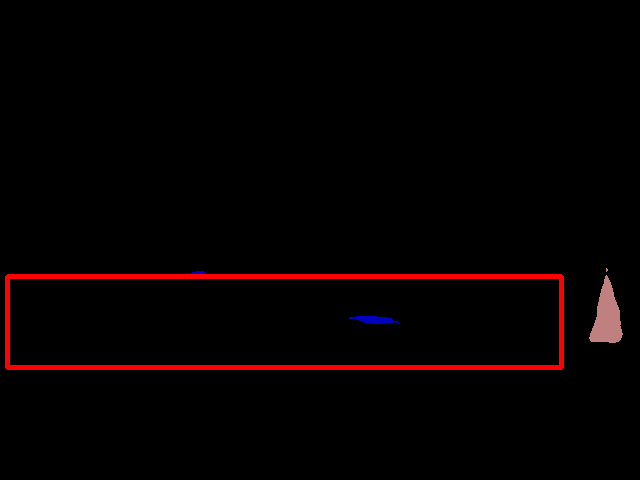}\hspace{0.5mm} &
                \includegraphics[width=1.85cm,height=1.85cm]{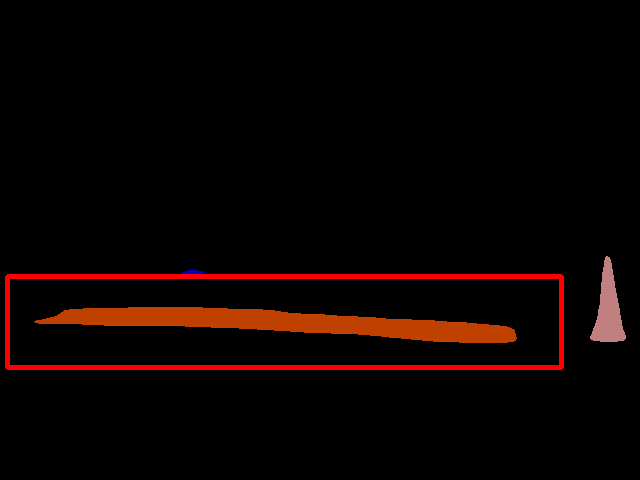}\hspace{0.5mm} &	
                \includegraphics[width=1.85cm,height=1.85cm]{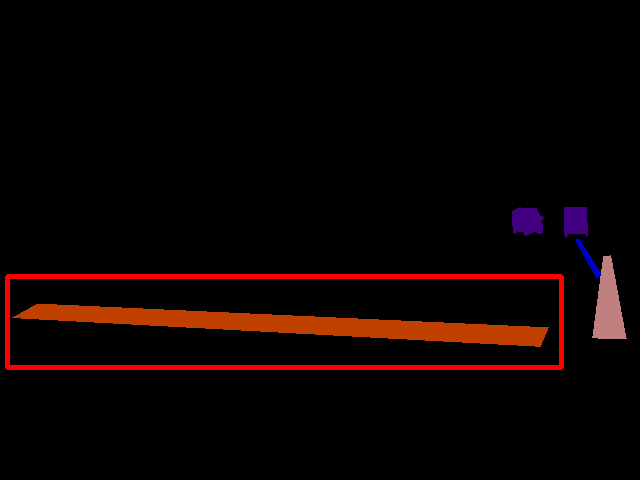}\hspace{0.5mm} \\
			\vspace{-0.5mm}
   \includegraphics[width=1.85cm,height=1.85cm]{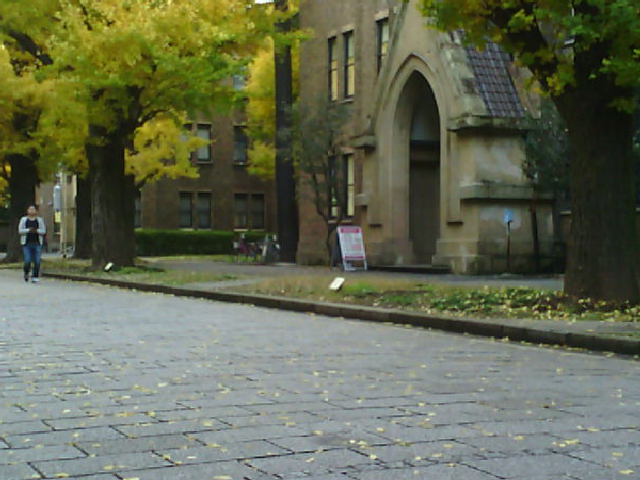}\hspace{0.5mm} &
			\includegraphics[width=1.85cm,height=1.85cm]{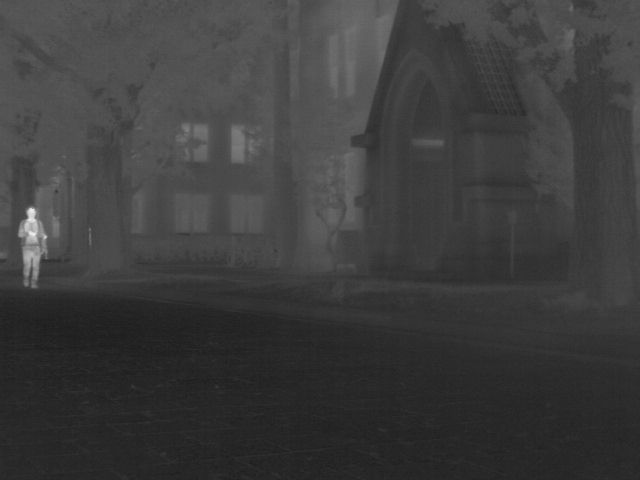}\hspace{0.5mm} &
			\includegraphics[width=1.85cm,height=1.85cm]{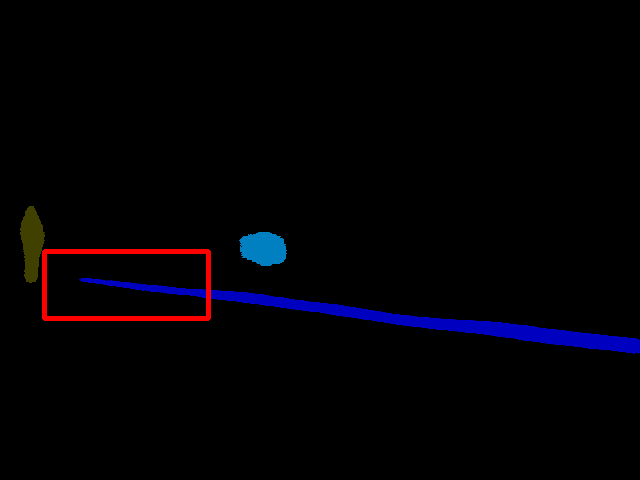}\hspace{0.5mm} &
			\includegraphics[width=1.85cm,height=1.85cm]{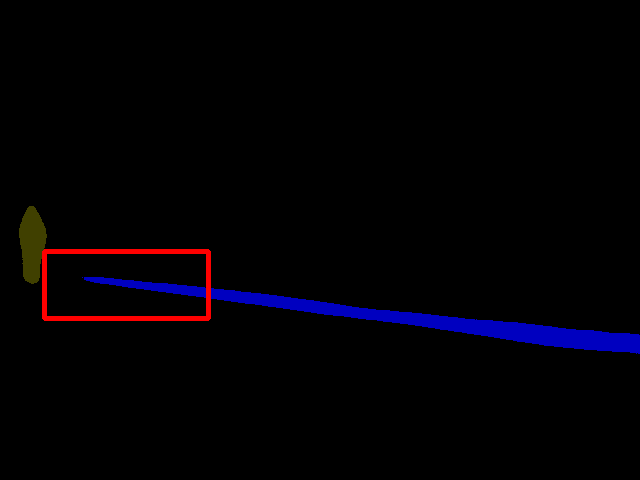}\hspace{0.5mm} &
			\includegraphics[width=1.85cm,height=1.85cm]{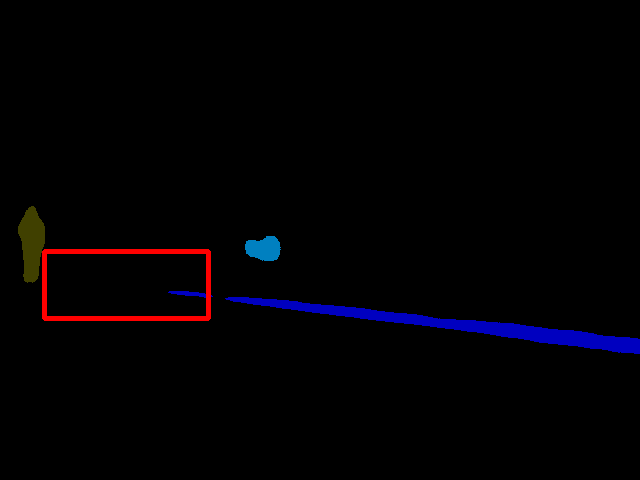}\hspace{0.5mm} &
                \includegraphics[width=1.85cm,height=1.85cm]{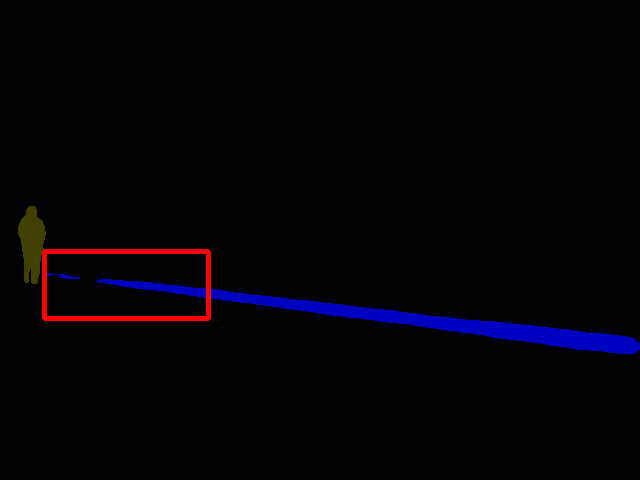}\hspace{0.5mm} 	&
                \includegraphics[width=1.85cm,height=1.85cm]{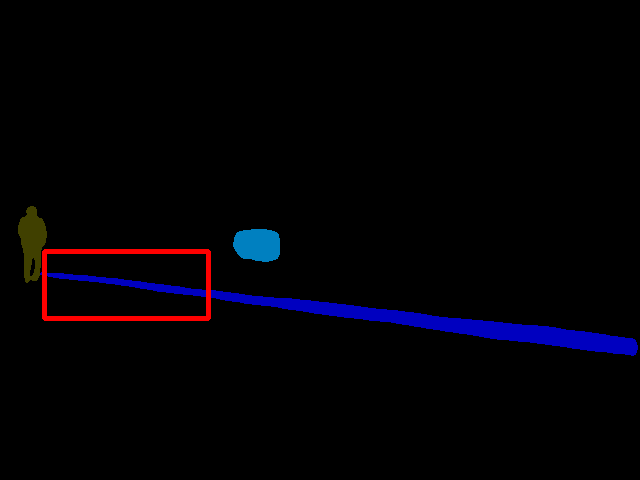}\hspace{0.5mm} &
                \includegraphics[width=1.85cm,height=1.85cm]{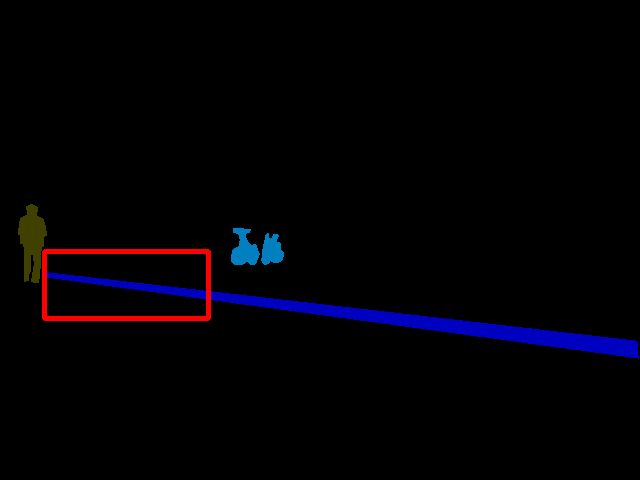}\hspace{0.5mm} \\
			\vspace{-0.5mm}
   \includegraphics[width=1.85cm,height=1.85cm]{figs/MFNet_Results/RGB/01404D.png}\hspace{0.5mm} &
			\includegraphics[width=1.85cm,height=1.85cm]{figs/MFNet_Results/Modal/01404D.png}\hspace{0.5mm} &
			\includegraphics[width=1.85cm,height=1.85cm]{figs/MFNet_Results/Pred_RTFNet/01404D.png}\hspace{0.5mm} &
			\includegraphics[width=1.85cm,height=1.85cm]{figs/MFNet_Results/Pred_FEANet/01404D.png}\hspace{0.5mm} &
			\includegraphics[width=1.85cm,height=1.85cm]{figs/MFNet_Results/Pred_EAEFNet/01404D.png}\hspace{0.5mm} &
                \includegraphics[width=1.85cm,height=1.85cm]{figs/MFNet_Results/Pred_CMX/01404D.png}\hspace{0.5mm} &
                \includegraphics[width=1.85cm,height=1.85cm]{figs/MFNet_Results/Pred_Sigma/01404D.png}\hspace{0.5mm} &
                \includegraphics[width=1.85cm,height=1.85cm]{figs/MFNet_Results/gt/01404D.png}\hspace{0.5mm} \\
			\vspace{-0.5mm}
   \includegraphics[width=1.85cm,height=1.85cm]{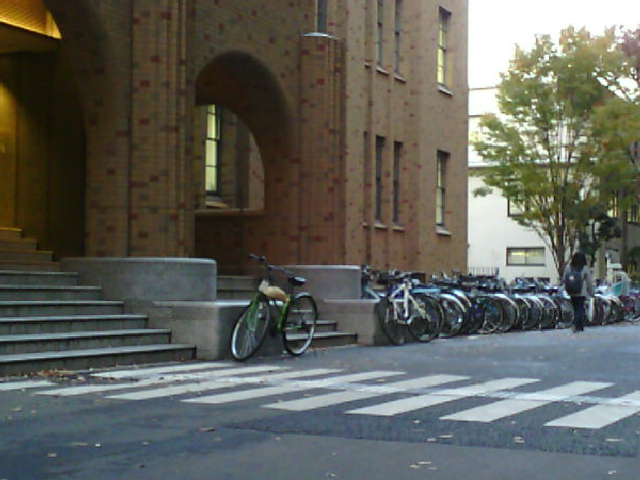}\hspace{0.5mm} &
			\includegraphics[width=1.85cm,height=1.85cm]{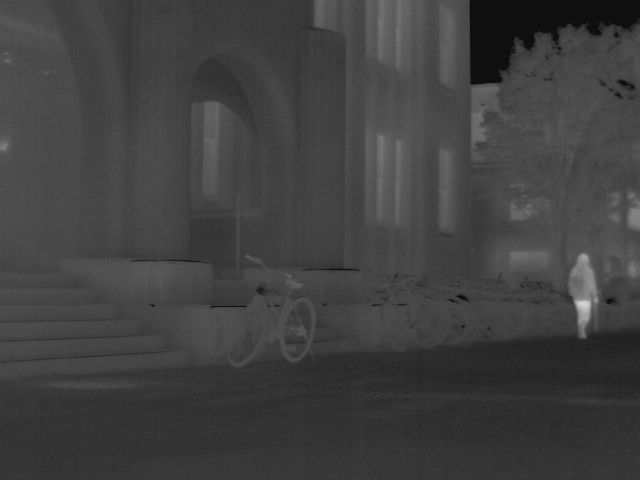}\hspace{0.5mm} &
			\includegraphics[width=1.85cm,height=1.85cm]{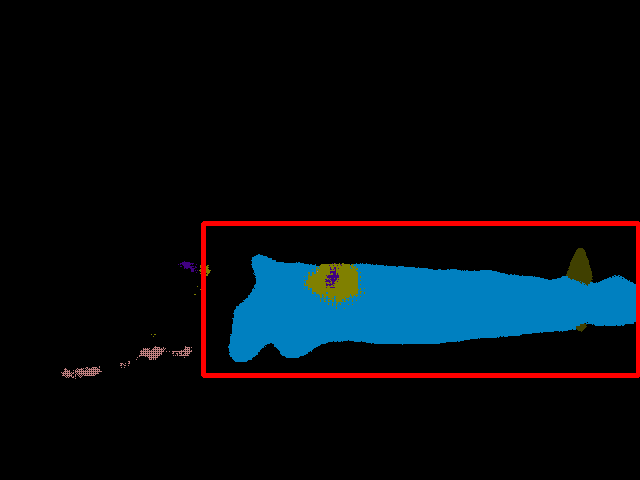}\hspace{0.5mm} &
			\includegraphics[width=1.85cm,height=1.85cm]{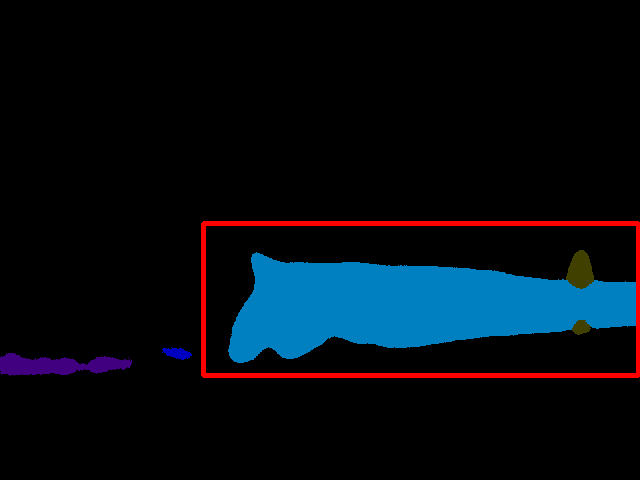}\hspace{0.5mm} &
			\includegraphics[width=1.85cm,height=1.85cm]{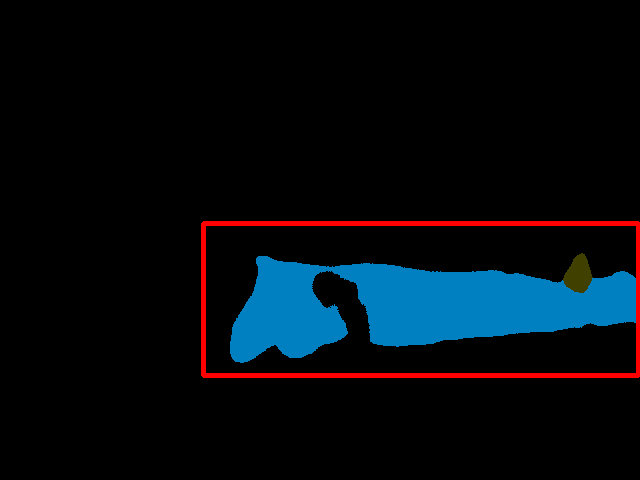}\hspace{0.5mm} &
                \includegraphics[width=1.85cm,height=1.85cm]{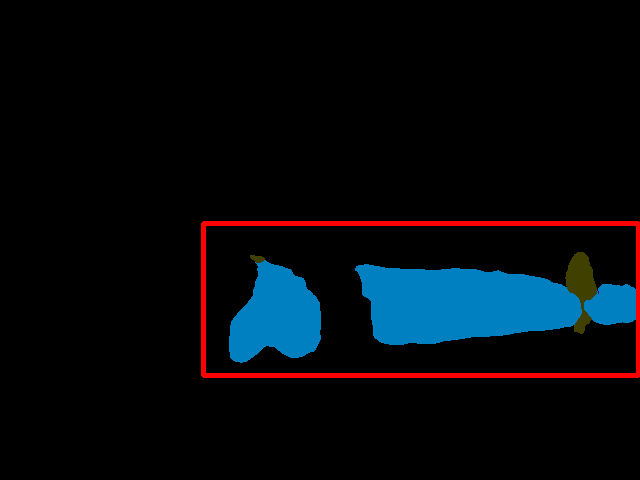}\hspace{0.5mm} &
                \includegraphics[width=1.85cm,height=1.85cm]{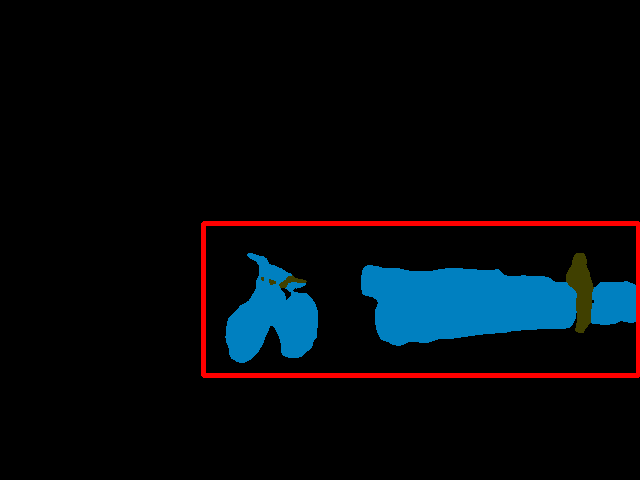}\hspace{0.5mm} &
                \includegraphics[width=1.85cm,height=1.85cm]{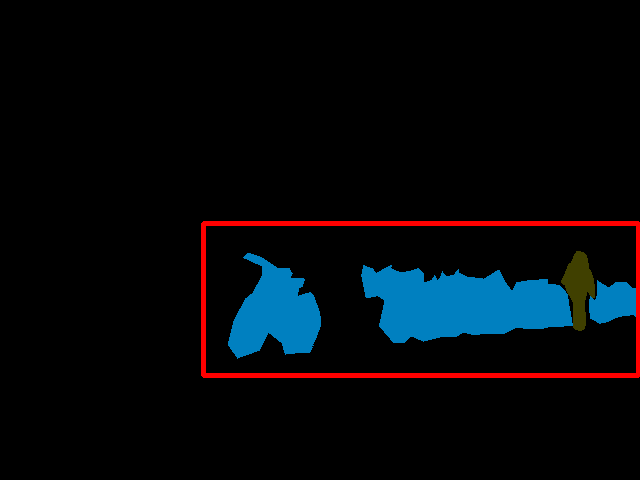}\hspace{0.5mm} \\
			\vspace{-0.5mm}
			\includegraphics[width=1.85cm,height=1.85cm]{figs/MFNet_Results/RGB/01435D.png}\hspace{0.5mm} &
			\includegraphics[width=1.85cm,height=1.85cm]{figs/MFNet_Results/Modal/01435D.png}\hspace{0.5mm} &
			\includegraphics[width=1.85cm,height=1.85cm]{figs/MFNet_Results/Pred_RTFNet/01435D.png}\hspace{0.5mm} &
			\includegraphics[width=1.85cm,height=1.85cm]{figs/MFNet_Results/Pred_FEANet/01435D.png}\hspace{0.5mm} &
			\includegraphics[width=1.85cm,height=1.85cm]{figs/MFNet_Results/Pred_EAEFNet/01435D.png}\hspace{0.5mm} &
                \includegraphics[width=1.85cm,height=1.85cm]{figs/MFNet_Results/Pred_CMX/01435D.png}\hspace{0.5mm} &
                \includegraphics[width=1.85cm,height=1.85cm]{figs/MFNet_Results/Pred_Sigma/01435D.png}\hspace{0.5mm} &
                \includegraphics[width=1.85cm,height=1.85cm]{figs/MFNet_Results/gt/01435D.png}\hspace{0.5mm} & \vspace{-1.2mm} \\ [1pt]
			{\scriptsize RGB} & {\scriptsize Thermal} & {\scriptsize RTFNet~\cite{sun2019rtfnet}} & {\scriptsize FEANet~\cite{deng2021feanet}} & {\scriptsize EAEFNet~\cite{liang2023explicit}} & {\scriptsize CMX~\cite{liu2022cmx}} & {\scriptsize Sigma (Ours)} & {\scriptsize Ground Truth}\ \\
		\end{tabular}
	}
	\vspace{-2.8mm}
	\caption{Qualitative comparison on MFNet~\cite{ha2017mfnet} dataset. More qualitative results can be found in the appendix.}
	\label{fig:qualitative mfnet supp}

 \vspace{-3pt}
\end{figure*}

\begin{figure*}[!]
	\centering
 \vspace{-10pt}
	\resizebox{1\textwidth}{!}
	{
		\begin{tabular}{@{}c@{}c@{}c@{}c@{}c@{}c@{}c@{}c@{}c@{}c@{}c}
			\vspace{-0.5mm}
   \includegraphics[width=1.85cm,height=1.85cm]{figs/NYU_Results/RGB/1298.jpg}\hspace{0.5mm} &
			\includegraphics[width=1.85cm,height=1.85cm]{figs/NYU_Results/Modal/1298.jpg}\hspace{0.5mm} &
			\includegraphics[width=1.85cm,height=1.85cm]{figs/NYU_Results/pred_sagate/1298.png}\hspace{0.5mm} &
			\includegraphics[width=1.85cm,height=1.85cm]{figs/NYU_Results/pred_multimae/1298.png}\hspace{0.5mm} &
			\includegraphics[width=1.85cm,height=1.85cm]{figs/NYU_Results/pred_cmx/1298.png}\hspace{0.5mm} &
                \includegraphics[width=1.85cm,height=1.85cm]{figs/NYU_Results/pred_cmnext/1298.png}\hspace{0.5mm} &
                \includegraphics[width=1.85cm,height=1.85cm]{figs/NYU_Results/pred_sigma/1298.png}\hspace{0.5mm} &
                \includegraphics[width=1.85cm,height=1.85cm]{figs/NYU_Results/gt/1298.png}\hspace{0.5mm} 
			 \\
			\vspace{-0.5mm}
   \includegraphics[width=1.85cm,height=1.85cm]{figs/NYU_Results/RGB/1191.jpg}\hspace{0.5mm} &
			\includegraphics[width=1.85cm,height=1.85cm]{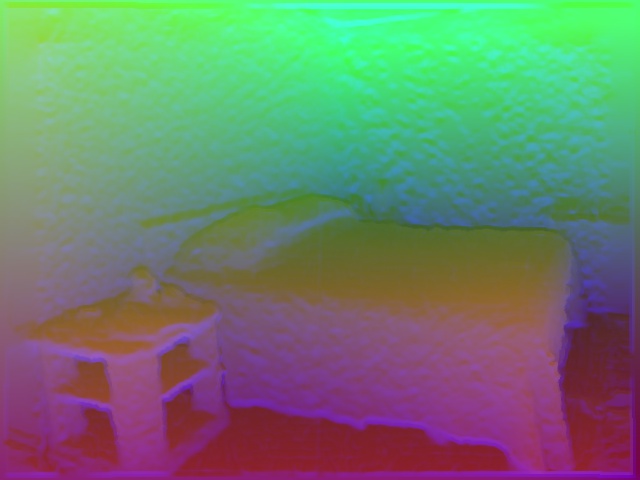}\hspace{0.5mm} &
			\includegraphics[width=1.85cm,height=1.85cm]{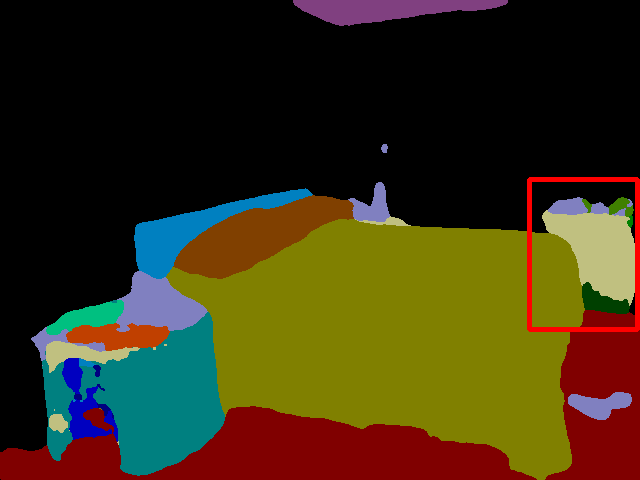}\hspace{0.5mm} &
			\includegraphics[width=1.85cm,height=1.85cm]{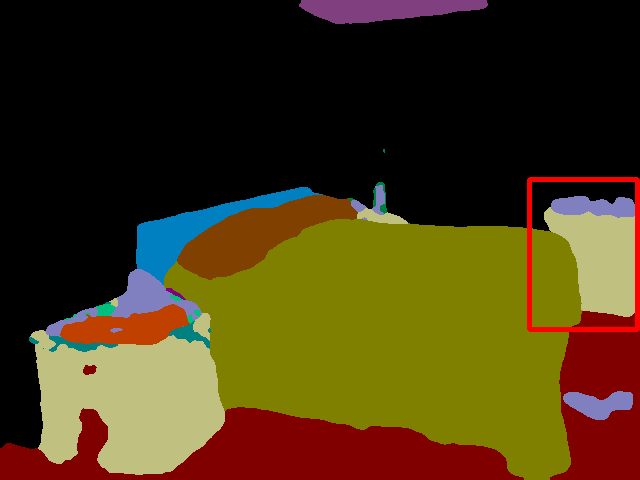}\hspace{0.5mm} &
			\includegraphics[width=1.85cm,height=1.85cm]{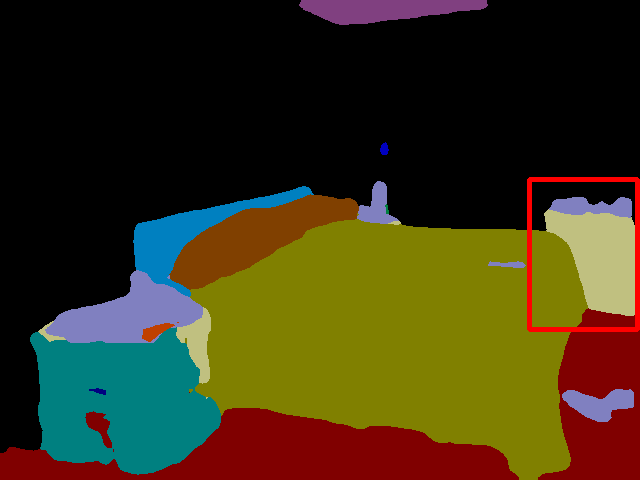}\hspace{0.5mm} &
                \includegraphics[width=1.85cm,height=1.85cm]{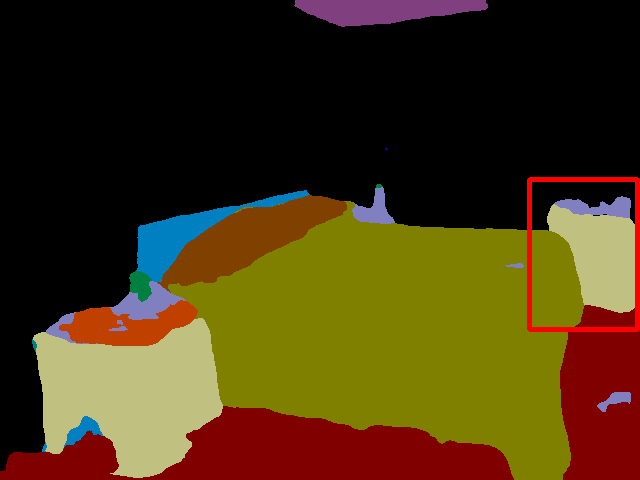}\hspace{0.5mm} &
                \includegraphics[width=1.85cm,height=1.85cm]{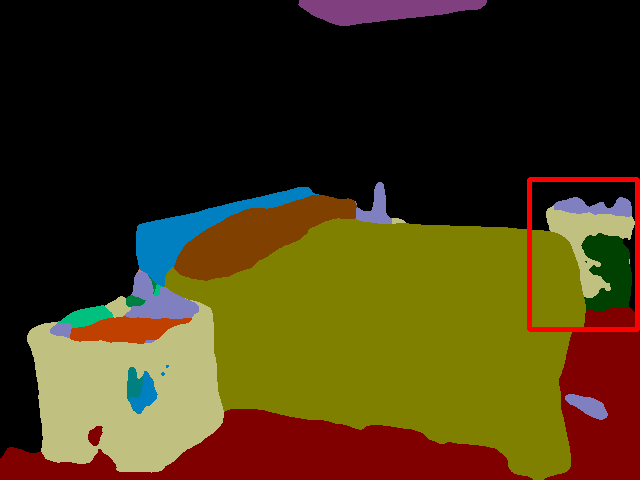}\hspace{0.5mm} &	
                \includegraphics[width=1.85cm,height=1.85cm]{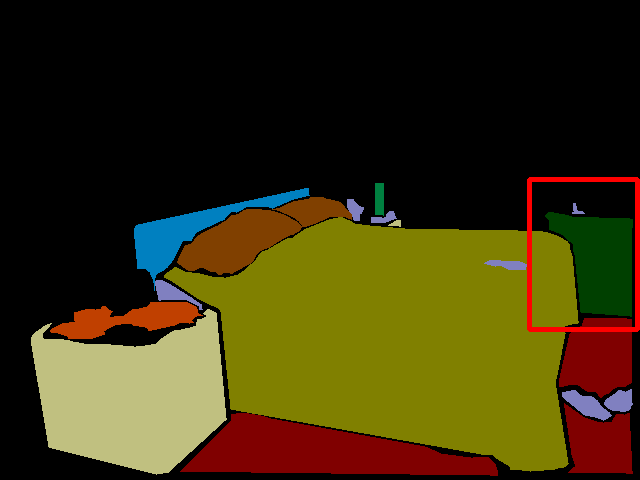}\hspace{0.5mm} \\
			\vspace{-0.5mm}
   \includegraphics[width=1.85cm,height=1.85cm]{figs/NYU_Results/RGB/36.jpg}\hspace{0.5mm} &
			\includegraphics[width=1.85cm,height=1.85cm]{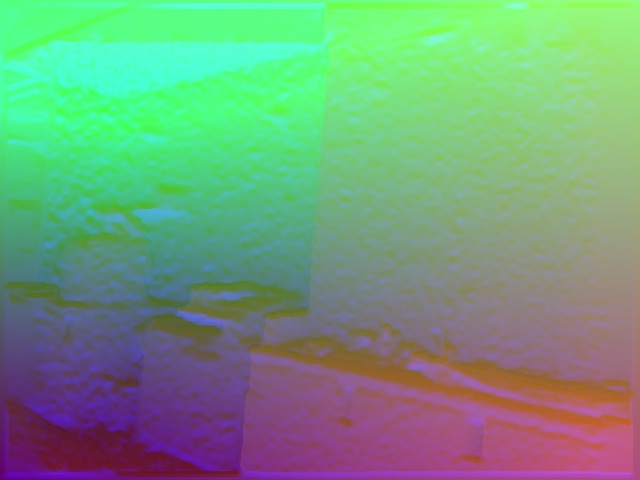}\hspace{0.5mm} &
			\includegraphics[width=1.85cm,height=1.85cm]{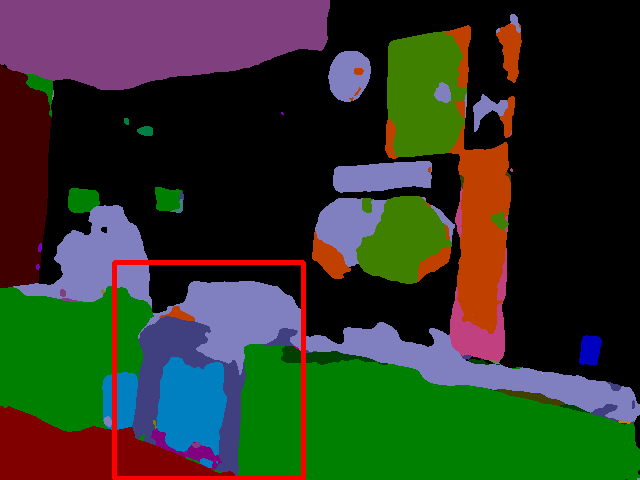}\hspace{0.5mm} &
			\includegraphics[width=1.85cm,height=1.85cm]{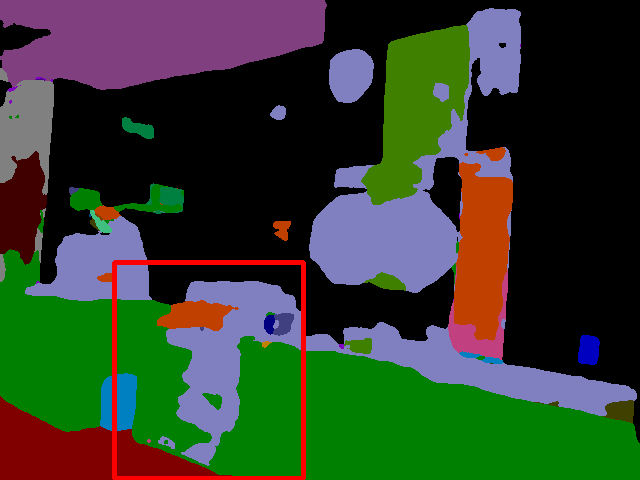}\hspace{0.5mm} &
			\includegraphics[width=1.85cm,height=1.85cm]{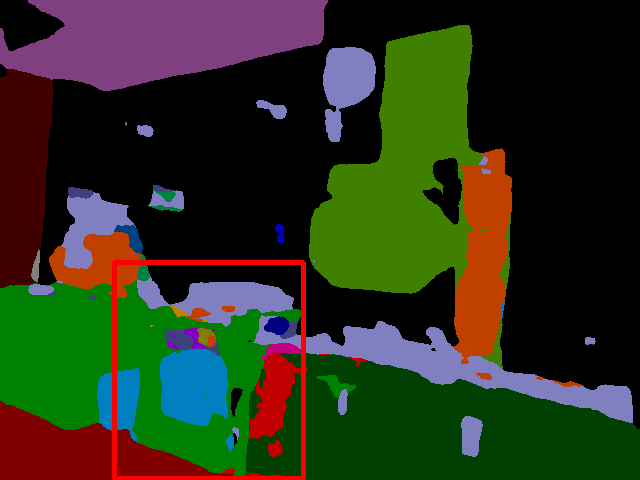}\hspace{0.5mm} &
                \includegraphics[width=1.85cm,height=1.85cm]{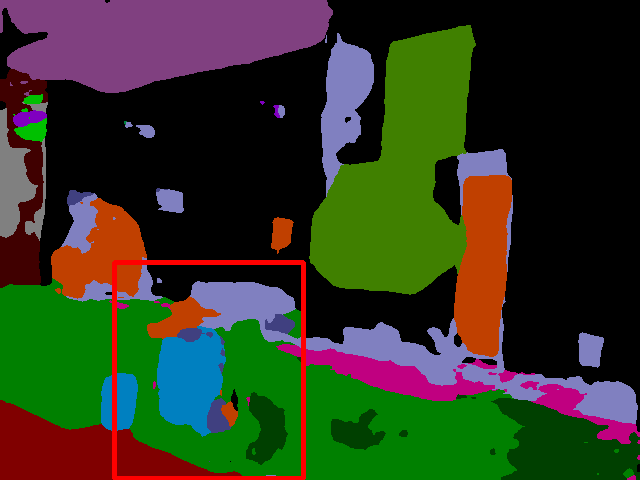}\hspace{0.5mm} &
                \includegraphics[width=1.85cm,height=1.85cm]{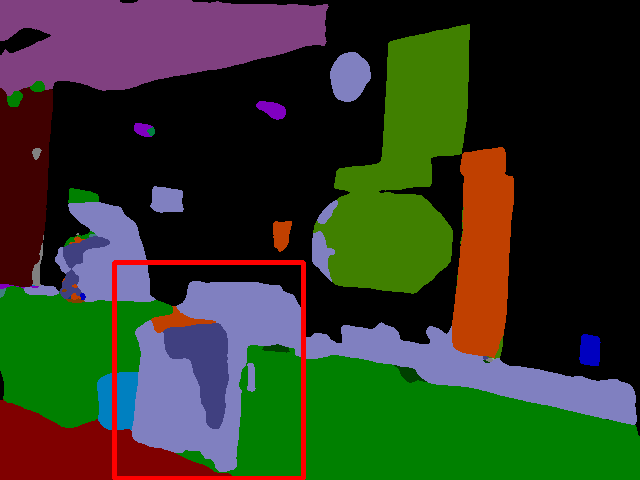}\hspace{0.5mm} &	
                \includegraphics[width=1.85cm,height=1.85cm]{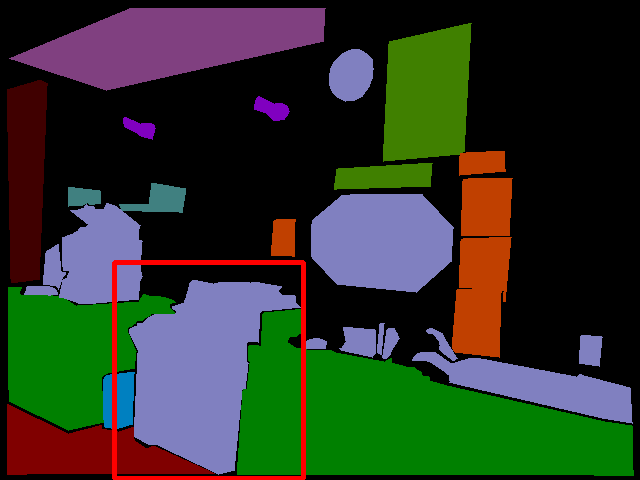}\hspace{0.5mm} \\
			\vspace{-0.5mm}

\includegraphics[width=1.85cm,height=1.85cm]{figs/NYU_Results/RGB/189.jpg}\hspace{0.5mm} &
\includegraphics[width=1.85cm,height=1.85cm]{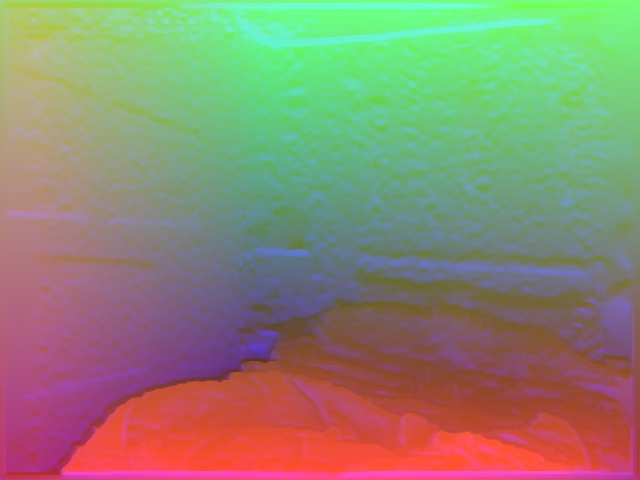}\hspace{0.5mm} &
\includegraphics[width=1.85cm,height=1.85cm]{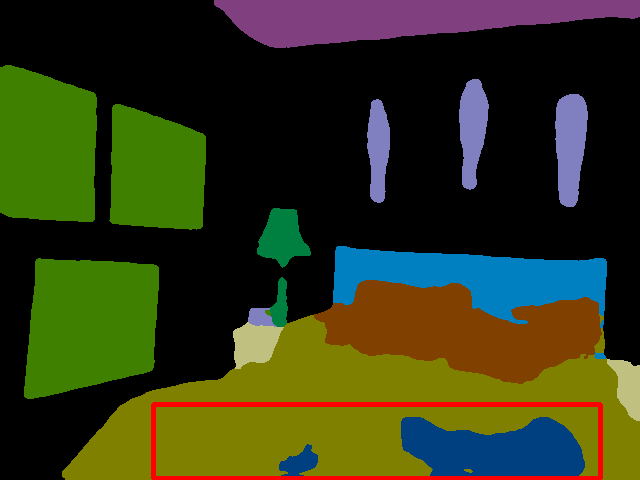}\hspace{0.5mm} &
\includegraphics[width=1.85cm,height=1.85cm]{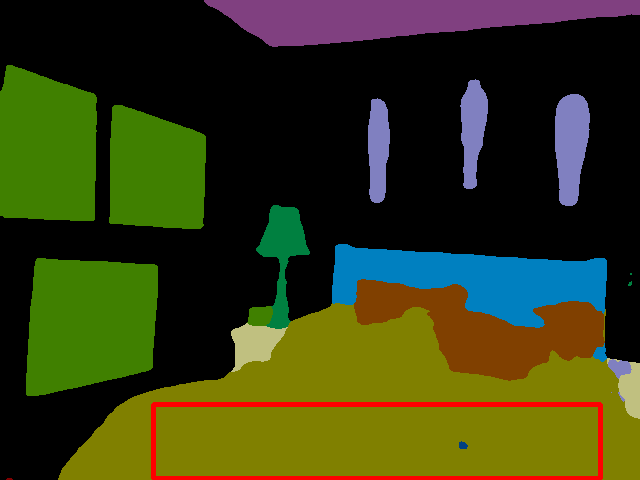}\hspace{0.5mm} &
\includegraphics[width=1.85cm,height=1.85cm]{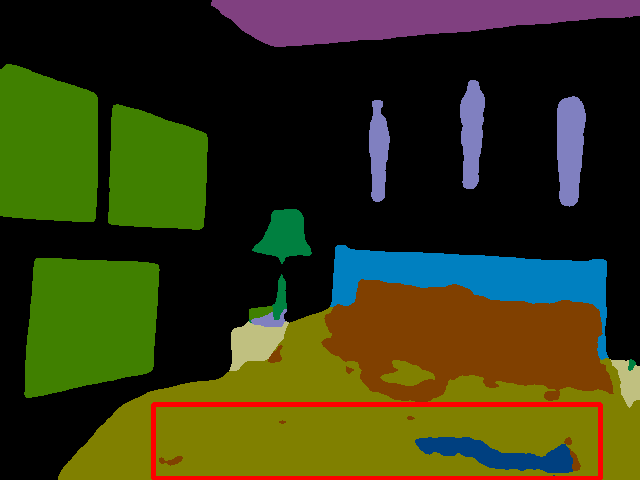}\hspace{0.5mm} &
\includegraphics[width=1.85cm,height=1.85cm]{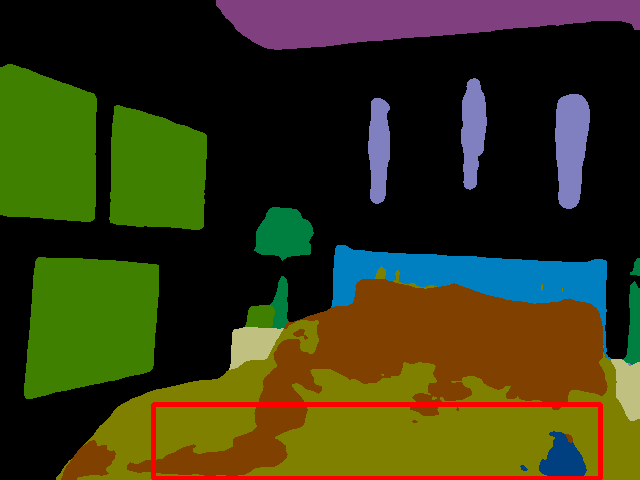}\hspace{0.5mm} &
\includegraphics[width=1.85cm,height=1.85cm]{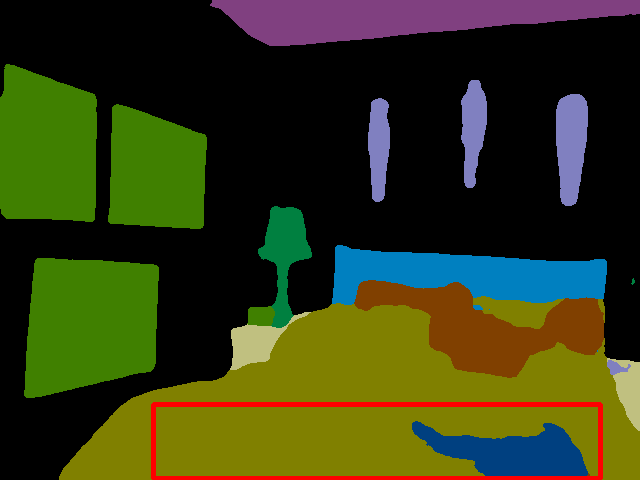}\hspace{0.5mm} &	
\includegraphics[width=1.85cm,height=1.85cm]{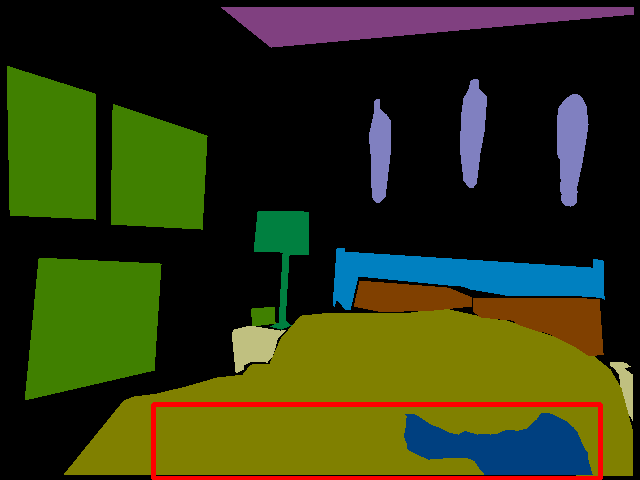}\hspace{0.5mm} \\
\vspace{-0.5mm}

\includegraphics[width=1.85cm,height=1.85cm]{figs/NYU_Results/RGB/549.jpg}\hspace{0.5mm} &
\includegraphics[width=1.85cm,height=1.85cm]{figs/NYU_Results/Modal/549.jpg}\hspace{0.5mm} &
\includegraphics[width=1.85cm,height=1.85cm]{figs/NYU_Results/pred_sagate/549.png}\hspace{0.5mm} &
\includegraphics[width=1.85cm,height=1.85cm]{figs/NYU_Results/pred_multimae/549.png}\hspace{0.5mm} &
\includegraphics[width=1.85cm,height=1.85cm]{figs/NYU_Results/pred_cmx/549.png}\hspace{0.5mm} &
\includegraphics[width=1.85cm,height=1.85cm]{figs/NYU_Results/pred_cmnext/549.png}\hspace{0.5mm} &
\includegraphics[width=1.85cm,height=1.85cm]{figs/NYU_Results/pred_sigma/549.png}\hspace{0.5mm} &	
\includegraphics[width=1.85cm,height=1.85cm]{figs/NYU_Results/gt/549.png}\hspace{0.5mm} \\
\vspace{-0.5mm}

\includegraphics[width=1.85cm,height=1.85cm]{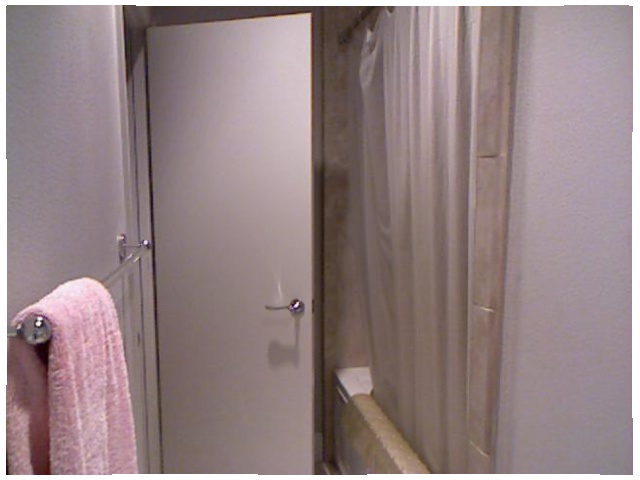}\hspace{0.5mm} &
\includegraphics[width=1.85cm,height=1.85cm]{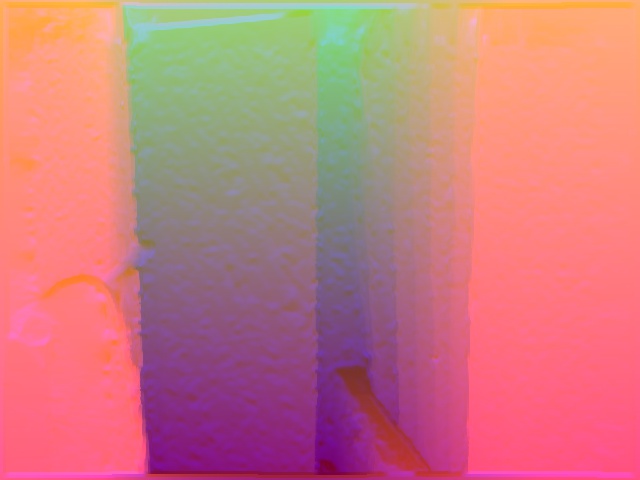}\hspace{0.5mm} &
\includegraphics[width=1.85cm,height=1.85cm]{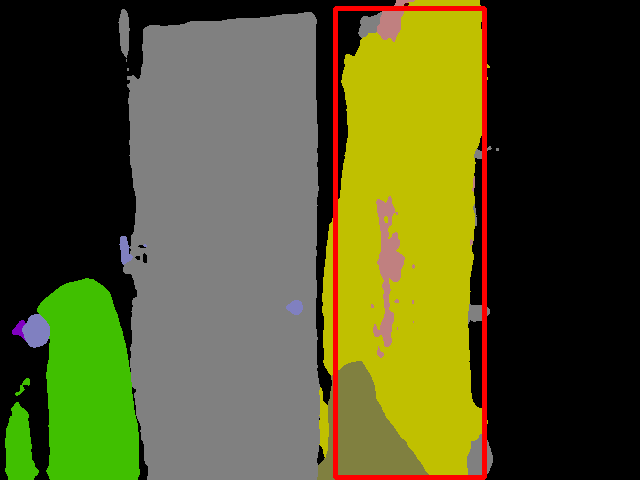}\hspace{0.5mm} &
\includegraphics[width=1.85cm,height=1.85cm]{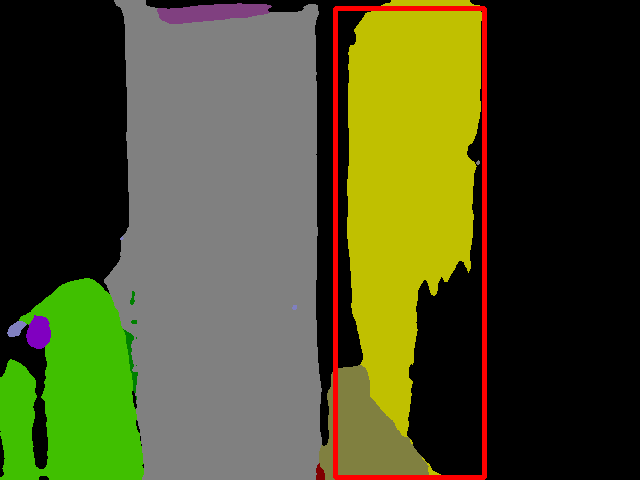}\hspace{0.5mm} &
\includegraphics[width=1.85cm,height=1.85cm]{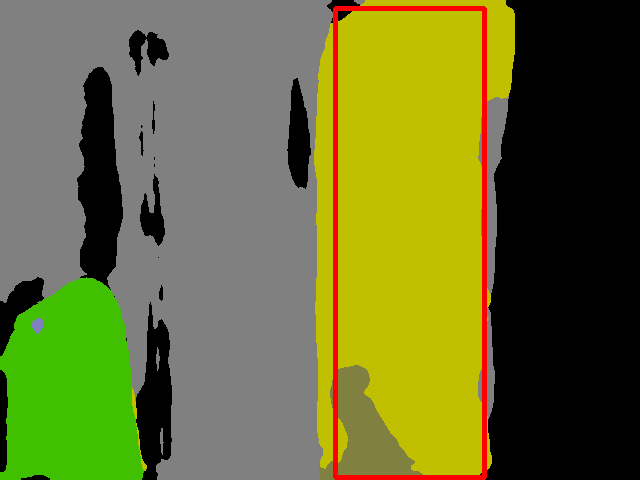}\hspace{0.5mm} &
\includegraphics[width=1.85cm,height=1.85cm]{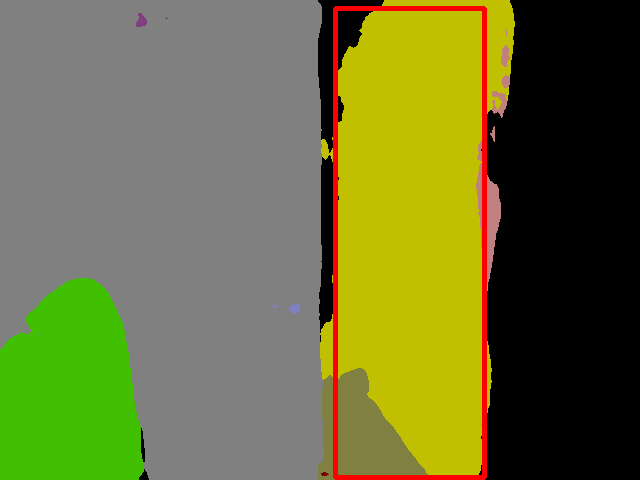}\hspace{0.5mm} &
\includegraphics[width=1.85cm,height=1.85cm]{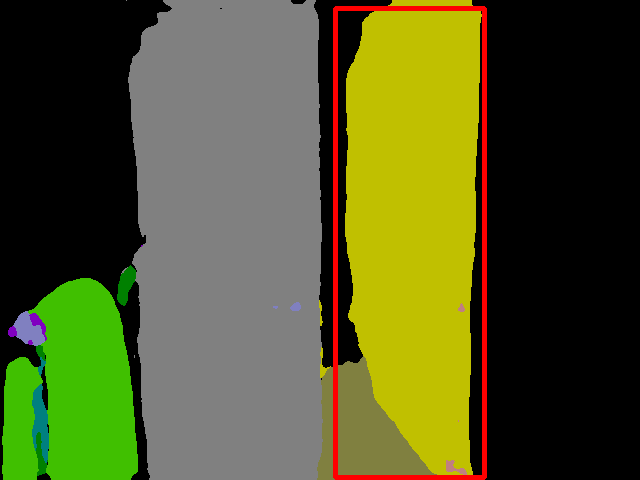}\hspace{0.5mm} &	
\includegraphics[width=1.85cm,height=1.85cm]{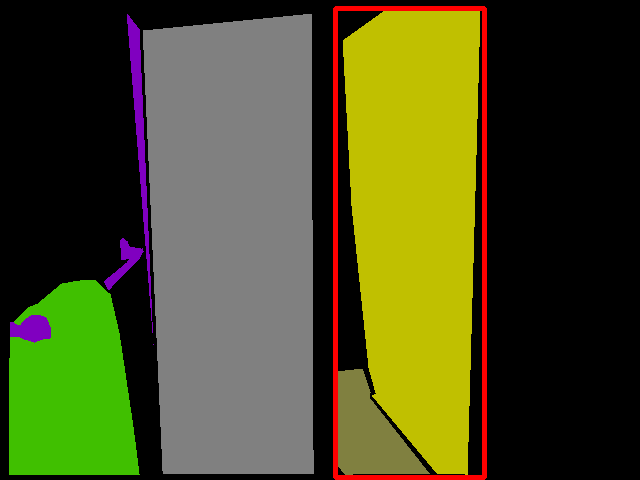}\hspace{0.5mm} \\
\vspace{-0.5mm}
   \includegraphics[width=1.85cm,height=1.85cm]{figs/NYU_Results/RGB/870.jpg}\hspace{0.5mm} &
			\includegraphics[width=1.85cm,height=1.85cm]{figs/NYU_Results/Modal/870.jpg}\hspace{0.5mm} &
			\includegraphics[width=1.85cm,height=1.85cm]{figs/NYU_Results/pred_sagate/870.png}\hspace{0.5mm} &
			\includegraphics[width=1.85cm,height=1.85cm]{figs/NYU_Results/pred_multimae/870.png}\hspace{0.5mm} &
			\includegraphics[width=1.85cm,height=1.85cm]{figs/NYU_Results/pred_cmx/870.png}\hspace{0.5mm} &
                \includegraphics[width=1.85cm,height=1.85cm]{figs/NYU_Results/pred_cmnext/870.png}\hspace{0.5mm} &
                \includegraphics[width=1.85cm,height=1.85cm]{figs/NYU_Results/pred_sigma/870.png}\hspace{0.5mm} &
                \includegraphics[width=1.85cm,height=1.85cm]{figs/NYU_Results/gt/870.png}\hspace{0.5mm} \\
			\vspace{-0.5mm}
        \includegraphics[width=1.85cm,height=1.85cm]{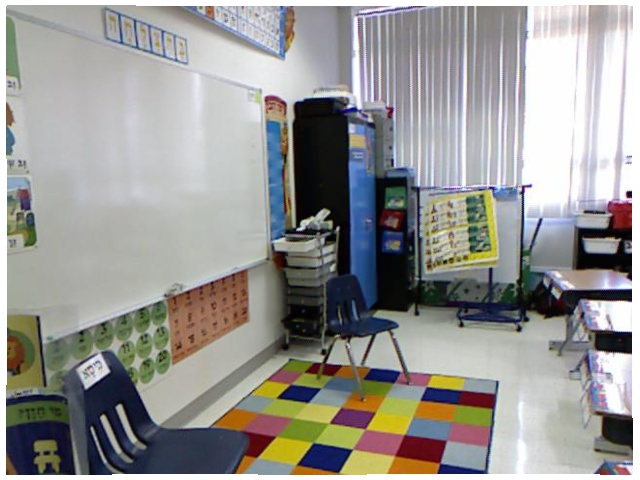}\hspace{0.5mm} &
			\includegraphics[width=1.85cm,height=1.85cm]{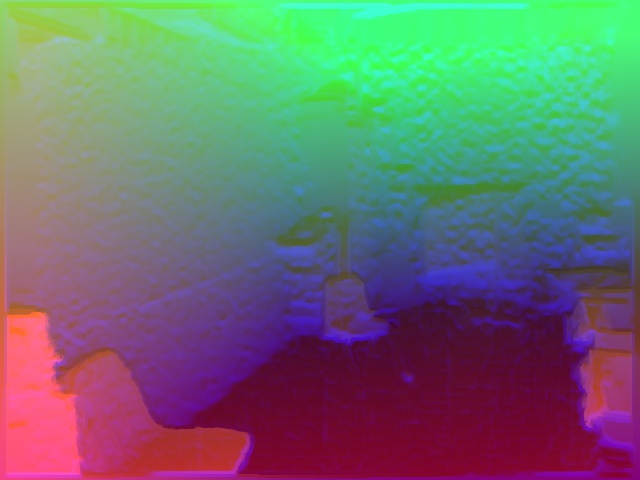}\hspace{0.5mm} &
			\includegraphics[width=1.85cm,height=1.85cm]{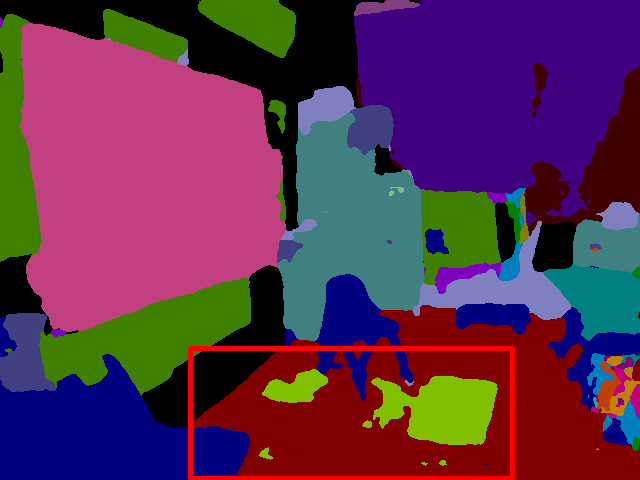}\hspace{0.5mm} &
			\includegraphics[width=1.85cm,height=1.85cm]{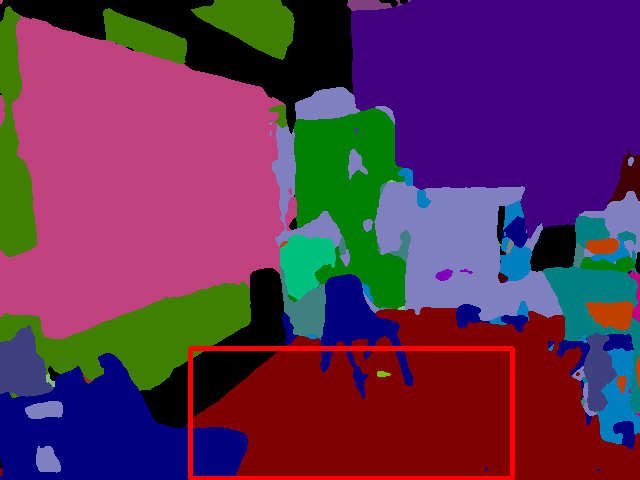}\hspace{0.5mm} &
			\includegraphics[width=1.85cm,height=1.85cm]{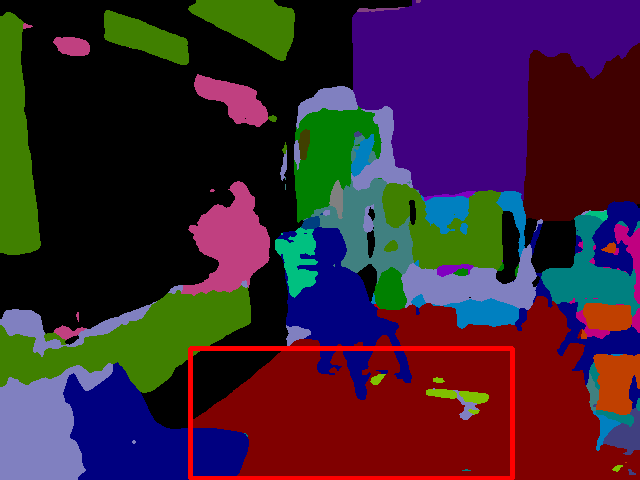}\hspace{0.5mm} &
                \includegraphics[width=1.85cm,height=1.85cm]{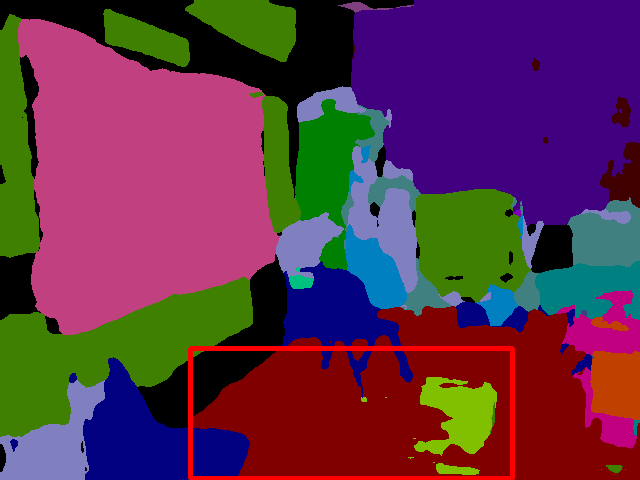}\hspace{0.5mm} &
                \includegraphics[width=1.85cm,height=1.85cm]{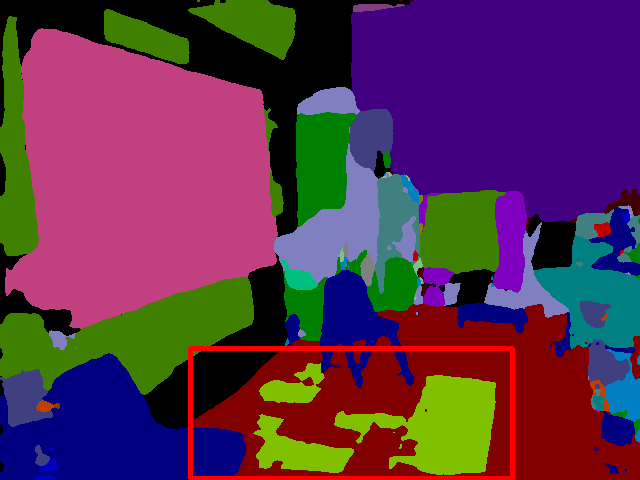}\hspace{0.5mm} &
                \includegraphics[width=1.85cm,height=1.85cm]{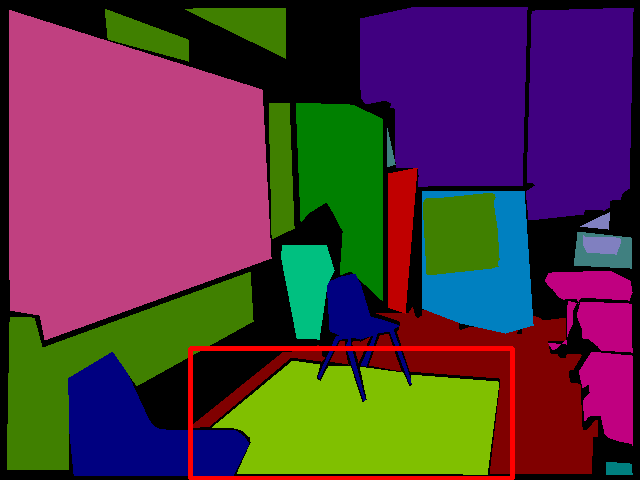}\hspace{0.5mm} &
			\vspace{-1.2mm} \\ [1pt]
			{\scriptsize RGB} & {\scriptsize HHA} & {\scriptsize SA-Gate~\cite{chen2020bi}} & {\scriptsize MultiMAE~\cite{bachmann2022multimae}} & {\scriptsize CMX~\cite{liu2022cmx}} & {\scriptsize CMNeXt~\cite{zhang2023delivering}} & {\scriptsize Sigma (Ours)} & {\scriptsize Ground Truth}\ \\
		\end{tabular}
	}
	\vspace{-2.8mm}
	\caption{Qualitative comparison on NYU Depth V2~\cite{silberman2012indoor} dataset. We use HHA images for better visualization of depth modality. More qualitative results can be found in the appendix.}
	\label{fig:qualitative nyudepth supp}
\vspace{-15pt}
\end{figure*}

In Fig.~\ref{fig:qualitative mfnet supp} and Fig.~\ref{fig:qualitative nyudepth supp}, we show more qualitative results of our method compared to others.

\end{document}